\documentclass[journal]{IEEEtran}

\IEEEoverridecommandlockouts        

\usepackage{cite}
\usepackage{amsmath,amssymb,amsfonts}
\usepackage{graphicx}
\usepackage{textcomp}
\usepackage{xcolor}
\usepackage{color}													
\usepackage{xfrac}	
\usepackage{colortbl}
\usepackage{siunitx}
\usepackage[english]{babel}	
\usepackage{multirow}
\usepackage[hidelinks]{hyperref}
\usepackage{blindtext}
\usepackage{mathtools,tikz}
\usepackage{hyperref}
\usepackage{tablefootnote}
\usepackage{algorithmic} 
\usepackage{textcomp} 
\usepackage{wasysym}
\usepackage{gensymb} 
\usepackage[ruled, linesnumbered]{algorithm2e}
\usepackage[caption=false, font=footnotesize]{subfig}
\usepackage{tabularx}
\usepackage{enumerate}
\usepackage{enumitem}
\usepackage{tikz}

\newif\ifhighlightchanges

\ifhighlightchanges
	\newcommand{\highlightred}[1]{\textcolor{red}{#1}}
\else
	\newcommand{\highlightred}[1]{#1}
\fi

\newcommand{\bs}[1]{\boldsymbol{#1}}
\newcommand{\mr}[1]{\mathrm{#1}}

\newcommand{\akr}{SafePR}
\newcommand{\anonymbin}{false}
\newcommand{\anonym}[3]{\ifthenelse{\equal{#1}{true}}{#2} {#3} }

\definecolor{gruen}{RGB}{0,152,70}
\definecolor{blau}{RGB}{0,0,255}
\definecolor{orange}{RGB}{255,110,0}
\definecolor{olivengruen}{RGB}{110, 117, 14}
\definecolor{gruenmarker}{RGB}{0, 158, 34}
\definecolor{klemmungfarbe}{RGB}{124,60,70}
\definecolor{kollisionfarbe}{RGB}{184,122,50}
\definecolor{platform}{HTML}{ceb301}
\definecolor{1link}{HTML}{c20078}
\definecolor{2link}{HTML}{e50000}
\definecolor{clamping}{HTML}{9a0eea}
\definecolor{dq}{HTML}{13eac9}
\definecolor{ddq}{HTML}{677a04}
\definecolor{tau}{HTML}{5a7d9a}
\definecolor{platform2}{HTML}{c79fef}
\definecolor{1link2}{HTML}{be03fd}
\definecolor{2link2}{HTML}{caa0ff}
\definecolor{clamping2}{HTML}{06c2ac}

\def\spc{7pt} 

\tikzset{myarrow/.style={-stealth,shorten >=\spc, shorten <=\spc}}

\title{\LARGE \bf
	\akr: Unified Approach for Safe Parallel Robots by\\
	Contact Detection and Reaction with Redundancy Resolution
}

\author{
	\anonym{\anonymbin}{Anonymous}{
	Aran Mohammad, Tim-Lukas Habich, Thomas Seel and Moritz Schappler
	\thanks{The authors acknowledge the support of the German Research Foundation~(DFG) under grant number 444769341.
			All authors are with the Institute of Mechatronic Systems, Leibniz University Hannover, 30823 Garbsen, Germany,
		{\tt \small aran.mohammad@imes.uni-hannover.de}}%
	}
}
\makeatletter
\newcommand{\removelatexerror}{\let\@latex@error\@gobble}
\makeatother

\newif\ifcopyright
\copyrighttrue
\begin{document}
	
	\ifcopyright
	\fboxrule=0.4pt \fboxsep=3pt
	
	\fbox{\begin{minipage}{1.1\linewidth}  
%
			This work has been submitted to the IEEE for possible publication. Copyright may be transferred without notice, after which this version may no longer be accessible.
	\end{minipage}}
	\else
	\fi
	\graphicspath{{./graphics/}}
	\maketitle
	\thispagestyle{empty}
	\pagestyle{empty}
	
	\begin{abstract}
		Fast and safe motion is crucial for the successful deployment of physically interactive robots.
		Parallel robots (PRs) offer the potential for higher speeds while maintaining the same energy limits due to their low moving masses.
		However, they require methods for contact detection and reaction while avoiding singularities and self-collisions.
		We address this issue and present \akr~— a unified approach for the detection and localization, including the distinction between collision and clamping to perform a reaction that is safe for humans and feasible for PRs.
		Our approach uses information from the encoders and motor currents to estimate forces via a generalized-momentum observer.
		Neural networks and particle filters classify and localize the contacts.
		We introduce reactions with redundancy resolution to avoid self-collisions and type-II singularities.
		Our approach detected and terminated~72 real-world collision and clamping contacts with end-effector speeds of up to~1.5\,m/s, each within~25–275\,ms.
		The forces were below the thresholds from ISO/TS~15066.
		By using built-in sensors, SafePR enables safe interaction with already assembled PRs without the need for new hardware components.
	\end{abstract}
	
	\section{Introduction}
\begin{figure}[h!]
			\vspace{1.5mm}			
			\centering
			\includegraphics[width=1\columnwidth]{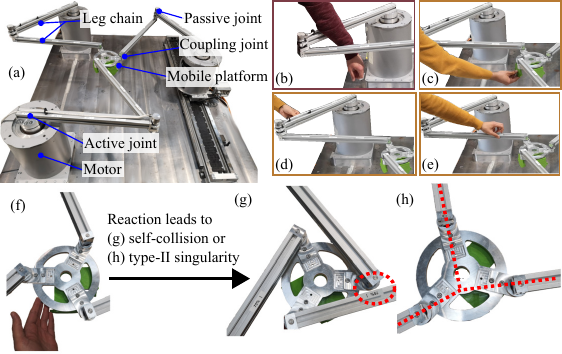}
			\caption{(a) The considered parallel robot with contact scenarios: (b) chain \textcolor{klemmungfarbe}{clamping}, \textcolor{kollisionfarbe}{collision} at the (c) platform, (d) first and (e) second link. This work addresses the scenarios that~(f) a reaction to a contact leads to (g)~self-collision or~(h) type-II singularity and, thus, an increase in the risk of injury to humans or damage to the robot.}
			\label{fig:titelbild}
			\vspace{-1.5mm}
		\end{figure}

		Safety in physical human-robot collaboration (HRC) is addressed by limiting the \emph{kinetic energy} resulting from the effective \emph{masses and relative speed} of the human and the robot in case of a collision~\cite{InternationalOrganizationforStandardization.2016}. 
		Accordingly, lightweight serial robots reduce the collision energy, but they lack high-speed and high-payload capabilities~\cite{InternationalFederationofRobotics.2024}. 
		Another approach is using PRs to reduce moving masses further and maintain the \emph{same energy limits} at \textit{higher speeds} or to \textit{decrease the kinetic energy} for the same speed. 
		PRs consist of typically base-mounted drives with kinematic chains connected to a mobile platform~\cite{Merlet.2006}.
		A PR is shown in Fig.~\ref{fig:titelbild}(a).		
		
		Whether serial or parallel --- safety must be ensured. 
		Therefore, collisions and clamping contacts, as shown in Fig.~\ref{fig:titelbild}(b)--(e), must be detected and terminated by a reaction \emph{safely and feasibly}~\cite{Haddadin.2017}.
		\subsection{Related Work: Contact Detection and Reaction for Safety}
			Detection can be realized apart from proprioceptive information by tactile skin~\cite{Dahiya.2013,Albini.2017} or by camera systems~\cite{Hoang.2022, Escarabajal.2023}. 
			In~\cite{Hoang.2022}, the contact point between the human and a PR is obtained via cameras, followed by a recursive Newton-Euler algorithm for determining the force.
			An admittance control~\cite{Escarabajal.2023} based on measured contact forces and visually determined end-effector coordinates can be performed for planned contacts, like in rehabilitation tasks.
			
			Built-in sensors are better suited to meet the requirements of \textit{fast and robust detection in dynamic contact scenarios} due to their smaller sample times and reduced time delays.
			Contact detection, isolation and identification can be performed via a physics-based disturbance observer using proprioceptive information~\cite{Haddadin.2017}. 
			Data-driven approaches enable classification into intentional and unintentional contacts~\cite{Golz.2015,Zhang.2021, Lippi.2021}.
			
			The results of detection, isolation, and identification determine \emph{reactions}. 
			For serial robots, different strategies are presented in~\cite{Luca.2006}: Contact detection via a momentum observer leads to an admittance control for performing a reflex motion.
			If the contact location on the robot structure is known, Cartesian interaction control can be used to reconfigure the robot depending on the contact force~\cite{Magrini.2014, Magrini.2015}.
			Information on the distance and the relative velocity between the human and the robot can be used to formulate an optimization problem for position tracking with the maximum permissible kinetic energy in an impact as a constraint~\cite{Meguenani.2015}.
			Energy-based control laws provide the advantage of explicitly integrating safety-related energy limits~\cite{Lachner.2021}.
			In~\cite{Vorndamme.2024}, a pipeline for the assessment of robot safety and reaction strategies, like an admittance control, zero-g or retraction movements, is introduced and evaluated with a serial-kinematic cobot and contact hazards, such as collisions and clamping.
			
			Works on \emph{parallel robots} showed that a reaction can be performed via compliance~\cite{Dutta.2019} or admittance control~\cite{Sun.2024}, assuming slow collisions so that determining acceleration is not required for force estimation.
			However, the capability to perform at high speed is necessary for the successful commercial utilization of PRs.
			In~\cite{Mohammad.2023}, different momentum-based disturbance observers detect collisions and clamping at speeds of up to~$\SI{0.9}{\meter / \second}$, and the PR reacts in the form of a zero-g mode.
			\highlightred{The contact-detection results can be improved by integrating inertial sensors mounted on the end-effector platform of a PR~\cite{Mohammad.2025}.}
			An explicitly commanded contact removal in the form of a retraction and a structure opening is presented in~\cite{Mohammad.2023_Reaction}, which uses neural networks to distinguish clamping contacts and collisions occurring at end-effector speeds of~$\SI{0.4}{\meter / \second}$.
			An extension to collision isolation and identification is introduced in~\cite{Mohammad.2023_IsolLoc}.\\
			The approaches do not yet consider singular configurations and self-collisions, as shown in Fig.~\ref{fig:titelbild}(g,h).
			These reactions would lead to uncontrollable behavior with commanded high motor torques and, thus, an increase in the risk of injury to humans or damaging the robot.
			
			We conclude that the deployment of PRs in HRC can only be realized with \textit{feasible} reaction strategies via consideration of the inherent limitations of PRs, in particular self-collisions and singularities within the workspace.
			Compliance with these limitations is crucial for the \textit{feasibility} and, hence, the success of the contact reaction.
		\begin{figure*}[t!]
			\vspace{1.5mm}
			\centering
			\includegraphics[width=\textwidth]{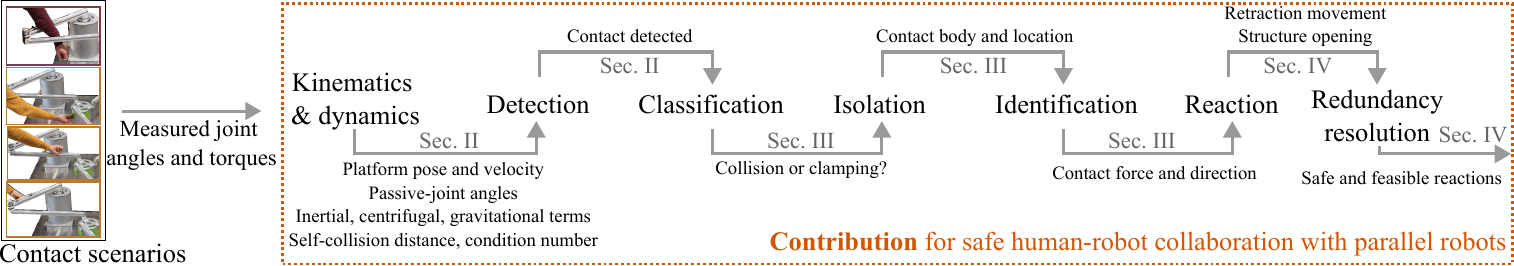} 
			\caption{
				Contact detection and reaction with \akr: 
				Contact type~(\textcolor{klemmungfarbe}{clamping}, \textcolor{kollisionfarbe}{collision}), location and forces are estimated based on built-in sensors for real-time reactions in the form of a structure opening or retraction movements while fulfilling limitations regarding self-collisions and type-II singularities.
			}
			\label{fig:gesamtbild}
			\vspace{-1.5mm}
		\end{figure*}
		\subsection{Related Work: Redundancy Resolution for Feasibility}
			In various studies with serial robots, redundancy is resolved to \emph{respect limitations}, \emph{optimize objective functions} or apply an \emph{interaction control law}.			
			In~\cite{Luca.2008}, redundant orientation degrees of freedom~(DoF) are admittance-controlled in the contact, while the end effector is still position-controlled. 
			If threshold values of the external joint torques are exceeded, the robot is fully admittance-controlled, which is extended to a two-contact scenario in~\cite{Magrini.2014}.
			
			The determinant of the inertia matrix can be selected as an objective function to minimize the robot's inertia and reduce forces during contact~\cite{Walker.1994}.
			Diagonalization of the inertia matrix is achieved in~\cite{Ficuciello.2015} so that an external force only causes an acceleration in the same dimension. 
			The inertia matrix can be projected onto a contact location, resulting in the reflected mass, which is optimized in~\cite{Mansfeld.2017,Sutjipto.2021, Khurana.2024} by a redundancy resolution to shape the inertia.
			
			Multiple target functions can be addressed by defining a \emph{task hierarchy}. 
			The motivation is that a task such as position tracking is only performed insofar as higher-prioritized tasks, such as maintaining the joint-angle limits or avoiding singularities, are ensured.
			The different hierarchy levels can be realized by quadratic optimization problems.
			The solutions are set as constraints of further low-priority optimization problems.
			In~\cite{Liu.2016}, a generalized hierarchical control is presented by defining the tasks as minimization problems and relating them to each other in hierarchies via \emph{constraints with nullspace projectors}.
			An extension to \emph{dynamic consistency} is achieved in~\cite{Dehio.2019} to change the order of the task hierarchy, to include or exclude different tasks, and to minimize the kinetic energy.
			The solution to multiple quadratic optimization problems is demonstrated in~\cite{Osorio.2020} to satisfy inequality constraints on the position, velocity and acceleration of points on the robot structure. 
			However, an optimization scheme has the disadvantage of feasibility and increased computational effort since a constrained optimization problem must be solved for each task during each time step, which can be decisive for safety in dynamic contact scenarios.
			
			Alternatively, an analytical control law with \emph{nullspace projection on velocity, acceleration or torque level} is possible~\cite{OussamaKhatib.1993, JunNakanishi.2008, AlexanderDietrich.2015, Fiore.2023}.
			These approaches rely on accurate models since imperfections weaken their theoretical advantages~\cite{Hermus.2022}.
			\emph{Operational- and joint-space limits}~\cite{Antonelli.2009b, Flacco.2012, Flacco.2012b, MunozOsorio.2019} or \emph{singularity and self-collision}~\cite{Dietrich.2011,Dietrich.2012} can then be addressed by activated inequality constraints~\cite{Moe.2016, Arrichiello.2017}.
			
			An HRC-independent optimization of performance characteristics is widely represented in the state of research on PRs.
			Redundancy of a PR can arise from actuation~(additional actuators or actuated leg chains), kinematics~(additional joints in a leg chain) and the task~(more platform DoF than required by the task, functional redundancy)~\cite{Gosselin.2018,Schappler.2023}.
			The redundancy can be resolved for \emph{avoiding singularities}~\cite{Cha.2007, Gosselin.2016, AbhishekAgarwal.2016, Schappler.2021}, \emph{maintaining joint-angle limits}~\cite{CavacantiSantos.2017} or \emph{minimizing the position error}~\cite{Kotlarski.2010}.	
			Singularity avoidance is realized in~\cite{AbhishekAgarwal.2016} via an acceleration-based nullspace projection in addition to position tracking. 
			Consideration of the joint-angle and operational-space limits is carried out in~\cite{Schappler.2023,CavacantiSantos.2017} with~(differential) dynamic programming for offline optimization of a known trajectory.
			Locally optimizing motion within a one-dimensional nullspace, as in~\cite{Schappler.2021}, is online-capable, but building weighted sums of multiple objective functions does not ensure a strict task hierarchy, which is necessary for critical scenarios like avoiding singularities while performing a collision retraction.
			
			Kinematic redundancy of PRs can further be used for singularity avoidance and gripper actuation~\cite{Wen.2021} or inferring interaction by an operator and switching to collaborative mode~\cite{Yigit.2023_IROS}.
		\subsection{Contributions}
			Redundancy resolution is a powerful method to maintain limits in contact scenarios, as shown by the related work on serial-kinematic robots. 
			For PRs, nullspace projections are only explored between end-effector and joint coordinates, but do not relate to a collided body or clamping gap.
			However, \emph{reaction strategies} require the contact coordinates in unexpected-contact cases and a combination with \emph{redundancy resolution} has not yet been investigated, which represents a \emph{research gap}.
			This article addresses this gap and introduces online-capable redundancy resolution for physical interaction with PRs.
			In particular, if the result of the contact detection requires a retraction with effectively one coordinate opposite to the collision direction, a \emph{high number of redundant DoF} results, which also holds for PRs without kinematic or actuation redundancy.
			
			The recent works~\cite{Mohammad.2023, Mohammad.2023_IsolLoc, Mohammad.2023_Reaction} provide the single-piece elements for recognizing a collision or clamping on the entire robot structure and modeling the \emph{kinetostatic relationship} between the specific contact location and the drives.
			This article merges these works into one unified approach, termed \textit{\akr}, and extends the sequence of contact detection and reaction with redundancy resolution for parallel robots, as shown in Fig.~\ref{fig:gesamtbild}.
			\akr~enables real-time capable contact-removal reactions by using only standard built-in sensors to handle the PRs' inherent limits. 
			In detail, this work's contributions are:
			\begin{enumerate}[label=\textbf{C\arabic*}]
				\item \label{contribution:Detection_Reaction}We merge the single methods from~\cite{Mohammad.2023,Mohammad.2023_IsolLoc,Mohammad.2023_Reaction} for the detection, classification, isolation and identification of collision and clamping contacts into one unified approach.
				Multiple reaction strategies based on the previously obtained information are introduced in a redundancy-resolution scheme to incorporate the collision point and clamping joint explicitly.
				\item \label{contribution:RedRes} Redundancy is resolved on velocity, acceleration and torque level while integrating type-II singularity and self-collision as inequality constraints for feasible reactions.
				\item \label{contribution:Reactions_RedRes_Results}We validate this approach for 72 real-world collision and clamping experiments with a planar PR and show the safety regarding force thresholds mentioned in ISO/TS~15066, as well as the feasibility by fulfilling limitations regarding self-collision and type-II singularities.
				\item \label{contribution:software}The software and extended documentation are published open source\footnote{\anonym{\anonymbin}{The link to the website is not mentioned due to the double-anonymous reviewing process, and it will be inserted here after acceptance.}{\url{https://aranmoha.github.io/SafePR/}}} to give other researchers the opportunity to reproduce and to transfer it to their robot systems.
			\end{enumerate}
				The new methods are required for contacts that can occur at any location and in any robot pose, thus enabling safe PRs for HRC with feasible reactions for the first time.
				Also, they are suitable for spatial and already assembled PRs since only standard built-in sensors are used.
				
				The article begins with the kinematics and dynamics modeling in Section~\ref{sec:preliminaries}, followed by the disturbance observer and the impedance control. 
				The isolation and identification of contacts are described in Sec.~\ref{sec:IsolIdent}.
				The redundancy resolution is then introduced in Sec.~\ref{sec:redundancy_resolution}.
				Section~\ref{sec:validation} commences with the test bench's description and continues with an evaluation of collision and clamping experiments. 
				\highlightred{Finally, the contributions and limitations of \akr~are summarized in Sec.~\ref{sec:Contributions_limitations}.}
				Section~\ref{sec:conlusions} concludes this article. 
	\section{Preliminaries} \label{sec:preliminaries}
		This section starts with a description of kinematics~(\ref{ssec:kinematics}) and dynamics~(\ref{ssec:dynamics}), followed by the formulation of Cartesian impedance control~(\ref{ssec:controller}) and of the generalized momentum observer~(\ref{ssec:observer}) from~\cite{Mohammad.2023}.
		\subsection{Kinematics} \label{ssec:kinematics}
\begin{figure}[b!]
				\vspace{-0.5mm}
				\centering
				\includegraphics[width=\columnwidth]{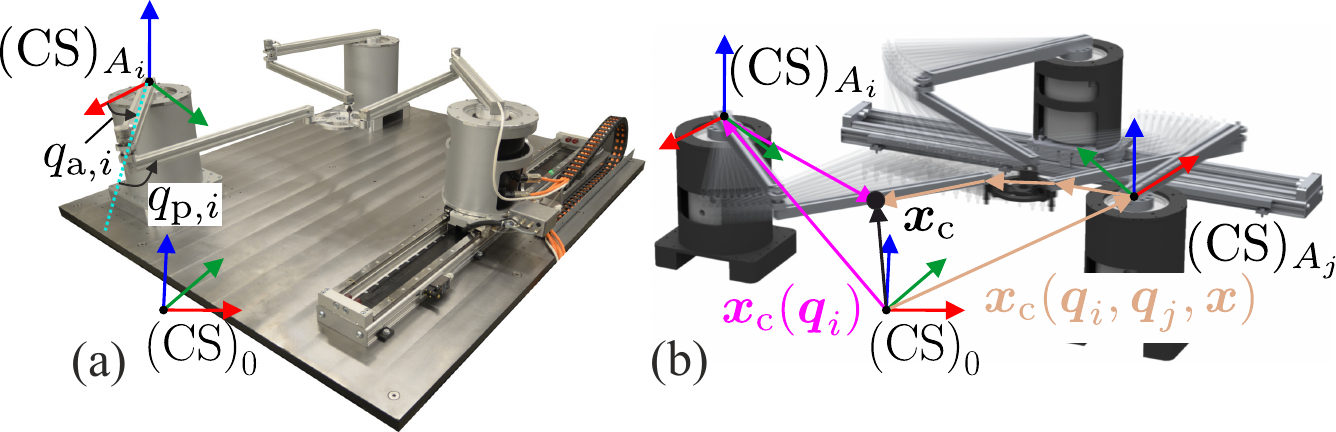}
				\caption{(a) The 3-\underline{R}RR parallel robot from~\cite{Mohammad.2023} with (b) a contact at~$\bs{x}_\mr{c}$}
				\label{fig:22_Versuchsreihe_3RRR_PKM_real_Kontaktort_red}
				\vspace{-1.5mm}
			\end{figure}

			Figure~\ref{fig:22_Versuchsreihe_3RRR_PKM_real_Kontaktort_red}(a) depicts a planar {3-\underline{R}RR} parallel robot\footnotemark\footnotetext{The letter R denotes a revolute joint, and an underlining represents an actuated joint.}~with~$n_{\bs{x}}$ platform DoF, $n_\mr{leg}$ kinematic chains and~$n_{\bs{q}_\mr{a}}$ active joints~\cite{Thanh.2012}, which is considered as fully-parallel~($n_{\bs{x}}{=}n_\mr{leg}{=}n_{\bs{q}_\mr{a}}{=}3$), since the prismatic joint discussed in~\cite{Kotlarski.2010} is not used. 
			Operational-space coordinates (end-effector/platform pose), active, passive, and coupling-joint angles are respectively given by~$\bs{x}^\mr{T}{=}[\bs{x}_{\mr{t}}^\mr{T},x_{\mr{r}}]{\in}\mathbb{R}^{n_{\bs{x}}}$, $\bs{q}_\mr{a}{\in}\mathbb{R}^{n_{\bs{q}_\mr{a}}}$, $\bs{q}_\mr{p}{\in}\mathbb{R}^{3}$ and~$\bs{q}_\mr{c}{\in}\mathbb{R}^{3}$.
			The~$n_i{=}3$ joint angles (active, passive, platform coupling) of each leg chain in~$\bs{q}_i{\in}\mathbb{R}^{n_i}$ are stacked as~$\bs{q}^\mr{T}{=}[\bs{q}_1^\mr{T}, \bs{q}_2^\mr{T}, \dots, \bs{q}_{n_\mr{leg}}^\mr{T}] {\in} \mathbb{R}^{ n_\mr{leg} n_i }$. 
		
			The kinematic constraints~$\bs{\mathit{\delta}} (\bs{q}, \bs{x}){=}\bs{0}$ are constructed by closing vector loops~\cite{Merlet.2006}. 
			The reduced kinematic constraints~$\bs{\mathit{\delta}}_\mr{red}(\bs{q}_\mr{a}, \bs{x}){=}\bs{0}$ are obtained by eliminating the passive-joint angles. 
			The active-joint angles are then analytically calculated (inverse kinematics).
			Passive-joint angles are measured and then the Newton-Raphson approach is applied to calculate the platform's pose. 
			
			For differential kinematics, the time derivative of the kinematic constraints gives
			\begin{align}\label{eq:DifKin_Jac1}
				\dot{\bs{q}}&= - \bs{\mathit{\delta}}_{\partial \bs{q}}^{-1}\bs{\mathit{\delta}}_{\partial \bs{x}} \dot{\bs{x}}=\bs{J}_{q,x}\dot{\bs{x}},\\
				\label{eq:DifKin_Jac2}
				\dot{\bs{x}}&= - \left(\bs{\mathit{\delta}}_{\mr{red},\partial \bs{x}}\right)^{-1} \left(\bs{\mathit{\delta}}_{\mr{red},\partial \bs{q}_\mr{a}}\right) \dot{\bs{q}}_\mr{a}=\bs{J}_{x,q_\mr{a}}\dot{\bs{q}}_\mr{a}
			\end{align}
			using the notation~$\bs{a}_{\partial \bs{b}}{\coloneqq} \sfrac{\partial \bs{a}}{\partial \bs{b}}$ and the Jacobian matrices\footnotemark~$\bs{J}_{q, x}{\in}\mathbb{R}^{n_{\bs{q}_\mr{a}} n_i \times n_{\bs{x}}}$ and~$\bs{J}_{x, q_\mr{a}}{\in}\mathbb{R}^{n_{\bs{x}}\times n_{\bs{q}_\mr{a}} }$.\footnotetext{For the sake of readability, dependencies on~$\bs{q}$ and~$\bs{x}$ are omitted.}
			
			The contact coordinates~$\bs{x}_\mr{c}$ of point~$\mr{C}$ on the robot structure are modeled via joint angles~$\bs{q}$~\cite{Mohammad.2023}, see Fig.~\ref{fig:22_Versuchsreihe_3RRR_PKM_real_Kontaktort_red}(b). 
			The time derivative results in~$\dot{\bs{x}}_\mr{c}{=}\bs{J}_{x_\mr{c},q}\dot{\bs{q}}$ with the Jacobian~$\bs{J}_{x_\mr{c},q}$. 
			Based on~(\ref{eq:DifKin_Jac1}) and~(\ref{eq:DifKin_Jac2}), the projection
			\begin{align}\label{eq:contactJacobian}
					\dot{\bs{x}}_\mr{c} &= \bs{J}_{x_\mr{c},q} \dot{ \bs{q}}= \bs{J}_{x_\mr{c}, x} \dot{\bs{x}} = \bs{J}_{x_\mr{c}, q_\mr{a}} \dot{\bs{q}}_\mr{a}
			\end{align}
			relates platform and joint coordinates with the contact coordinates using the contact's Jacobian matrices~$\bs{J}_{x_\mr{c}, x}$ and~$\bs{J}_{x_\mr{c}, q_\mr{a}}$.
		\subsection{Dynamics} \label{ssec:dynamics}
			The Lagrangian equations of the second kind, the \textit{subsystem} and \textit{coordinate-partitioning} methods formulate the equations of motion in the operational space without the constraint forces~\cite{Thanh.2009}.
			The dynamics model\footnote{Generalized forces~$\bs{F}{=}(\bs{f}^\mr{T}, \bs{m}^\mr{T})^\mr{T}{\in}\mathbb{R}^{n_{\bs{x}}}$ (including forces~$\bs{f}$ and moments~$\bs{m}$) are expressed in operational-space coordinates.} is
			\begin{align} \label{eq:dyn_x}
				\bs{M}_x \ddot{\bs{x}} + \bs{c}_x + \bs{g}_x +	\bs{F}_{\mr{fr},x} &= \bs{F}_\mr{a} + \bs{F}_\mr{ext,mP},		
			\end{align} 
			where~$\bs{M}_x$ denotes the inertia matrix,~$\bs{c}_x{=}\bs{C}_x\dot{\bs{x}}$ the vector of the centrifugal and Coriolis effects,~$\bs{g}_x$ the gravitational terms,~$\bs{F}_{\mr{fr},x}$ the viscous and Coulomb-friction components,~$\bs{F}_\mr{a}$ the forces resulting from the motor torques at the active joints and~$\bs{F}_\mr{ext,mP}$ external forces. 
			The formulation for the actuated-joint coordinates is given by
			\begin{align} \label{eq:dyn_qa}
				\bs{M}_{q_\mr{a}} \ddot{\bs{q}}_\mr{a} + \bs{c}_{q_\mr{a}} + \bs{g}_{q_\mr{a}} + \bs{\tau}_{\mr{fr},q_\mr{a}} &= \bs{\tau}_\mr{a} +	\bs{\tau}_\mr{a,ext}.				
			\end{align} 
			The forces~$\bs{F}_\mr{a}$ are projected by the principle of virtual work as~$\bs{\tau}_\mr{a}{=}\bs{J}_{x,q_\mr{a}}^\mr{T}\bs{F}_\mr{a}$. 
			A link contact with the external forces~$\bs{F}_\mr{ext,link}$ affects the platform and the actuated-joint coordinates via
			\begin{subequations}\label{eq:trafo_link_mP_Drives}
				\begin{align}\label{eq:trafo_link_mP_Drives1}
					\bs{F}_\mr{ext,mP}&=\bs{J}_{x_\mr{c},x}^\mr{T} \bs{F}_\mr{ext,link} \quad \text{and}\\
					\bs{\tau}_\mr{a,ext}&=\bs{J}_{x_\mr{c},q_\mr{a}}^\mr{T} \bs{F}_\mr{ext,link}
				\end{align} 
			\end{subequations}
			in a configuration-dependent way.
		\subsection{Cartesian Impedance Control in Operational Space} \label{ssec:controller}
			Cartesian impedance control for PRs~\cite{Taghirad.2013} is described by
			\begin{equation} \label{eq:ImpRegX}
				\bs{F}_\mr{a} = \hat{\bs{c}}_x + \hat{\bs{g}}_x + \hat{\bs{F}}_{\mr{fr},x} + \hat{\bs{M}}_x \ddot{\bs{x}}_\mr{d} + \bs{K}_\mr{d} \bs{e}_x + \bs{D}_\mr{d} \dot{\bs{e}}_x 
			\end{equation}
			in platform coordinates with the compensation of the dynamics,~${\bs{K}_\mr{d}{=}\mr{diag}(k_{\mr{d},1}, \dots ,k_{\mr{d},n_{\bs{x}}}){>}\bs{0}}$ as the desired stiffness matrix and the pose error~$\bs{e}_x$ between the desired~$\bs{x}_\mr{d}$ and actual pose~$\bs{x}$.
			The position and orientation error are calculated via~$\bs{e}_{x,\mr{t}}{=}\bs{x}_\mr{t,d}{-}\bs{x}_\mr{t}$ and
			\begin{equation}\label{eq:orientationerror}
				\bs{e}_{x,\mr{r}}=\tilde{\bs{\varphi}}({^0}\bs{R}_\mr{E}^\mr{T}(\bs{x}_\mr{r}) {^0}\bs{R}_\mr{E}(\bs{x}_\mr{r,d}))
			\end{equation}
			with Cardan angles~(Tait-Bryan angles)~$\tilde{\bs{\varphi}}$~\cite{Schappler.2021}.
			The factorization damping design~\cite{AlbuSchaffer.2003} is described via
			\begin{align} 
				\bs{D}_\mr{d} = \tilde{\bs{M}}_{x} \bs{D}_{\xi,\mr{ic}} \tilde{\bs{K}}_\mr{d} + \tilde{\bs{K}}_\mr{d} \bs{D}_{\xi,\mr{ic}} \tilde{\bs{M}}_{x}
			\end{align} 
			with~$\bs{K}_\mr{d}{=}\tilde{\bs{K}}_\mr{d} \tilde{\bs{K}}_\mr{d}$, $\bs{M}_x{=}\tilde{\bs{M}}_{x} \tilde{\bs{M}}_{x}$ (due to the symmetric positive-definite~$\bs{M}_x$), ${\bs{D}_{\xi,\mr{ic}}{=}\mr{diag}(D_{\xi,\mr{ic},1}, \dots ,D_{\xi,\mr{ic},n_{\bs{x}}})}$ and~$D_{\xi,\mr{ic},i}{>}0$. 
			The ideal closed-loop error dynamics results in
			\begin{align}\label{eq:ic_closedDynamics}
				\bs{M}_x \ddot{\bs{e}}_x + \bs{D}_\mr{d} \dot{\bs{e}}_x + \bs{K}_\mr{d} \bs{e}_x = -\bs{F}_\mr{ext,mP}
			\end{align}
			with the external force as input and the pose error as output.
		\subsection{Generalized-Momentum Observer} \label{ssec:observer}
			Following~\cite{Haddadin.2017}, the generalized momentum~$\bs{p}_x{=}\bs{M}_x \dot{\bs{x}}$ is used to set up a residual in the operational space since the minimal coordinates for the dynamics of PRs are expressed in~$\bs{x}$. 
			The residual's time derivative is~$\sfrac{\mr{d}}{\mr{dt}} \hat{\bs{F}}_\mr{ext,mP} {=} \bs{K}_{\mr{o}} (\dot{\bs{p}}_x {-} \dot{\hat{\bs{p}}}_x )$ with the observer gain matrix~$\bs{K}_\mr{o}{=}\mr{diag}(k_{\mr{o},1}, \dots ,k_{\mr{o},n_{\bs{x}}})$ and~$k_{\mr{o},i}{>}0$~(unit $1/\mr{s}$). 
			The generalized-momentum observer~(MO) is constructed by expressing~(\ref{eq:dyn_x}) as~$\hat{\bs{M}}_x\ddot{\bs{x}}$ and substituting it with the term~$\sfrac{\mr{d}}{\mr{dt}}\hat{\bs{p}}_x$ in the time integral of~$\sfrac{\mr{d}}{\mr{dt}} \hat{\bs{F}}_\mr{ext,mP}$. 
			With~$\sfrac{\mr{d}}{\mr{dt}}\hat{\bs{M}}_x{=}\hat{\bs{C}}_x^\mr{T}{+}\hat{\bs{C}}_x$~\cite{Haddadin.2017, Ott.2008}, the external-force estimation is realized by
			\begin{align} \label{eq:mo}
				\hat{\bs{F}}_\mr{ext,mP} &{=} \bs{K}_\mr{o} \left( \hat{\bs{M}}_x \dot{\bs{x}} - \int_{0}^t (\bs{F}_\mr{a} {-} \hat{\bs{\beta}} {+} \hat{\bs{F}}_\mr{ext,mP}) \mr{d}\tilde{t} \right)  \\ 
				\label{eq:mo_beta}
				\text{and } \hat{\bs{\beta}} {=} \hat{\bs{g}}_x &{+} \hat{\bs{F}}_{\mr{fr},x} {+}( \hat{\bs{C}}_x {-}\dot{\hat{\bs{M}}}_x )\dot{\bs{x}} {=} \hat{\bs{g}}_x {+} \hat{\bs{F}}_{\mr{fr},x}{-}\hat{\bs{C}}_x^\mr{T} \dot{\bs{x}}.
			\end{align}
			By well-identified dynamics terms in (\ref{eq:mo_beta}), the MO estimates the external force in platform coordinates with the linear and decoupled error dynamics
			\begin{align}
				\bs{K}_\mr{o}^{-1} \dot{\hat{\bs{F}}}_\mr{ext,mP} + \hat{\bs{F}}_\mr{ext,mP}=\bs{F}_\mr{ext,mP},
			\end{align} 
			which enables online contact detection.
	\section{Isolation and Identification} \label{sec:IsolIdent}
		The theory of this section's methods is reproduced from~\cite{Mohammad.2023_IsolLoc,Mohammad.2023_Reaction} for completeness.
		Collision effects on the mobile platform~(\ref{ssec:Coll_mP}) and at the kinematic chain's links~(\ref{ssec:Coll_Link}) are presented at the beginning. 
		The collided-body classification algorithm with a feedforward neural network~(\ref{ssec:ClassAlg_FNN}) is described afterward.
		Then, a particle filter is introduced (\ref{ssec:IsolItend_PF}).
		Finally, the clamped-chain classification~(\ref{ssec:ContactClass}) is shown.
		\subsection{Collision on the Mobile Platform} \label{ssec:Coll_mP} 
			A contact wrench~$\bs{F}_\mr{ext,mP}$ with forces~$\bs{f}_\mr{ext,mP}$ and moments~$\bs{m}_\mr{ext,mP}$ is expressed in operational-space coordinates and affects the mobile platform.
			$\hat{\bs{F}}_\mr{ext,mP}$ and~$\hat{\bs{\tau}}_{\mr{a,ext}}{=}\bs{J}_{x,q_\mr{a}}^\mr{T}\hat{\bs{F}}_\mr{ext,mP}$ are then estimated by the MO from~(\ref{eq:mo}). 
			\begin{figure}[t!]
				\vspace{1.5mm}
				\centering
				\includegraphics[width=\columnwidth]{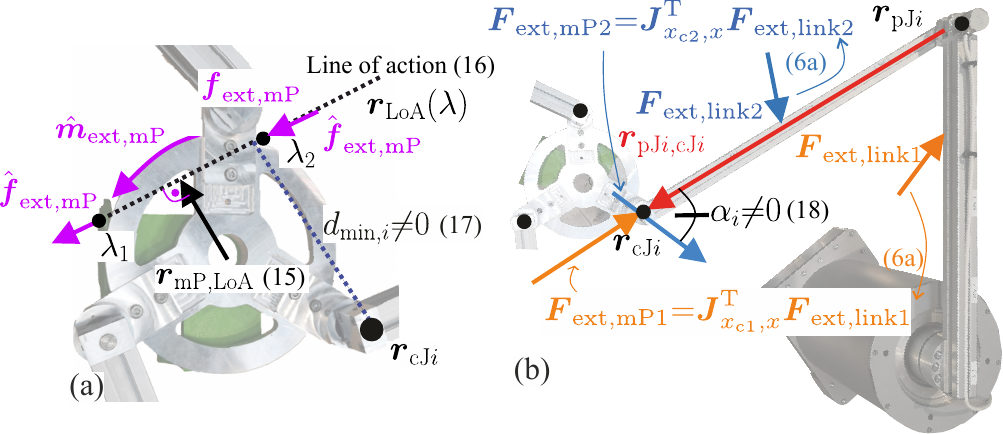}
				\caption{Kinetostatic analysis from~\cite{Mohammad.2023_IsolLoc}. (a) External force~$\bs{f}_{\mr{ext,mP}}$ on the platform with the estimate~$\hat{\bs{F}}_\mr{ext,mP}{=}(\hat{\bs{f}}_{\mr{ext,mP}}^\mr{T}, \hat{\bs{m}}_{\mr{ext,mP}}^\mr{T})^\mr{T}$, the minimum lever~$\bs{r}_\mr{mP,LoA}$, the line of action~$\bs{r}_\mr{LoA}(\lambda)$ and intersection points at~$\lambda_1,\lambda_2$. 
					The minimum distance~$d_{\min,i}$ is between~$\bs{r}_\mr{LoA}(\lambda)$ and the coupling point~$\bs{r}_{\mr{cJ}i}$. 
					(b) Link forces~$\bs{F}_{\mr{ext,link1/2}}$ with their projections~$\bs{F}_{\mr{ext,mP1/2}}$ on the platform coordinates. $\bs{F}_\mr{ext,mP2}$ and~$\bs{r}_{\mr{pJ}i,\mr{cJ}i}$ include the angle~$\alpha_i$.}
				\label{fig:KinetostaticAnalysis}
				\vspace{-1.5mm}
			\end{figure}
			Figure~\ref{fig:KinetostaticAnalysis}(a) depicts the following procedure. 
			An assumption for unwanted contacts is made by~$\bs{m}_\mr{ext,mP}{=}\bs{0}$~\cite{Haddadin.2017}, resulting in
			\begin{equation} 
				\begin{split}
					\hat{\bs{m}}_\mr{ext,mP} {=} \bs{r} {\times} \hat{\bs{f}}_\mr{ext,mP} {=} \bs{S} 	(\bs{r})\hat{\bs{f}}_\mr{ext,mP} {=} \bs{S}^\mr{T} ( \hat{\bs{f}}_\mr{ext,mP}) \bs{r}
				\end{split}
			\end{equation}
			with~$\bs{S}$ as a skew-symmetric-matrix operator and~$\bs{r}$ as a lever between the platform's body-fixed coordinate system to any point on the line of action~(LoA)~$\bs{r}_\mr{LoA}(\lambda)$. 
			Using the Moore-Penrose inverse~$(\dagger)$ of~$\bs{S}^\mr{T}(\hat{\bs{f}}_\mr{ext,mP})$~\cite{Haddadin.2017}, the minimum-distance vector 
			\begin{equation} \label{eq:CalcMPLoA}
				\begin{split} 
					\bs{r}_\mr{mP,LoA} &= ( \bs{S}^\mr{T} ( \hat{\bs{f}}_\mr{ext,mP}) )^\dagger \hat{\bs{m}}_\mr{ext,mP}
				\end{split}
			\end{equation} 
			from the platform's coordinate system's origin to~$\bs{r}_\mr{LoA}(\lambda)$ of the external force is calculated. 
			Now the LoA
			\vspace{-2mm}
			\begin{equation} \label{eq:CalcLoA}
				\bs{r}_\mr{LoA}(\lambda){=}\bs{x}_\mr{t}{+}\bs{r}_\mr{mP, LoA}{+}\lambda \hat{\bs{n}}_{f_\mr{ext,mP}} \text{ with } \hat{\bs{n}}_{f_\mr{ext,mP}}{=}\frac{\hat{\bs{f}}_\mr{ext,mP}}{||\hat{\bs{f}}_\mr{ext,mP} ||_2}
			\end{equation} 
			can be determined with the scalar variable~$\lambda$, leading to the intersections~$\lambda_1$ and $\lambda_2$ with the known platform hull. 
			These two cases correspond to a pull~($\lambda_1$) and a push~($\lambda_2$) force. 
			It is assumed that undesired contacts are the latter.
			\highlightred{The scalar variables $\lambda_1$ and $\lambda_2$ result from the determination of the intersection points of the outer shell of the mobile end-effector platform with the line of action $\bs{r}_\mr{LoA}(\lambda)$ of the estimated external forces (see Fig.~\ref{fig:KinetostaticAnalysis}(a)).
			The end-effector platform is represented by simple geometries, allowing for quick calculation of the intersection.
			For the planar parallel robot, which can only sense and act forces in the plane and moments around the vertical axis of the mobile end-effector platform, a circle is assumed as a simplifying geometry.
			The parameters of the circular equation, i.e. center and radius, are known at all times from the geometric data and the kinematics.
			The linear equation~$\bs{r}_\mr{LoA}(\lambda)$ is already known from the modeling, and its entries $r_{\mr{LoA},x}$ and $r_{\mr{LoA},y}$ can be inserted into the circular equation in order to obtain the possible solutions $\lambda_1$ and $\lambda_2$ by rearranging.
			Substituting these solutions into the line-of-action equation gives possible positions $\bs{r}_\mr{LoA}(\lambda_1)$ and $\bs{r}_\mr{LoA}(\lambda_2)$.}
			Thus, a collision on the platform is localized, and together with the MO's estimation in~(\ref{eq:mo}), the isolation and identification of platform collisions are complete.
		\subsection{Collision at a Leg Chain} \label{ssec:Coll_Link}
			Figure~\ref{fig:KinetostaticAnalysis}(b) shows the two link forces~$\bs{F}_\mr{ext,link1/2}$ at the $i$-th chain's first and second link, and their projections~$\bs{F}_\mr{ext,mP1/2}$ to the platform coordinates via the corresponding contact Jacobian matrix~$\bs{J}_{x_{\mr{c1/2}},x}$ from~(\ref{eq:trafo_link_mP_Drives1}).		
			The two link collisions in Fig.~\ref{fig:KinetostaticAnalysis}(b) differ from the platform contact in Fig.~\ref{fig:KinetostaticAnalysis}(a) since the minimum distance 
			\begin{equation} \label{eq:MinDistd}
				\begin{split} 
					d_{\mr{min},i} = ||\left(\bs{r}_{\mr{cJ}i}-(\bs{x}_\mr{t}{+}\bs{r}_\mr{mP, LoA1/2})\right) \times \hat{\bs{n}}_{f_\mr{ext,mP1/2}}||_2
				\end{split}
			\end{equation} 
			from~$\bs{r}_\mr{LoA1/2}(\lambda)$ to the~$i$-th coupling joint~$\bs{r}_{\mr{cJ}i}$ is zero.
			This enables the determination of the affected leg chain on which the contact force acts. 
			Since the link's contact force affects the platform via the~$i$-th kinematic chain, the force's projection in platform coordinates intersects with the coupling joint~$\mr{cJ}i$.
			
			The distinction between a collision on the first and on the second link is possible as follows. 
			The vector~$\bs{r}_{\mr{pJ}i,\mr{cJ}i}$ from the passive joint~($\mr{pJ}i$) to the coupling joint~($\mr{cJ}i$) of the~$i$-th chain and~$\hat{\bs{n}}_{f_\mr{ext,mP}}$ define the angle
			\begin{equation} \label{eq:Anglealpha}
				\begin{split} 
					\alpha_i{=}\angle (\hat{\bs{n}}_{f_\mr{ext,mP}}, \bs{r}_{\mr{pJ}i,\mr{cJ}i}).
				\end{split}
			\end{equation} 
			The lines of action~$\bs{r}_{\mr{LoA1/2}}$ of~$\bs{F}_{\mr{ext,mP1/2}}$ show in comparison the difference that~$\bs{r}_{\mr{LoA,link}1}$ with~$\alpha_{\mr{link}1}{=\SI{180}{\degree}}$ is antiparallel~(or parallel by~$\SI{0}{\degree}$) to~$\bs{r}_{\mr{pJ}i,\mr{cJ}i}$. 
			In contrast,~$\bs{F}_\mr{ext,mP2}$ with~$\bs{r}_{\mr{pJ}i,\mr{cJ}i}$ includes the angle~$\SI{0}{\degree}{<}|\alpha_{\mr{link}2}|{<}\SI{180}{\degree}$. 
			Furthermore, the distinction of links can be made based on~$\hat{\bs{\tau}}_{\mr{a,ext}}$, since~$\bs{F}_\mr{ext,link1}$ only acts on the affected actuated joint.\footnote{Isolation for a collision at the first link is only realized up to the body classification since only one drive is excited, and no position can be determined.}
			
			\begin{figure}[tb!]
				\removelatexerror
				\vspace{1.5mm}
				\begin{algorithm}[H]
					\caption{Calculate features $\bs{d}, \bs{\alpha}$ and $n_{\tau_\mr{a}} $}\label{alg:classifyContBody_geomFeat}
					{\small 
						\SetKwInOut{Input}{Input}
						\SetKwInOut{Output}{Output}
						\Input{$\hat{\bs{F}}_\mr{ext,mP},\hat{\bs{\tau}}_\mr{a,ext}, \bs{q},\bs{x}, \varepsilon_{\hat{\tau}_{\mr{a}}}$}
						\Output{Collision-body-relevant features $\bs{d}, \bs{\alpha}, n_{\tau_\mr{a}}$}
						$\bs{r}_\mr{mP,LoA}\gets$ Minimal lever by (\ref{eq:CalcMPLoA})\;					
						$\bs{r}_\mr{LoA}(\lambda)\gets$ Line of action by (\ref{eq:CalcLoA})\;	
						$n_{\tau_\mr{a}} \gets 0$ \tcp*{number of affected drives}\
						$\bs{d} {\gets}\bs{0}$ \tcp*{minimal distances for $n_\mr{leg}$ chains}\
						$\bs{\alpha} {\gets} \bs{0}$ \tcp*{angles for $n_\mr{leg}$ chains}\
						\For{$i:=1$ to $n_{\mr{leg}}$} 
						{\tcp{calculate features}
							$\bs{r}_{\mr{cJ}i}\gets$ $i$-th coupling-joint position by serial forward kinematics\;		
							$d_i \gets$ Minimal distance $d_{\min,i}$ by (\ref{eq:MinDistd}) in row $i$ of $\bs{d}$\;
							$\bs{r}_{\mr{pJ}i,\mr{cJ}i}\gets$ Vector from passive joint to coupling joint\;	
							$\alpha_i\gets$ Angle $\alpha_i$ by (\ref{eq:Anglealpha}) in row $i$ of $\bs{\alpha}$\;
							\If{$|\hat{\tau}_{\mr{a}_i\mr{,ext}}|{>}\varepsilon_{\hat{\tau}_{\mr{a}}}$}
							{
								{$n_{\tau_\mr{a}}\gets n_{\tau_\mr{a}}{+}1$\;}
							}
						}
					}
				\end{algorithm}
				\vspace{-0.5mm}
			\end{figure}
			By these considerations and by generalizing from the particular case of Fig.~\ref{fig:KinetostaticAnalysis}, we hypothesize that~$\hat{\bs{\tau}}_\mr{a,ext}$, $\bs{d}^\mr{T}{=}[d_{\min,1}, \dots, d_{\min,{n_\mr{leg}}}]$ and $ \bs{\alpha}^\mr{T}{=}[\alpha_1, \dots, \alpha_{n_\mr{leg}}]$ can be used to classify the robot's collided body: The robot configuration is captured in the features~$\bs{d}, \bs{\alpha}$ and~$\hat{\bs{\tau}}_\mr{a,ext}$, favoring generalization for contacts in new joint-angle configurations.
			This \emph{reduces the necessity for an extensive sampling} of the high-dimensional configuration space. 
			Algorithm~\ref{alg:classifyContBody_geomFeat} summarizes the calculation of~$\bs{d},\bs{\alpha}$ and the number~$n_{\tau_\mr{a}}$ of affected drives for collision-body classification.
			The inputs~$\bs{q}$ and~$\bs{x}$ to Alg. \ref{alg:classifyContBody_geomFeat} follow from the kinematics modeling, while~$\hat{\bs{F}}_\mr{ext,mP}$ and~$\hat{\bs{\tau}}_{\mr{a,ext}}$ are obtained by~(\ref{eq:mo}). A threshold~$\varepsilon_{\hat{\tau}_{\mr{a}}}$ is predefined for determining~$n_{\tau_\mr{a}}$ and avoiding false-positive classification.
		\subsection{Collision-Body Classification with a Neural Network} \label{ssec:ClassAlg_FNN}
			A feedforward neural network (FNN) is selected as a classification algorithm for the collided body. 
			The optimization method Adam is chosen to train the classifier with the physically modeled inputs~$\hat{\bs{F}}_\mr{ext,mP}$, $\hat{\bs{\tau}}_\mr{a,ext}$, $\bs{d}$, $\bs{\alpha}$ and the collided body as output.
			This dataset is obtained from previous collision experiments with the known collided bodies consisting of six links and the mobile platform. 
			The hyperbolic tangent function is selected as the activation function in the hidden layers. 
			Since the inputs are available in robot operation, real-time prediction is possible.
			A regularization term, the number of hidden layers and neurons are determined by a grid search in a hyperparameter optimization to avoid underfitting and overfitting. 
		\subsection{Particle Filter for the Second Links} \label{ssec:IsolItend_PF} 
			If the FNN predicts a second link as the collided body, a particle filter with~$R$ particles is initiated for that second link. 
			The~$r$-th particle at the~$k$-th time step is represented by
			\begin{align}
				\bs{p}_k^{[r]} = [\hat{l}_\mr{c}^{[r]},\hat{f}_{y}^{[r]}]^\mr{T}
			\end{align}
			with the estimated contact force~$\hat{f}_y^{[r]}$ at the second link.
			Two assumptions are made to favor real-time capability. 
			The first is that a contact force affects orthogonally on the second link.
			Secondly, the links are assumed as thin rods, and a variable~$0 {\le} \hat{l}_\mr{c}^{[r]} {\le} 1$ normalized to the link length is introduced, which is expressed and invariant in the body-fixed coordinate system of the second link.
			At the passive joint~$\mr{pJ}i$, $\hat{l}_\mr{c}^{[r]}{=}0$ holds and increases along the second link to the coupling joint~$\mr{cJ}i$.
			This allows one-dimensional collision isolation and identification for a planar\footnotemark~PR, which is supported by the usually high ratio of link length to link radius.
			\footnotetext{Two-dimensional identification for a spatial PR}
			\highlightred{The particle filter can be adapted to non-orthogonal collisions via $\bs{p}_k^{[r]} {=} [l_\mr{c}^{[r]},\hat{f}_{x}^{[r]},\hat{f}_{y}^{[r]}]^\mr{T}$, where $\hat{f}_{x}^{[r]}$ and $\hat{f}_{y}^{[r]}$ are force components for angled contact forces.}
			Each particle position is updated in the motion model 
			\begin{align} \label{eq:pf_motionmodel}
				\bs{p}_{k+1}^{[r]} \sim \mathcal{N}(\bs{p}_{k}^{[r]}, \bs{\Sigma}_\mr{mot})
			\end{align}
			by sampling a normal distribution with a covariance matrix~$\bs{\Sigma}_\mr{mot}$.
			The measurement model with the weights 
			\begin{align} \label{eq:pf_measurementmodel}
				w_k^{[r]}{=}\exp \left(
				{-}\frac{1}{2} 
				( \hat{\bs{\tau}}_{\mr{a,ext}} {-} \hat{\bs{\tau}}_{\mr{a,ext}}^{[r]})^\mr{T}
				\bs{\Sigma}_\mr{meas}^{-1} 
				( \hat{\bs{\tau}}_{\mr{a,ext}}{-}\hat{\bs{\tau}}_{\mr{a,ext}}^{[r]} )
				\right)
			\end{align}
			includes a covariance matrix~$\bs{\Sigma}_\mr{meas}$ and the estimated external joint torques~$\hat{\bs{\tau}}_\mr{a,ext}$.
			In (\ref{eq:pf_measurementmodel}),~$\hat{\bs{\tau}}_{\mr{a,ext}}^{[r]}$ represents the projection 
			\begin{align}
				\hat{\bs{\tau}}_\mr{a,ext}^{[r]} = (\bs{J}_{\bs{x}_\mr{c,t},q_\mr{a}}^{[r]})^\mr{T} \hat{\bs{f}}_\mr{c}^{[r]}
			\end{align}
			by (\ref{eq:trafo_link_mP_Drives}) of the estimation~$\hat{\bs{f}}_\mr{c}^{[r]}{=}^0\bs{R}_{\mr{pJ}i} [0,\hat{f}_{y}^{[r]},0]^\mr{T}$ expressed in~$\left(\mr{CS}\right)_{\mr{pJ}i}$ of the $i$-th chain's second link.
			Since the contact point~$\bs{x}_\mr{c,t}$ depends on~$\hat{l}_\mr{c}^{[r]}$, the contact's Jacobian matrix~$\bs{J}_{\bs{x}_\mr{c,t},q_\mr{a}}^{[r]}$ relates the $r$-th particle's contact point onto the actuated-joint coordinates.
			Thus, the particle positions are weighted with~$w_k^{[r]}$ according to their fit to~$\hat{\bs{\tau}}_\mr{a,ext}$ and an importance resampling is performed according to the weights.
		\subsection{Clamping-Chain Classification with Neural Networks} \label{ssec:ContactClass}
			Two FNNs are trained for clamping classification with the same hyperparameter optimization scheme as in~(\ref{ssec:ClassAlg_FNN}). 
			\subsubsection{Clamping Classification}\label{ssec:ContactClass_I}
				The first FNN classifies a contact into~$\{$Collision, Clamping$\}$ based on the input~$\hat{\bs{F}}_\mr{ext,mP}$ from~(\ref{eq:mo}).
			\subsubsection{Chain Classification}\label{ssec:ContactClass_II}		
				If a clamping contact is predicted, a second FNN categorizes the contact into the~$n_\mr{leg}$ classes~$\{\mr{C}_1, \dots, \mr{C}_{n_\mr{leg}} \}$ using the estimated external forces~$\hat{\bs{F}}_\mr{ext,mP}$ and the minimum distances~$\bs{d}$ from (\ref{eq:MinDistd}).
				The~$i$-th class represents a clamping between the two links of the~$i$-th kinematic chain.
	\section{Redundancy Resolution} \label{sec:redundancy_resolution}
		This section introduces the theory of the contributions~\ref{contribution:Detection_Reaction} and~\ref{contribution:RedRes}.
		The reaction method~(\ref{ssec:Retraction}) from~\cite{Mohammad.2023_Reaction} and its limitations~(\ref{ssec:problem_solution_formulation}) are discussed to motivate the redundancy resolution with collision and clamping-reaction strategies~(\ref{ssec:CollClam_Reactions}) on kinematics~(\ref{ssec:redres_kin}) and dynamics level~(\ref{ssec:redres_dyn}) following~\cite{Khatib.1987,JunNakanishi.2008,AlexanderDietrich.2015}.
		The integration of inequality constraints into the redundancy resolution follows the approach in~\cite{Moe.2016}, and its adaptation to the different nullspace projections for PRs in HRC with continuous control laws~(\ref{ssec:IEC_Stetigkeit}) is then presented.
		
		\subsection{Simple Reaction Strategy: Retraction Movement} \label{ssec:Retraction}
\begin{figure}[t!]
				\vspace{1.5mm} 
				\centering
				\includegraphics[width=0.8\columnwidth]{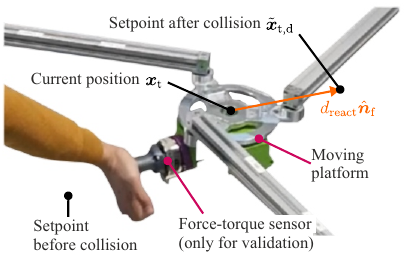}
				\caption{Retraction movement from~\cite{Mohammad.2023_Reaction}}
				\label{fig:reaktion_rueckz_strukoeff}
				\vspace{-1.5mm}
			\end{figure}

			The direction~$\hat{\bs{n}}_{f_\mr{ext,mP}}$ of the external force's LoA from~(\ref{eq:CalcLoA}) allows the calculation of a new target position 
			\begin{align} \label{eq:retr_tr}
				\tilde{\bs{x}}_\mr{t,d}=\bs{x}_\mr{t}+d_\mr{react} \hat{\bs{n}}_{f_\mr{ext,mP}}
			\end{align}
			with a predefined distance~$d_\mr{react}$ and the current platform position~$\bs{x}_\mr{t}$.
			A retraction-trajectory planning is initiated to the new target pose~$\tilde{\bs{x}}_\mr{d}$ without requiring any information on the contact location or type.
			Since the control law of the PR is formulated in the operational space and the closed kinematic loops are captured in the kinetostatic projection~$\bs{\tau}_\mr{a}{=}\bs{J}_{x,q_\mr{a}}^\mr{T}\bs{F}_\mr{a}$, the retraction movement occurs with all~$n_\mr{leg}$ kinematic chains, as shown in Fig.~\ref{fig:reaktion_rueckz_strukoeff}.	
			This reaction strategy can be extended by a reduction of the translational stiffness~$\bs{K}'_\mr{t,d}{<}\bs{K}_\mr{t,d}$ of the controller up to~$\bs{K}'_\mr{t,d}{=}\bs{0}$ (zero-g mode).	
		\subsection{Problem and Solution Formulation}\label{ssec:problem_solution_formulation}
			Although the described reaction does not require any information on the contact location, it has two disadvantages: (i)~It requires the retraction movement to be \textit{feasible} and is not restricted by singularities or self-collisions. 
			If these conditions are not met, the robot may be destroyed, which poses an increased risk to humans.
			(ii) Independent of singularities, utilizing a desired \textit{interaction control} law at the contact location is impossible since the desired stiffness and damping matrices are expressed in the platform coordinates in~(\ref{eq:ImpRegX}). 
			Thus, a desired parameterization of the robot-environment dynamics is only possible in these coordinates, while the compliance at the contact location is biased according to the contact's Jacobian matrix.
			These two limitations are addressed in the following by extending the retraction movement to include a redundancy resolution involving the contact's Jacobian matrix.
			We hypothesize that the reaction only has to occur opposite to the contact direction and effectively corresponds to a one-dimensional task for collisions and a two-dimensional task for clamping contacts.
			
			Considered are~$n_\mathcal{T}$ tasks with their coordinates in~$\bs{\sigma}_i{=}\bs{f}(\bs{q}_\mr{a},\bs{x}){\in}\mathbb{R}^{n_{\bs{\sigma}_i}}$ and~$n_{\bs{\sigma}_i}{<}n_{\bs{q}_\mr{a}}$.
			The time derivatives are given by
			\begin{align}
				\label{eq:DifKinSigma_d}
				\dot{\bs{\sigma}}_i &= \bs{J}_i \dot{\bs{q}}_\mr{a} \quad \text{and}\\
				\label{eq:DifKinSigma_dd}
				\ddot{\bs{\sigma}}_i &= \dot{\bs{J}}_i \dot{\bs{q}}_\mr{a} +\bs{J}_i \ddot{\bs{q}}_\mr{a}
			\end{align}
			with the Jacobian matrices~$\bs{J}_i {\in}\mathbb{R}^{n_{\bs{\sigma}_i}{\times}n_{\bs{q}_\mr{a}}}$.
			The degree of redundancy is defined as~$n_{\bs{q}_\mr{a}}{-}n_{\bs{\sigma}}$ with~$n_{\bs{\sigma}}{=}\sum_{i=1}^{n_\mathcal{T}}n_{\bs{\sigma}_i}{\le}n_{\bs{q}_\mr{a}}$.
		\subsection{Collision and Clamping Reaction}\label{ssec:CollClam_Reactions}
			The generic representation in~(\ref{eq:DifKinSigma_d}) and~(\ref{eq:DifKinSigma_dd}) allows the following usage of the localization and identification results from Sec.~\ref{sec:IsolIdent} for reaction tasks.
			\subsubsection{Reaction for a Collision on Mobile Platform}
				If the mobile platform is classified as a collision body, the robot retracts parallel to the direction~$\hat{\bs{n}}_{f_\mr{ext,mP}}$ via
					\begin{align}\label{eq:react_Coll_mobPl}
						\dot{\bs{\sigma}}_i&=\dot{\bs{x}}_\mr{t} \quad \text{with} \quad
						\bs{J}_i=\bs{J}_{x_\mr{t},q_\mr{a}}.
					\end{align} 
				Here, the platform orientation is redundant for the reaction.
				The translational DoF perpendicular to~$\hat{\bs{n}}_{f_\mr{ext,mP}}$ can also be redundant by a one-dimensional formulation of the retraction along the line of action via
				\begin{align}\label{eq:react_Coll_mobPl_1D}
					\dot{\sigma}_i&{=} \hat{\bs{n}}_{f_\mr{ext,mP}}^\mr{T} \dot{\bs{x}}_\mr{t}{=}\dot{x}_\mr{LoA,t} \text{ with }
					\bs{J}_i=\hat{\bs{n}}_{f_\mr{ext,mP}}^\mr{T} \bs{J}_{x_\mr{t},q_\mr{a}}.
				\end{align} 
				This projects the translational platform velocity onto the line of action and is equivalent to a rotation with the matrix~${^\mr{LoA}}\bs{R}_\mr{0}$ between~$(\mr{CS})_0$ and a new coordinate system~$(\mr{CS})_\mr{LoA}$ to capture the retraction along~$\hat{\bs{n}}_{f_\mr{ext,mP}}$ in only one axis.
			\subsubsection{Reaction for a Collision on a First Link}
				If the FNN from Sec.~\ref{ssec:ClassAlg_FNN} predicts a first link of the~$j$-th kinematic chain as a collision body, the reaction is performed via
					\begin{align}\label{eq:react_Coll_1Link}
						\dot{\sigma}_i&=\dot{q}_{\mr{a},j}  \quad \text{with} \quad
						\bs{J}_i=\bs{j}_i
					\end{align} 
				and~$\bs{j}_i^\mr{T}{\in}\mathbb{R}^{n_{\bs{q}_\mr{a}}}$ having zeros in the entries, except a one at the~$j$-th position. 
				The direction of the reaction is calculated from the sign of the~$j$-th entry of~$\hat{\bs{\tau}}_\mr{a,ext}$.
				Since only the~$j$-th actuator is required for this contact reaction, there are~$n_{\bs{q}_\mr{a}}{-}1$ redundant DoF.
			\subsubsection{Reaction for a Collision on a Second Link}
				This reaction method is based on the result of the particle filter from Sec.~\ref{ssec:IsolItend_PF}. 
				Using the estimated contact location~$\hat{l}_\mr{c}$ and contact force~$\hat{\bs{f}}_\mr{c}$, the task
					\begin{align}\label{eq:react_Coll_2Link}
						\dot{\bs{\sigma}}_i&=\dot{\bs{x}}_{\mr{c,t}}(\hat{l}_\mr{c}) \quad \text{with} \quad
						\bs{J}_i =\bs{J}_{x_\mr{c,t},q_\mr{a}}
					\end{align} 
				is formulated to retract from the contact location~$\mr{C}$ in the direction of~$\hat{\bs{f}}_\mr{c}$.
				Similarly to~(\ref{eq:react_Coll_mobPl_1D}), the DoF perpendicular to~$\hat{\bs{f}}_\mr{c}$ can be declared redundant by the task \begin{align}\label{eq:react_Coll_2Link_1D}
					\dot{\sigma}_i&= ({^{f_\mr{c}}}\bs{R}_\mr{0} \dot{\bs{x}}_\mr{c,t})_1=\dot{x}_\mr{loa,c,t}, \quad
					\bs{J}_i=({^{f_\mr{c}}}\bs{R}_\mr{0} \bs{J}_{x_\mr{c,t},q_\mr{a}})_1.
				\end{align} 
				with~${^{f_\mr{c}}}\bs{R}_\mr{0}$ rotating~$(\mr{CS})_0$ so that its first axis aligns with the direction of~$\hat{\bs{f}}_\mr{c}$ and~$(\cdot)_1$ expressing the selection of the argument's first row.
				Since only up to~$\mr{rank}(\bs{J}_{x_\mr{c,t},q_\mr{a}})$ DoF are required for this reaction, there are~$n_{\bs{q}_\mr{a}}{-}\mr{rank}(\bs{J}_{x_\mr{c,t},q_\mr{a}})$ redundant DoF.
			\subsubsection{Structure Opening for a Clamping Contact}
				If a clamping contact with the~$j$-th chain is predicted by the FNN from Sec.~\ref{ssec:ContactClass}, two tasks 
				\begin{subequations}\label{eq:react_Cla_Chain}
					\begin{align}
						\label{eq:klemm1}
						\dot{\sigma}_i &= \dot{q}_{\mr{a},j}, \quad \bs{J}_i = \bs{j}_i,\\
						\dot{\sigma}_{i+1} &= \dot{q}_{\mr{p},j}, \quad
						\bs{J}_{i+1} = \bs{j}_{q,q_\mr{a},j} 
						\label{eq:klemm2}
					\end{align} 
				\end{subequations}
				are initiated with~$\bs{j}_{q,q_\mr{a},j}$ as the row of~$\bs{J}_{q,q_\mr{a}}{=}\bs{J}_{q,x} \bs{J}_{x,q_\mr{a}}$ corresponding to the~$j$-th chain's passive joint.
				\begin{figure}[t!]
					\vspace{1.5mm} 
					\centering
					\includegraphics[width=0.8\columnwidth]{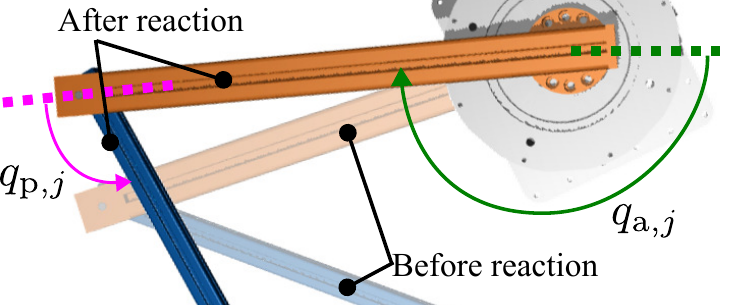}
					\caption{Clamping kinematic chain before (transparent) and after the reaction. Both links of the chain open the gap.}
					\label{fig:clampingreaction}
					\vspace{-1.5mm}
				\end{figure}	
				Figure~\ref{fig:clampingreaction} illustrates the clamping reaction described in~(\ref{eq:react_Cla_Chain}). 
				
				Reactions for contact cases limited by singularities of type~II and self-collisions are now considered.
				A \textit{feasible} reference reaction implemented by~$\dot{\bs{q}}_\mr{a,ref},\ddot{\bs{q}}_\mr{a,ref}$ or~$\bs{\tau}_\mr{a,ref}$ is described in the following, which incorporates singularity and self-collision avoidance and simultaneously removes contact. 
		\subsection{Redundancy Resolution on Kinematics Level} \label{ssec:redres_kin}
			The redundancy resolution for~$n_\mathcal{T}$ tasks can be formulated on the velocity level by
			\begin{subequations} \label{eq:rr_dq}
				\begin{align}
					\dot{\bs{q}}_\mr{a,ref} &= \sum_{i=1}^{n_\mathcal{T}-1} \bs{J}_i^\dagger 	\dot{\bs{\sigma}}_{\mr{ref},i} {+} \bs{N}_i 	\bs{J}_{i+1}^\dagger 	\dot{\bs{\sigma}}_{\mr{ref},{i+1}},\\
					\label{eq:rr_dq_sigma_ref}
		   			\dot{\bs{\sigma}}_{\mr{ref},i} & = \dot{\bs{\sigma}}_{\mr{d},i} {+} 	\bs{K}_{\dot{\sigma},i} 	(\bs{\sigma}_{\mr{d},i}{-}\bs{\sigma}_i),\\	
					\bs{N}_i &=\bs{I} {-} \bs{J}_{\mr{A},i}^\dagger \bs{J}_{\mr{A},i},\\
					\label{eq:rr_AugJacobi}
					\bs{J}_{\mr{A},i} &= [\bs{J}_1^\mr{T}, \bs{J}_2^\mr{T}, {\dotsc}, 	\bs{J}_i^\mr{T}]^\mr{T} \quad \text{and}\\
					\bs{\tau}_\mr{a,d}&= \bs{K}_{q_\mr{a}} \dot{\bs{q}}_\mr{a,ref} \Delta t {+} 	\bs{D}_{q_\mr{a}} 	(\dot{\bs{q}}_\mr{a,ref}{-}\dot{\bs{q}}_\mr{a}) {+} \hat{\bs{b}}_{q_\mr{a}}
				\end{align}
			\end{subequations}
			with~$\bs{N}_i$ as the nullspace projection matrix,~$\bs{J}_{\mr{A},i}$ as the augmented Jacobian matrix,~$\bs{K}_{\dot{\sigma}, i}$ as the gain of the task error~$(\bs{\sigma}_{\mr{d},i}{-}\bs{\sigma}_i)$ and with~$\hat{\bs{b}}_{q_\mr{a}}{=}\hat{\bs{c}}_{q_\mr{a}} {+} \hat{\bs{g}}_{q_\mr{a}} {+} \hat{\bs{\tau}}_{\mr{fr},q_\mr{a}}$.

			Better tracking performance can be achieved by formulating at acceleration level via
			\begin{subequations} \label{eq:rr_ddq}
				\begin{align}
					\ddot{\bs{q}}_\mr{a,ref} &{=} \sum_{i=1}^{n_\mathcal{T}-1} \bs{J}_i^\dagger 	(\ddot{\bs{\sigma}}_{\mr{ref},i} {-} 	\dot{\bs{J}}_i\dot{\bs{q}}_\mr{a} ) {+} \bs{N}_i \bs{J}_{i+1}^\dagger 	(\ddot{\bs{\sigma}}_{\mr{ref},i+1} {-} 	\dot{\bs{J}}_{i+1} \dot{\bs{q}}_\mr{a} ),\\
					\label{eq:rr_ddq_sigma_ref}
					\ddot{\bs{\sigma}}_{\mr{ref},i} &{=} \ddot{\bs{\sigma}}_{\mr{d},i} {+} 	\bs{K}_{\ddot{\sigma},i} 	(\bs{\sigma}_{\mr{d},i}{-}\bs{\sigma}_i) {+} \bs{D}_{\ddot{\sigma},i} (\dot{\bs{\sigma}}_{\mr{d},i}{-}\dot{\bs{\sigma}}_i)\quad \text{and}\\
					\bs{\tau}_\mr{a,d}&{=} \bs{M}_{q_\mr{a}} \ddot{\bs{q}}_\mr{a,ref} + 	\hat{\bs{b}}_{q_\mr{a}}.
				\end{align}
			\end{subequations}
		\subsection{Redundancy Resolution on Dynamics Level} \label{ssec:redres_dyn}
			The previous formulations at the velocity and acceleration level do not take the dynamics effects of the robot into account, which interact with the various tasks~\cite{Khatib.1987,JunNakanishi.2008,AlexanderDietrich.2015} and are shown in the following.
			Rearranging of~(\ref{eq:dyn_qa}) to
			\begin{align}
				\ddot{\bs{q}}_\mr{a} &= \bs{M}_{q_\mr{a}}^{-1} (\bs{\tau}_{\mr{a}} {+} \bs{\tau}_\mr{a,ext} {-} \bs{c}_{q_\mr{a}} {-} \bs{g}_{q_\mr{a}} {-} \bs{\tau}_{\mr{fr},q_\mr{a}} ) 	
			\end{align}
			and substituting into~(\ref{eq:DifKinSigma_dd}) leads to
				\begin{align}
					\label{eq:dyn_sigma_previous}
					\ddot{\bs{\sigma}}_i{=} \dot{\bs{J}}_i \dot{\bs{q}}_\mr{a} {+} \bs{J}_i \bs{M}_{q_\mr{a}}^{-1} (\bs{\tau}_{\mr{a}} {+} \bs{\tau}_{\mr{a,ext}}  {-} 	\bs{c}_{q_\mr{a}} 	{-} \bs{g}_{q_\mr{a}} {-} \bs{\tau}_{\mr{fr},q_\mr{a}}).
				\end{align}
			A non-redundant case with~$\bs{\tau}_{\mr{a}}{=}\bs{\tau}_{\mr{a,ref}} {=} \bs{J}_i^\mr{T}\bs{F}_{\mr{a,ref},i}$ is considered first.
			By multiplying
			\begin{align} \label{eq:mass_matrix_task}
				\bs{M}_i &= (\bs{J}_i\bs{M}_{q_\mr{a}}^{-1} \bs{J}_i^\mr{T})^{-1}
			\end{align} 			
			from left, the robot's dynamics
			\begin{align} \label{eq:dyn_sigma1}
				\bs{M}_i \ddot{\bs{\sigma}}_i {+} \bs{c}_i {+} \bs{g}_i {+} \bs{F}_{\mr{fr},i} 	&= \bs{F}_{\mr{a,ref,}i} {+} \bs{F}_{\mr{ext},i}
			\end{align} 
			expressed in the task coordinates follows. 
			Equation~\ref{eq:dyn_sigma1} contains 
			\begin{subequations}
			\begin{align} \label{eq:dyn_sigma2}
				\bar{\bs{J}}_{i} = \bs{M}_{q_\mr{a}}^{-1} \bs{J}_{i}^\mr{T}  (\bs{J}_i\bs{M}_{q_\mr{a}}^{-1} \bs{J}_i^\mr{T})^{-1}&= \bs{M}_{q_\mr{a}}^{-1} \bs{J}_{i}^\mr{T} \bs{M}_i,\\
				\bs{c}_i = \bs{M}_i (\bs{J}_i\bs{M}_{q_\mr{a}}^{-1}\bs{c}_{q_\mr{a}} {-} 	\dot{\bs{J}}_i\dot{\bs{q}}_\mr{a} ) &=\bar{\bs{J}}_{i}^\mr{T} 	\bs{c}_{q_\mr{a}}  - \bs{M}_i \dot{\bs{J}}_i\dot{\bs{q}_\mr{a}},\\
				\bs{g}_i = \bs{M}_i\bs{J}_i\bs{M}_{q_\mr{a}}^{-1}\bs{g}_{q_\mr{a}} &= 	\bar{\bs{J}}_{i}^\mr{T} \bs{g}_{q_\mr{a}},\\
				\bs{F}_{\mr{fr},i} = \bs{M}_i\bs{J}_i\bs{M}_{q_\mr{a}}^{-1}\bs{\tau}_{\mr{fr},q_\mr{a}} &= 	\bar{\bs{J}}_{i}^\mr{T} \bs{\tau}_{\mr{fr},q_\mr{a}} \quad \text{and}\\
				\bs{F}_{\mr{ext},i} = \bs{M}_i\bs{J}_i\bs{M}_{q_\mr{a}}^{-1}\bs{\tau}_\mr{a,ext} 	&= \bar{\bs{J}}_{i}^\mr{T} \bs{\tau}_\mr{a,ext}
			\end{align} 
			\end{subequations}
			with the dynamically consistent pseudoinverse~$\bar{\bs{J}}_{i}$~\cite{Khatib.1987}.
			For the redundant case, the principle of virtual work states that a force~$\bs{F}_{\mr{a},i}$ expressed in the~$i$-th task-space coordinates is projected onto the actuated-joint coordinates via
			\begin{align} \label{eq:PrinVirtWork_TaskCoord}
				\bs{\tau}_{\mr{a}} =\bs{J}_i^\mr{T}\bs{F}_{\mr{a},i} + \bar{\bs{N}}_i \bs{\tau}_{\mr{a},i+1}
			\end{align} 
			with the nullspace projection matrix~$\bar{\bs{N}}_i {=}\bs{I} {-} \bs{J}_{\mr{A},i}^\mr{T} \bar{\bs{J}}_{\mr{A},i}^\mr{T}$ and a torque~$\bs{\tau}_{\mr{a},i+1}$ in the nullspace of the augmented Jacobian matrix~$\bs{J}_{\mr{A},i}$ from~(\ref{eq:rr_AugJacobi}).
			If all task coordinates are stacked vertically, it follows
			\begin{align} \label{eq:dyn_sigma_all}
				[\ddot{\bs{\sigma}}_{1}^\mr{T},\cdots,\ddot{\bs{\sigma}}_{i}^\mr{T}]^\mr{T}	&{=} \dot{\bs{J}}_{\mr{A},i} \dot{\bs{q}}_\mr{a} {+} \bs{J}_{\mr{A},i} \bs{M}_{q_\mr{a}}^{-1} (\bs{\tau}_{\mr{a}} {+} \bs{\tau}_{\mr{a,ext}}  {-} 	\bs{b}_{q_\mr{a}}).
			\end{align}
			Substituting~(\ref{eq:PrinVirtWork_TaskCoord}) in~(\ref{eq:dyn_sigma_all}), the arbitrary torque~$\bs{\tau}_{\mr{a},i+1}$ vanishes due to the nullspace matrix's attribute described via
			\begin{align}
				\bs{M}_{\mr{A},i} &= (\bs{J}_{\mr{A},i} \bs{M}_{q_\mr{a}}^{-1} \bs{J}_{\mr{A},i}^\mr{T} )^{-1},\\
				\bar{\bs{J}}_{\mr{A},i} &= \bs{M}_{q_\mr{a}}^{-1} \bs{J}_{\mr{A},i}^\mr{T} \bs{M}_{\mr{A},i} \quad \text{and}\\
				\label{eq:nullspace_zero}
				\bs{J}_{\mr{A},i} \bs{M}_{q_\mr{a}}^{-1}\bar{\bs{N}}_i &{=\bs{J}_{\mr{A},i} \bs{M}_{q_\mr{a}}^{-1}(\bs{I} {-} \bs{J}_{\mr{A},i}^\mr{T} \bar{\bs{J}}_{\mr{A},i}^\mr{T}})\\
				\nonumber
				&{=}\bs{J}_{\mr{A},i} \bs{M}_{q_\mr{a}}^{-1} - \bs{J}_{\mr{A},i} \bs{M}_{q_\mr{a}}^{-1} \bs{J}_{\mr{A},i}^\mr{T} \bar{\bs{J}}_{\mr{A},i}^\mr{T} \\
				\nonumber
				&{=} \bs{J}_{\mr{A},i} \underbrace{\bs{M}_{q_\mr{a}}^{-1}}_{\bs{M}_{q_\mr{a}}^{-\mr{T}}} {-} \underbrace{\bs{J}_{\mr{A},i} \bs{M}_{q_\mr{a}}^{-1} \bs{J}_{\mr{A},i}^\mr{T}}_{\bs{M}_{\mr{A},i}^{-1}{=}\bs{M}_{\mr{A},i}^{-\mr{T}}} \bs{M}_{\mr{A},i}^\mr{T} \bs{J}_{\mr{A},i}  \bs{M}_{q_\mr{a}}^{-\mr{T}}\\
				\nonumber
				&{=} \bs{J}_{\mr{A},i} \bs{M}_{q_\mr{a}}^{-\mr{T}} - \bs{M}_{\mr{A},i}^{-\mr{T}} \bs{M}_{\mr{A},i}^\mr{T} \bs{J}_{\mr{A},i} \bs{M}_{q_\mr{a}}^{-\mr{T}} = \bs{0}.
			\end{align}
			Equation~\ref{eq:nullspace_zero} results in the zero matrix due to the symmetry of~$\bs{M}_{q_\mr{a}}$ and~$\bs{M}_{\mr{A},i}$.
			Generalized and formulated for~$n_\mathcal{T}$ tasks, it follows
			\begin{subequations} \label{eq:rr_tau}
				\begin{align}
					\bs{\tau}_\mr{a,ref} &{=} \sum_{i=1}^{n_\mathcal{T}-1} \bs{J}_i^\mr{T} \bs{F}_{\mr{a,ref},i} {+} \bar{\bs{N}}_i \bs{J}_{i+1}^\mr{T} \bs{F}_{\mr{a,ref},i+1},\\
					\label{eq:rr_tau_sigma_ref}
					\bs{F}_{\mr{a,ref},i} &{=} \bs{M}_{i} \ddot{\bs{\sigma}}_{\mr{d},i} {+} \bs{K}_{F,i} (\bs{\sigma}_{\mr{d},i}{-}\bs{\sigma}_i) {+} 	\bs{D}_{F,i} (\dot{\bs{\sigma}}_{\mr{d},i}{-}\dot{\bs{\sigma}}_i)\\
				     \text{and } \bs{\tau}_\mr{a,d}&{=}\bs{\tau}_{\mr{a,ref}}  {+} \hat{\bs{b}}_{q_\mr{a}}.
		   		\end{align}
			\end{subequations}
			By substituting~$\bs{F}_{\mr{a,ref},i}{+}\bar{\bs{J}}_i^\mr{T}\hat{\bs{b}}_{q_\mr{a}}$ from~(\ref{eq:rr_tau_sigma_ref}) into in~(\ref{eq:dyn_sigma1}), the closed-loop error dynamics 
			\begin{align}\label{eq:taskCoord_closedDynamics}
				\bs{M}_i \ddot{\bs{e}}_i + \bs{D}_{F,i} \dot{\bs{e}}_i + \bs{K}_{F,i} \bs{e}_i  = \bs{F}_{\mr{ext},i}
			\end{align}
			results with~$\bs{e}_i{=}(\bs{\sigma}_i{-}\bs{\sigma}_{\mr{d},i})$ in the $i$-th task coordinates.
			The control law described in~(\ref{eq:rr_tau}) is similar to the Cartesian impedance control from Sec.~\ref{ssec:controller} with~$\bs{\sigma}_1{=}\bs{x}$ and~$\bs{J}_1{=}\bs{J}_{x,q_\mr{a}}$.
			
			The damping matrix 
			\begin{align}
				\bs{D}_{F, i} &= \tilde{\bs{M}}_{i} \bs{D}_{\xi,i} \tilde{\bs{K}}_{F,i} + \tilde{\bs{K}}_{F,i} \bs{D}_{\xi,i} \tilde{\bs{M}}_{i}
			\end{align}
			is calculated with~$(\cdot){=}\tilde{(\cdot)}\tilde{(\cdot)}$ as in Sec.~\ref{ssec:controller} to achieve a configuration-independent damping behavior using the stiffness matrix~$\bs{K}_{F,i}$ and the mass matrix~$\bs{M}_i$ from~(\ref{eq:mass_matrix_task}) expressed in the task coordinates.
			Since the nullspace projection modifies the energetic flow of the control law in~(\ref{eq:rr_tau_sigma_ref}), the usage of Lyapunov functions describing only the single tasks is not sufficient to prove convergence~\cite{AlbuSchaeffer.2023} and the stability investigation is an open topic of active research~\cite{AlbuSchaeffer.2023,Dietrich.2012,AlexanderDietrich.2015, Dietrich.2020}, out of this papers' scope.
			Here, Lyapunov functions are used to obtain an insight into the parameterization of the matrices~$\bs{K}_{F,i}$ and~$\bs{D}_{F,i}$.
			The positive semi-definite Lyapunov function
			\begin{align} \label{eq:lyapunov}
				V_i= \frac{1}{2} \dot{\bs{e}}_i^\mr{T} \bs{M}_i \dot{\bs{e}}_i +  \frac{1}{2} \bs{e}_i^\mr{T} \bs{K}_{F,i} \bs{e}_i,
			\end{align}
			is chosen, whose time derivative is given by
			\begin{subequations} \label{eq:lyapunov_derivative}
				\begin{align} 
				\dot{V}_i{=}& \dot{\bs{e}}_i^\mr{T} (\bs{M}_i \ddot{\bs{e}}_i) {+} \frac{1}{2} \dot{\bs{e}}_i^\mr{T} \dot{\bs{M}}_i \dot{\bs{e}}_i {+}  \dot{\bs{e}}_i^\mr{T} \bs{K}_{F,i} \bs{e}_i {+}\frac{1}{2} \bs{e}_i^\mr{T} \dot{\bs{K}}_{F,i} \bs{e}_i,\\
				{=}& \dot{\bs{e}}_i^\mr{T}(\bs{F}_{\mr{ext},i} {-} \bs{D}_{F,i} \dot{\bs{e}}_i {-} \bs{K}_{F,i} \bs{e}_i){+}\\
				\nonumber
				& \frac{1}{2} \dot{\bs{e}}_i^\mr{T} \dot{\bs{M}}_i \dot{\bs{e}}_i {+}  \dot{\bs{e}}_i^\mr{T} \bs{K}_{F,i} \bs{e}_i {+}\frac{1}{2} \bs{e}_i^\mr{T} \dot{\bs{K}}_{F,i} \bs{e}_i,\\
				\dot{V}_i{=}&\dot{\bs{e}}_i^\mr{T} \bs{F}_{\mr{ext},i} {+} \frac{1}{2} \dot{\bs{e}}_i^\mr{T} \dot{\bs{M}}_i \dot{\bs{e}}_i {-} 	\dot{\bs{e}}_i^\mr{T} \bs{D}_{F,i} \dot{\bs{e}}_i {+} \frac{1}{2}  \bs{e}_i^\mr{T} \dot{\bs{K}}_{F,i} \bs{e}_i.
			\end{align}	
		\end{subequations}
			Equations~\ref{eq:lyapunov} and~\ref{eq:lyapunov_derivative} show that the matrices~$\bs{K}_{F,i}$ and~$\bs{D}_{F,i}$ must be positive definite, while~$\dot{\bs{K}}_{F,i}$ must be negative definite.
			The latter implies that the positive task stiffness is allowed to decrease over time but may only be increased if~$\bs{e}_i{=}\bs{0}$.
		
		\subsection{Inequality Constraints and Continuity of the Control Law}\label{ssec:IEC_Stetigkeit}
			The previous methods only take into account equality constraints (ECs). 
			However, numerous tasks need to be formulated and considered as inequality constraints (IECs).
			In this work, the inequality constraints are integrated similarly to those in~\cite{Moe.2016}, but they are generalized to kinematics and dynamics nullspace projections for PRs in HRC.
			
			For this purpose~$n_\mathcal{T}{=}3$ tasks are defined for collision reactions and $4$ for structure opening.
			Two are treated as IECs and the other as ECs: (i, IEC) the condition number~$\sigma_\kappa {=} \mr{cond}(\bs{J}_{x,q_\mr{a}})$ to account for type-II singularities, (ii, IEC) the minimum distance~$\sigma_\mr{sc}$ for self-collision avoidance and (iii and iv, ECs) the coordinates~$\bs{\sigma}_\mr{r}$ of the reaction from Sec.~\ref{ssec:CollClam_Reactions} to a collision or clamping. 
			
			Although the practical relevance of the condition number~$\sigma_\kappa$ is limited~\cite{Merlet.2006_JMD}, its usage for singularity avoidance will be demonstrated in the following sections. 
			Regarding~$\sigma_\mr{sc}$, the PR's six links are assumed to be one-dimensional line segments.
			Each link's minimal distance is calculated by referring to the other chains' links.
			
			The fulfillment of the limits of the $k$-th joint can be integrated into the presented approach with~$\dot{\sigma}_{\mr{d},i}{=}\dot{q}_{\mr{d},k}{=}0$ in~(\ref{eq:react_Cla_Chain}) as the $i$-th task, but only singularities of type~II and self-collisions are considered, which are more complex due to the modeling.
			
			The corresponding Jacobian matrices~$\bs{J}_\kappa$~(i) and~$\bs{J}_\mr{sc}$~(ii) are calculated numerically as in Alg.~\ref{alg:Jacobian}.
			\begin{figure}[b!]
				\removelatexerror
				\vspace{1.5mm}
				\begin{algorithm}[H] \nonumber
					\caption{Compute Jacobian~$\bs{J}_\kappa$ and~$\bs{J}_\mr{sc}$}
					\label{alg:Jacobian}
					{\small 
						\SetKwInOut{Input}{Input}
						\SetKwInOut{Output}{Output}
						\Input{Actuated-joint coordinates $\bs{q}_\mr{a}$} 
						\Output{$\bs{J}_\kappa, \bs{J}_\mr{sc}$}
						$\bs{x}\gets$ Forward kinematics for current $\bs{q}_{\mr{a}}$\;
						$\bs{J}_{x,q_\mr{a}}\gets$ Jacobian for current $\bs{q}_\mr{a}, \bs{x}$\;
						$\kappa\gets$ Condition number for current $\bs{J}_{x,q_\mr{a}}$\; 
						$d\gets$ Minimal distance between all links for current $\bs{q}_\mr{a}, \bs{x}$\;
						\For(\tcp*[f]{iterate active joint}){$i{:=}1$ to $n_{\bs{q}_\mr{a}}$} 
						{
							$\bs{q}_{\mr{a},\delta}\gets$ $\bs{q}_\mr{a}$\;
							$\bs{q}_{\mr{a},\delta}\gets$ Add increment $\delta$ to $i$-th actuated joint's angle\;
							$\bs{x}_{\delta}\gets$ Forward kinematics for $\bs{q}_{\mr{a},\delta}$\;
							$\bs{J}_{x,q_\mr{a}, \delta}\gets$ Jacobian for $\bs{q}_{\mr{a},\delta}$ and~$\bs{x}_\delta$\;
							$\kappa_\delta \gets$ Condition number of $\bs{J}_{x,q_\mr{a}, \delta}$\;
							$\bs{J}_{\kappa,i}\gets$ $(\kappa_\delta{-}\kappa)/\delta$\;
							$\bs{q}_\delta\gets$ Full inverse kinematics for $\bs{x}_\delta$\;
							$d_\delta\gets$ Compute minimal distance for $\bs{q}_\delta$\;
							$\bs{J}_{\mr{sc},i}\gets$ $(d_\delta{-}d)/\delta$\;
						}
					}
				\end{algorithm}
				\vspace{0mm}
			\end{figure}
			Each IEC is defined by safety values~$\underline{\sigma}_{i,\mr{s}}$ and~$\overline{\sigma}_{i,\mr{s}}$ to define the range~$\underline{\sigma}_{i,\mr{s}}{<}\sigma_i{<}\overline{\sigma}_{i,\mr{s}}$. 
			In addition, activation values~$\underline{\sigma}_{i,\mr{a}}{>}\underline{\sigma}_{i,\mr{s}}$ and~$\overline{\sigma}_{i,\mr{a}}{<}\overline{\sigma}_{i,\mr{s}}$ are formulated to obtain a complete trajectory for~$\sigma_{\mr{d}, i},\dot{\sigma}_{\mr{d},i}$ and~$\ddot{\sigma}_{\mr{d},i}$ from~$\underline{\sigma}_{i,\mr{a}}$ or~$\overline{\sigma}_{i, \mr{a}}$ to the final value~$\underline{\sigma}_{i,\mr{s}}$ or~$\overline{\sigma}_{i,\mr{s}}$ as input to the control laws in~(\ref{eq:rr_dq_sigma_ref}), (\ref{eq:rr_ddq_sigma_ref}) and (\ref{eq:rr_tau_sigma_ref}). 
			\highlightred{The rationale behind defining safety and activation values is to avoid undesired system behaviors, such as high-frequency oscillations in task activation due to sensor noise or discretization effects in calculating the task values~$\sigma_i$.
			This strategy is similar to the approach from~\cite{Arrichiello.2017}.
			The complete trajectory from an activation value to a safety value avoids this effect for all three types of control formulations.}
			
			The priority order, Jacobian matrices, and heuristically determined safety and activation values of the IECs are shown in Table~\ref{table:priority}.
			\begin{table}[t]
				\vspace{1.5mm}
				\caption{Set-based and equality-based tasks after contact detection}
				\label{table:priority}
				\vspace{-3.5mm}
				\centering
				\begin{tikzpicture}
					\node[inner sep=\spc] (t)
					{
						\begin{tabular}{c c c c c} 
							\hline
							Type& $\bs{\sigma}_i$ & $\bs{J}_i$ & $[\underline{\sigma}_{i,\mr{s}},\overline{\sigma}_{i,\mr{s}}]$ & 	$[\underline{\sigma}_{i,\mr{a}},\overline{\sigma}_{i,\mr{a}}]$\\ 
							\hline\hline
							IEC & $\sigma_a{=}\sigma_\kappa$ & $\bs{J}_\kappa$ & $[-,\overline{\sigma}_{a,\mr{s}}]$&$[-, \overline{\sigma}_{a,\mr{a}}]$\\ 
							\hline
							\cellcolor{gray!10}IEC & \cellcolor{gray!10}$\sigma_b{=}\sigma_\mr{sc}$ & \cellcolor{gray!10}$\bs{J}_\mr{sc}$ & \cellcolor{gray!10}$[ 	\underline{\sigma}_{b,\mr{s}} ,-]$& \cellcolor{gray!10}$[ \underline{\sigma}_{b,\mr{a}} ,-]$\\
							\hline
							EC & $\bs{\sigma}_\mr{r}$ & $\bs{J}_{\sigma_\mr{r},q_\mr{a}}$ & $-$& $-$ \\
							\hline
						\end{tabular}
					};
					\draw[myarrow, shorten >=25pt] (t.south west) -- (t.north west) node[midway,above,sloped, align=left] {increasing\\priority};
					\end{tikzpicture}
				\vspace{-5.5mm}	
			\end{table}
			\begin{table}[t]
			\vspace{1.5mm}
			\caption{Set-based and equality-based tasks before contact detection}
			\label{table:priority_woContact}
			\vspace{-3.5mm}
			\centering
			\begin{tikzpicture}
				\node[inner sep=\spc] (t)
				{
					\begin{tabular}{c c c c c} 
						\hline
						Type& $\bs{\sigma}_i$ & $\bs{J}_i$ & $[\underline{\sigma}_{i,\mr{s}},\overline{\sigma}_{i,\mr{s}}]$ & 	$[\underline{\sigma}_{i,\mr{a}},\overline{\sigma}_{i,\mr{a}}]$\\ 
						\hline\hline
						IEC & $\sigma_a{=}\sigma_\kappa$ & $\bs{J}_\kappa$ & $[-,\overline{\sigma}_{a,\mr{s}}]$&$[-, \overline{\sigma}_{a,\mr{a}}]$\\ 
						\hline
						\cellcolor{gray!10}IEC & \cellcolor{gray!10}$\sigma_b{=}\sigma_\mr{sc}$ & \cellcolor{gray!10}$\bs{J}_\mr{sc}$ & \cellcolor{gray!10}$[ 	\underline{\sigma}_{b,\mr{s}} ,-]$& \cellcolor{gray!10}$[ \underline{\sigma}_{b,\mr{a}} ,-]$\\
						\hline
						EC & $\bs{\sigma}_1{=}\bs{x}_\mr{t}$ & $\bs{J}_{x_\mr{t},q_\mr{a}}$ & $-$& $-$ \\
						\hline
						\cellcolor{gray!10}EC & \cellcolor{gray!10}$\bs{\sigma}_2{=}\bs{x}_\mr{r}$ & \cellcolor{gray!10}$\bs{J}_{x_\mr{r},q_\mr{a}}$ & \cellcolor{gray!10}$-$& \cellcolor{gray!10}$-$ \\ 
						\hline
					\end{tabular}
				};
				\draw[myarrow, shorten >=25pt] (t.south west) -- (t.north west) node[midway,above,sloped, align=left] {increasing\\priority};
			\end{tikzpicture}
			\vspace{-5.5mm}	
			\end{table}
			
			IECs are identified by letters in the index and ECs by numbers.
			Task prioritization decreases with consecutive letters and numbers, while all IEC-related tasks are higher prioritized.
			The task coordinates~$\bs{\sigma}_\mr{r}$ and Jacobian matrix~$\bs{J}_{\sigma_\mr{r},q_\mr{a}}$ result from one of the reaction tasks in~(\ref{eq:react_Coll_mobPl})--(\ref{eq:react_Cla_Chain}), depending on the results of the contact classification and localization.
			For the non-contact case, the tasks shown in Table~\ref{table:priority_woContact} are selected with the matrices~$\bs{J}_{x_\mr{t},q_\mr{a}}$ and~$\bs{J}_{x_\mr{r},q_\mr{a}}$ from~$\bs{J}_{x,q_\mr{a}}$ given in~(\ref{eq:DifKin_Jac2}).
			
			\begin{figure}[t!]
				\removelatexerror
				\vspace{1.5mm}
				\begin{algorithm}[H] \nonumber
					\caption{Mode selection}
					\label{alg:mode_selection}
					{\small 
						\SetKwInOut{Input}{Input}
						\SetKwInOut{Output}{Output}
						\Input{$\sigma_a,\sigma_b,\bs{\sigma}_i,\dot{\bs{\sigma}}_i,\ddot{\bs{\sigma}}_i$} 
						\Output{Desired motor torque $\bs{ \tau}_\mr{a,d}$ from~(\ref{eq:rr_dq}), (\ref{eq:rr_ddq}) or~(\ref{eq:rr_tau})}
						$b_{\sigma_a}, b_{\sigma_b} \gets $ Check status of $\sigma_a, \sigma_b$ according to Alg.~\ref{alg:check_status}\label{alg:mode_selection_line_bsigma}\;
						\uIf(\tcp*[f]{IECs are met}){$ b_{\sigma_a} \land b_{\sigma_b}$}
						{$\bs{\tau}_\mr{a,d}{:=}\bs{\tau}_\mr{a,d}(\bs{\sigma}_i,\dot{\bs{\sigma}}_i,\ddot{\bs{\sigma}}_i)$\tcp*[f]{mode 1}}
						\Else(\tcp*[f]{at least one task must be activated}){
							\uIf(\tcp*[f]{activate IEC task $\sigma_a$}){$ \neg b_{\sigma_a} \land b_{\sigma_b}$}
							{
								$\sigma_a,\dot{\sigma}_a,\ddot{\sigma}_a\gets$Trajectory from~$\overline{\sigma}_{a,\mr{a}}$ to~$\overline{\sigma}_{a,\mr{s}}$\;
								$\bs{\tau}_\mr{a,d}{:=}\bs{\tau}_\mr{a,d}(\sigma_a,\dot{\sigma}_a,\ddot{\sigma}_a, \bs{\sigma}_i,\dot{\bs{\sigma}}_i,\ddot{\bs{\sigma}}_i)$\tcp*[f]{mode 2}
							}
							\uElseIf(\tcp*[f]{activate IEC task $\sigma_b$}){$ b_{\sigma_a} \land \neg b_{\sigma_b}$}
							{
								$\sigma_b,\dot{\sigma}_b,\ddot{\sigma}_b\gets$Trajectory from~$\underline{\sigma}_{b,\mr{a}}$ to~$\underline{\sigma}_{b,\mr{s}}$\;
								$\bs{\tau}_\mr{a,d}{:=}\bs{\tau}_\mr{a,d}(\sigma_b,\dot{\sigma}_b,\ddot{\sigma}_b, \bs{\sigma}_i,\dot{\bs{\sigma}}_i,\ddot{\bs{\sigma}}_i)$\tcp*[f]{mode 3}
							}		
							\Else(\tcp*[f]{activate IEC tasks $\sigma_a,\sigma_b$}){$\bs{\tau}_\mr{a,d}{:=}\bs{\tau}_\mr{a,d}(\sigma_a,\dot{\sigma}_a,\ddot{\sigma}_a, \sigma_b,\dot{\sigma}_b,\ddot{\sigma}_b, \bs{\sigma}_i,\dot{\bs{\sigma}}_i,\ddot{\bs{\sigma}}_i);$\tcp*[f]{mode 4}}			
						}
					}
				\end{algorithm}
				\vspace{0mm}
			\end{figure}
			\begin{figure}[t!]
				\removelatexerror
				\vspace{1.5mm}
				\begin{algorithm}[H] \nonumber
					\caption{Check status of set-based task $\sigma_{i}$}
					\label{alg:check_status}
					{\small 
						\SetKwInOut{Input}{Input}
						\SetKwInOut{Output}{Output}
						\SetKwFunction{FMain}{checkStatus}
						\SetKwProg{Fn}{Function}{:}{}
						\Fn{\FMain{$\sigma_i,\underline{\sigma}_{i,\mr{a}},\overline{\sigma}_{i,\mr{a}}$}}{
						\uIf{$(\sigma_i > \underline{\sigma}_{i,\mr{a}}) \land (\sigma_i < \overline{\sigma}_{i,\mr{a}})$}
						{$b_{\sigma_i}\gets\mr{true}$\tcp*{in the range $\underline{\sigma}_{i,\mr{a}}{<}\sigma_i{<}\overline{\sigma}_{i,\mr{a}}$}}
						\Else{$b_{\sigma_i}\gets\mr{false}$\;}
					}
					\KwRet $b_{\sigma_i}$\;
				}
				\end{algorithm}
				\vspace{0mm}
			\end{figure}
			Since two IECs are considered, there are four possible states regarding their respective activation.
			The check of the tasks~$\sigma_a$ and~$\sigma_b$ for activation and selection of the modes are shown in Algorithms~\ref{alg:mode_selection} and~\ref{alg:check_status}.
			In Algorithm~\ref{alg:mode_selection}, line~\ref{alg:mode_selection_line_bsigma} checks whether the respective activation values~$\underline{\sigma}_{i,\mr{a}},\overline{\sigma}_{i,\mr{a}}$ are reached. 
			Depending on the result, the desired motor torques~$\bs{\tau}_{\mr{a,d}}$ are calculated considering the activated IECs and therefore in four possible ways. 
			
			Attention has to be taken to the transitions between these four possible states, which cause oscillations and, in the worst case, instabilities due to the abrupt change in the control law.
			To avoid this, the target values of the two modes are linearly interpolated.
			\begin{figure}[bt!]
				\vspace{1.5mm}			
				\centering
				\includegraphics[width=1\columnwidth]{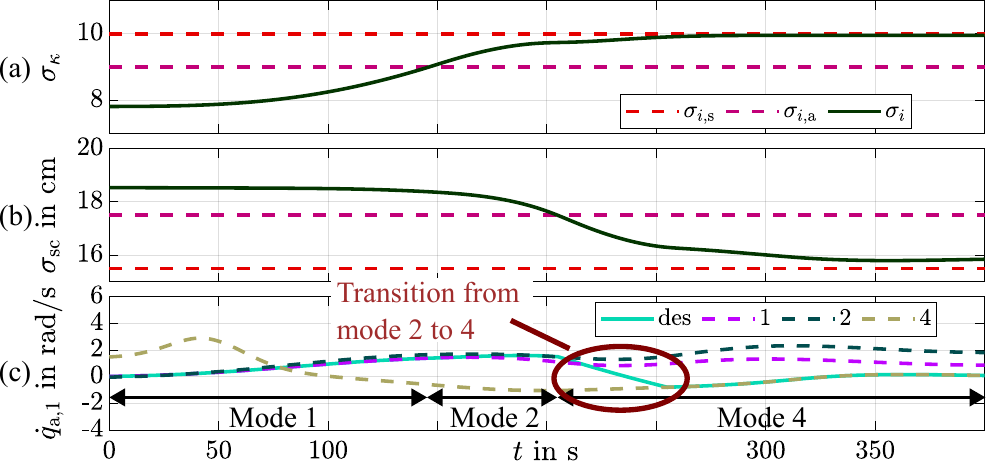}
				\caption{Curves of the performance characteristics (a)~$\sigma_\kappa$ and (b)~$\sigma_\mr{sc}$ to quantify the type-II singularities and self-collisions.
				(c) Curves of the required target drive speeds of the three modes when both IECs are inactive (mode 1), only~$\sigma_\kappa$ exceeds its activation value~$\overline{\sigma}_{\kappa,\mr{a}}$ and is active (mode 2) and when both IECs are activated (mode 4).
				The turquoise curve~($\mr{des}$) in the lower plot shows the chosen joint velocities.}
				\label{fig:Schaltvorgang_konti}
				\vspace{-1.5mm}
			\end{figure}
			Figure~\ref{fig:Schaltvorgang_konti} shows the linearly interpolated desired actuated joints' velocities in an exemplary movement with two mode transitions at~$t_{1,2}{=}\SI{150}{\milli \second}$ and~$t_{2,4}{=}\SI{220}{\milli \second}$.
			The second transition shows the linear transition from mode 2 to 4 in the time range~$220$--$\SI{260}{\milli \second}$. 
			Using this transition approach avoids abrupt changes in the command values.
			A transition duration of $\SI{10}{\milli\second}$ is defined in the following experiments. 

	\section{Validation} \label{sec:validation}
		This section begins with the test bench's description~(\ref{ssec:Val_Testbench}) and a performance evaluation of the different control laws~(\ref{ssec:Val_AccEvaluation}).
		Real-world results on contact reactions~(\ref{ssec:Val_Contacts_wo_RR}) are then presented, which are not restricted by the PR's limitations regarding singularities and self-collisions.
		After demonstrating in a simulation environment the safety-critical effects of disregarding these limitations~(\ref{ssec:Val_Simulation_wo_RR}), the results of fulfilling the IECs in a non-contact scenario~(\ref{ssec:Val_RR_wo_Contact}) are discussed.
		The results of the contact classification for localization and identification~(\ref{ssec:Val_IsolIdenti}) follow.
		Finally, results of the total detection-reaction process while fulfilling the IECs~(\ref{ssec:Val_RR_w_Contact}) are presented and evaluated regarding the force thresholds mentioned in~\cite{InternationalOrganizationforStandardization.2016}. 
						
		\begin{table*}[bt!]
			\vspace{1.5mm}
			\caption{Parameters for control laws}
			\label{table:controllerParam}
			\vspace{-1.5mm}
			\centering
			\begin{tabular}{l||c|c|c||c|c|c||c|c|c}
				\hline 
				\multirow{2}{*}{Redundancy-resolution mode}&\multicolumn{3}{c}{\textbf{Velocity}~(\ref{eq:rr_dq})} & \multicolumn{3}{c}{\textbf{Acceleration}~(\ref{eq:rr_ddq})}& \multicolumn{3}{c}{\textbf{Torque}~(\ref{eq:rr_tau})} \\
				& Symbol& Value & Unit & Symbol& Value & Unit& Symbol& Value & Unit\\
				\hline \hline
				\multicolumn{10}{c}{\textbf{Inequality constraints}} \\
				\hline 
				\multirow{2}{*}{Condition number}& \multirow{2}{*}{$k_{\dot\sigma,a}$} & \multirow{2}{*}{$20$} & \multirow{2}{*}{$1/\mr{s}$}& $k_{\ddot\sigma,a}$ & $1500$ & $1/\mr{s}^2$& $k_{F,a}$ & $1$ & $\mr{Nm}$ \\
				&&&& $d_{\ddot\sigma,a}$ & $62.5$ & $1/\mr{s}$& $d_{\xi,a}$ & $2$ & --- \\ \hline
				\multirow{2}{*}{Minimal distance}&\multirow{2}{*}{$k_{\dot\sigma,b}$} & \multirow{2}{*}{$30$} & \multirow{2}{*}{$1/\mr{s}$}& $k_{\ddot\sigma,b}$ & $1500$ & $1/\mr{s}^2$& $k_{F,b}$ & $16000$ & $\mr{N}/\mr{m}$ \\
				&&&&$d_{\ddot\sigma,b}$&$100$&$1/\mr{s}$& $d_{\xi,b}$ & $2$ & --- \\
				\hline \hline
				\multicolumn{10}{c}{\textbf{Reaction strategies} depending on contact classification, isolation and identification}  \\
				\hline 
				\multirow{2}{*}{Platform collision}& \multirow{2}{*}{$\bs{K}_{\dot\sigma,1}$} & \multirow{2}{*}{$\mr{diag}(30,30)$} & \multirow{2}{*}{$1/\mr{s}$}& $\bs{K}_{\ddot\sigma,1}$ & $\mr{diag}(2000,2000)$ & $1/\mr{s}^2$& $\bs{K}_{F,1}$ & $\mr{diag}(2000,2000)$ & $\mr{Nm}$ \\
				&&&& $\bs{D}_{\ddot\sigma,1}$ & $\mr{diag}(50,50)$ & $1/\mr{s}$& $\bs{D}_{\xi,1}$ & $\mr{diag}(1,1)$ & --- \\ \hline
				\multirow{2}{*}{1st link collision}&\multirow{2}{*}{$k_{\dot\sigma,1}$} & \multirow{2}{*}{$10$} & \multirow{2}{*}{$1/\mr{s}$}& $k_{\ddot\sigma,1}$ & $1250$ & $1/\mr{s}^2$& $k_{F,1}$ & $2500$ & $\mr{Nm}/\mr{rad}$ \\
				&&&&$d_{\ddot\sigma,1}$&$50$&$1/\mr{s}$& $d_{\xi,1}$ & $1$ & --- \\ \hline
				\multirow{2}{*}{2nd link collision}& \multirow{2}{*}{$\bs{K}_{\dot\sigma,1}$} & \multirow{2}{*}{$\mr{diag}(10,10)$} & \multirow{2}{*}{$1/\mr{s}$}& $\bs{K}_{\ddot\sigma,1}$ & $\mr{diag}(1250,1250)$ & $1/\mr{s}^2$& $\bs{K}_{F,1}$ & $\mr{diag}(2000,2000)$ & $\mr{N}/\mr{m}$ \\
				&&&& $\bs{D}_{\ddot\sigma,1}$ & $\mr{diag}(50,50)$ & $1/\mr{s}$& $\bs{D}_{\xi,1}$ & $\mr{diag}(1,1)$ & --- \\ \hline
				\multirow{4}{*}{Clamping}& \multirow{2}{*}{$k_{\dot\sigma,1,q_\mr{a}}$} & \multirow{2}{*}{$10$} & \multirow{2}{*}{$1/\mr{s}$}& $k_{\ddot\sigma,1,q_\mr{a}}$ & $1250$ & $1/\mr{s}^2$& $k_{F,1,q_\mr{a}}$ & $1000$ & $\mr{Nm}/\mr{rad}$ \\
				&&&& $d_{\ddot\sigma,1,q_\mr{a}}$ & $75$ & $1/\mr{s}$& $d_{\xi,1,q_\mr{a}}$ & $1$ & --- \\ 
				& \multirow{2}{*}{$k_{\dot\sigma,2,q_\mr{p}}$} & \multirow{2}{*}{$10$} & \multirow{2}{*}{$1/\mr{s}$}& $k_{\ddot\sigma,2,q_\mr{p}}$ & $1250$ & $1/\mr{s}^2$& $k_{F,2,q_\mr{p}}$ & $1000$ & $\mr{Nm}/\mr{rad}$ \\ 
				&&&& $d_{\ddot\sigma,2,q_\mr{p}}$ & $75$ & $1/\mr{s}$& $d_{\xi,2,q_\mr{p}}$ & $1$ & --- \\ 
				\hline \hline
				\multicolumn{10}{c}{\textbf{Robot control} during non-contact-phase and with deactivated IEC}  \\
				\hline 
				\multirow{2}{*}{Platform position}& \multirow{2}{*}{$\bs{K}_{\dot\sigma,1}$} & \multirow{2}{*}{$\mr{diag}(30,30)$} & \multirow{2}{*}{$1/\mr{s}$}& $\bs{K}_{\ddot\sigma,1}$ & $\mr{diag}(2000,2000)$ & $1/\mr{s}^2$& $\bs{K}_{F,1}$ & $\mr{diag}(2000,2000)$ & $\mr{N/m}$ \\
				&&&& $\bs{D}_{\ddot\sigma,1}$ & $\mr{diag}(50,50)$ & $1/\mr{s}$& $\bs{D}_{\xi,1}$ & $\mr{diag}(1,1)$ & --- \\ \hline
				\multirow{2}{*}{Platform orientation}&\multirow{2}{*}{$k_{\dot\sigma,2}$} & \multirow{2}{*}{$40$} & \multirow{2}{*}{$1/\mr{s}$}& $k_{\ddot\sigma,2}$ & $2500$ & $1/\mr{s}^2$& $k_{F,2}$ & $85$ & $\mr{Nm}/\mr{rad}$ \\
				&&&&$d_{\ddot\sigma,2}$&$75$&$1/\mr{s}$& $d_{\xi,2}$ & $1$ & --- \\ \hline
			\end{tabular}
		\end{table*}
		\subsection{Test Bench} \label{ssec:Val_Testbench}
			\begin{figure}[b!]
				\vspace{1.5mm} 
				\centering
				\includegraphics[width=\columnwidth]{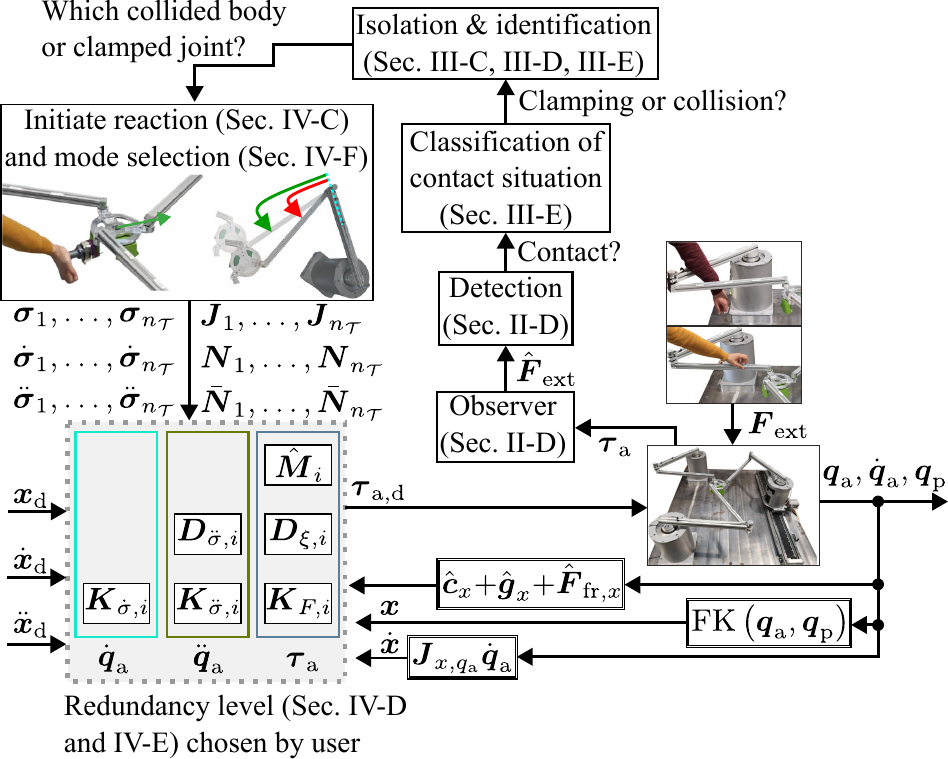}
				\caption{Block diagram of the algorithm illustrating the integration of the individual elements in \akr}
				\label{fig_Ablauf_3PRRR}
				\vspace{-1.5mm}
			\end{figure}	
			\begin{figure*}[t!]
				\vspace{1.5mm}			
				\centering
				\includegraphics[width=\textwidth]{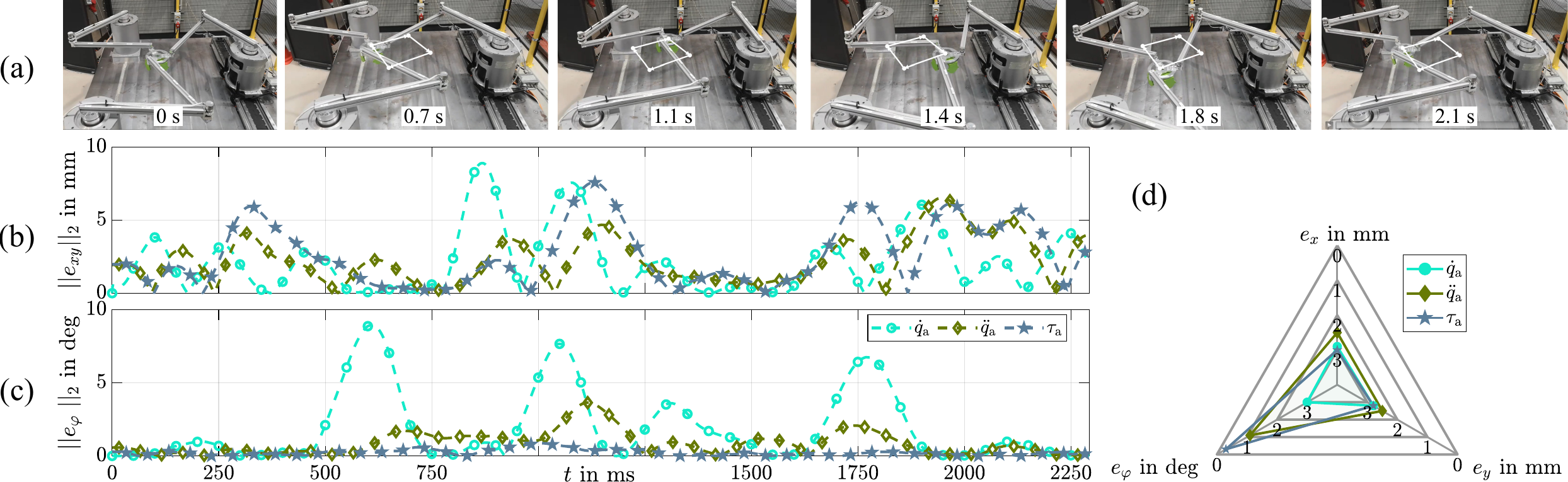}
				\caption{(a) Square trajectory (edge lengths~$\SI{300}{\milli \meter}$), Euclidean error of (b) translational and (c) rotational operational-space coordinates, and (d) root mean squared errors when applying the control laws on velocity ($\dot{q}_\mr{a},\mathcolor{dq}{\circ}$), acceleration~($\ddot{q}_\mr{a},\mathcolor{ddq}{\lozenge}$) and torque level~($\tau_\mr{a},\mathcolor{tau}{\bigstar}$).
				The acceleration- and torque-based control laws show lower orientation errors at high speeds of up to~$\SI{1.53}{\meter / \second}$.}
				\label{fig:120_Viereck_Regler_dq_ddq_tau_red}
				\vspace{-1.5mm}
			\end{figure*}
			\begin{figure}[bt!]
				\vspace{1.5mm} 
				\centering
				\includegraphics[width=\columnwidth]{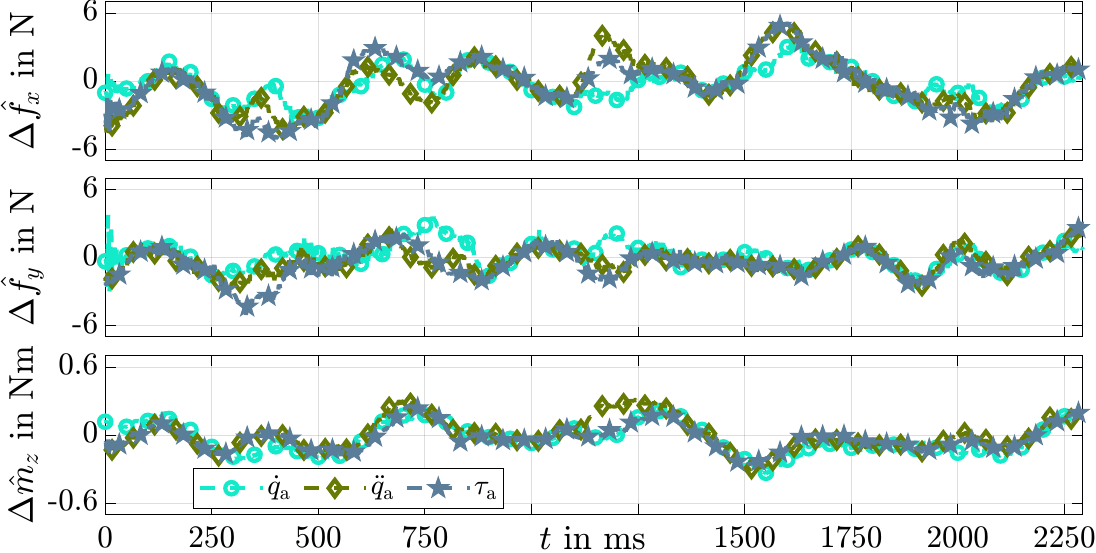}
				\caption{Force- and moment-estimation errors during the square trajectory.
					Low errors favor faster contact detection.}
				\label{fig:130_Viereck_Regler_Fext_error}
				\vspace{-1.5mm}
			\end{figure}	
			The active joints of the 3-\underline{R}RR parallel robot are actuated by three torque motors\footnote{KTY6288.4 from Georgii Kobold GmbH \& Co. KG} (gearless synchronous motors). 
			Absolute encoders\footnote{ECN1313 from Heidenhain GmbH \& Co. KG} measure angular positions with a system accuracy of~$\SI{0.0056}{\degree}$.
			The encoder signals are integrated into the data communication of the servo drive\footnote{S600 from Kollmorgen Europe GmbH}, which are then numerically differentiated and low-pass filtered with~$\SI{30}{\hertz}$ for the velocity computation. 
			Incremental encoders\footnote{RI36-H from Hengstler GmbH} measure passive joints' angles with an accuracy of~$\SI{0.1}{\degree}$.
			They are integrated into the data communication by a channel encoder interface\footnote{EL5101 from Beckhoff Automation} with $16$ bit. 
			Two force-torque sensors\footnote{KMS40 from Weiss Robotics, Mini40 from ATI Industrial Automation} (FTS) are used to measure the contact forces with~$\SI{500}{\hertz}$ and~$\SI{1000}{\hertz}$ to compare the different contact-reaction strategies. 
			Their measurement ranges are~$\pm\SI{120}{\newton}/\pm\SI{3}{Nm}$ and~$\pm\SI{240}{\newton}/\pm\SI{4}{Nm}$. 
			The EtherCAT protocol and the open-source tool EtherLab\footnote{\url{https://www.etherlab.org}} are used with an external-mode patch and a shared-memory real-time interface for data communication. 
			
			Figure~\ref{fig_Ablauf_3PRRR} represents the block diagram.
			The sampling rate of the control law is~$\SI{1}{\kilo\hertz}$. 
			The MO is parameterized with~$k_{\mathrm{o},i}{=}\frac{1}{\SI{50}{\milli \second}}$.
			
			Both IECs are only checked if a contact is detected.
			As soon as the reaction is finished and the user terminates the activated-IEC case via a graphical user interface, the IECs are deactivated again.
			
			The Matlab/Simulink software of the presented algorithms and hardware integration is published as open source with extended documentation, which represents contribution~\ref{contribution:software}.
			
		\subsection{Performance Evaluation of the Different Control Laws}\label{ssec:Val_AccEvaluation}
			Figure~\ref{fig:120_Viereck_Regler_dq_ddq_tau_red} shows a square trajectory with constant orientation and the resulting tracking errors regarding the operational-space coordinates.
			A jerk-limited motion profile with maximum velocities~$||\dot{\bs{x}}_{\mr{t}}||_2{=}\SI{1.53}{\meter / \second}$ and accelerations~$||\ddot{\bs{x}}_{\mr{t}}||_2{=}\SI{12}{\meter / \second^2}$ is performed with the respective control law.			
			
			In the experiments, the controller parameters in~(\ref{eq:rr_dq}), (\ref{eq:rr_ddq}) and~(\ref{eq:rr_tau}) are determined heuristically.
			Table~\ref{table:controllerParam} lists the parameters, which shows one disadvantage of the acceleration-based approach regarding the gain-tuning effort: Due to the formulation, the number of tuning parameters is doubled compared to the velocity approach, and low damping leads to the occurrence of oscillations.
			For the torque-based formulation, the factorization damping design~\cite{AlbuSchaffer.2003} simplifies the parameter tuning as long as the dynamics parameters are precisely identified.
			
			While in Fig.~\ref{fig:120_Viereck_Regler_dq_ddq_tau_red}(b)--(d), the position errors of all three approaches are similar (${<}\SI{10}{\milli \meter}$), the velocity-based approach has a higher orientation error over time~($\SI{9}{\degree}$) as well as on average ($\SI{3}{\degree}$).
			The higher orientation error can be critical with regard to the stability of the control law in configurations of a dynamic trajectory close to singularities, such as a retraction movement, since the associated rank loss of the Jacobian matrix leads to theoretically infinitely high motor torques.
			
			Figure~\ref{fig:130_Viereck_Regler_Fext_error} shows the MO's estimations during the contact-free square trajectory. 
			Due to the maximum errors of~$\SI{5}{\newton}$ and~$\SI{0.3}{Nm}$, the threshold values for contact detection are set to~$\SI{10}{\newton}$ and~$\SI{1}{Nm}$ to preclude false-positive detections.
		\subsection{Contact Experiments without Redundancy Resolution}\label{ssec:Val_Contacts_wo_RR}
\begin{figure}[t!]
				\vspace{1.5mm} 
				\centering
				\includegraphics[width=\columnwidth]{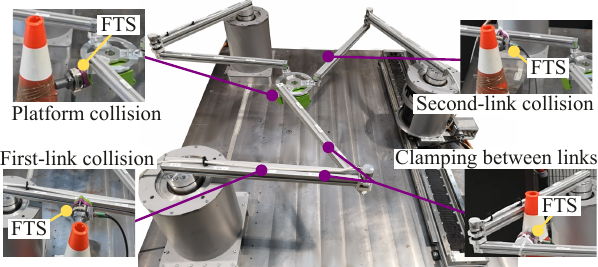}
				\caption{Collision and clamping experiments. A pylon is chosen as a contact partner and fixed to the base to achieve reproducible, comparable results.}
				\label{fig:coll_clamp_experiments}
				\vspace{-1.5mm}
			\end{figure}

			The detection and reaction results are now presented without being affected by the PR's limitations (singularities and risk of self-collision) due to the position in the workspace.
			For this purpose, contact experiments with a pylon, as in Fig.~\ref{fig:coll_clamp_experiments}, are carried out, and the performance of the reaction strategy \textit{retraction movement} from Sec.~\ref{ssec:Retraction} is evaluated.
			The pylon is fixed on the base to prevent its retraction.
			\begin{figure}[t!]
				\vspace{1.5mm} 
				\centering
				\includegraphics[width=\columnwidth]{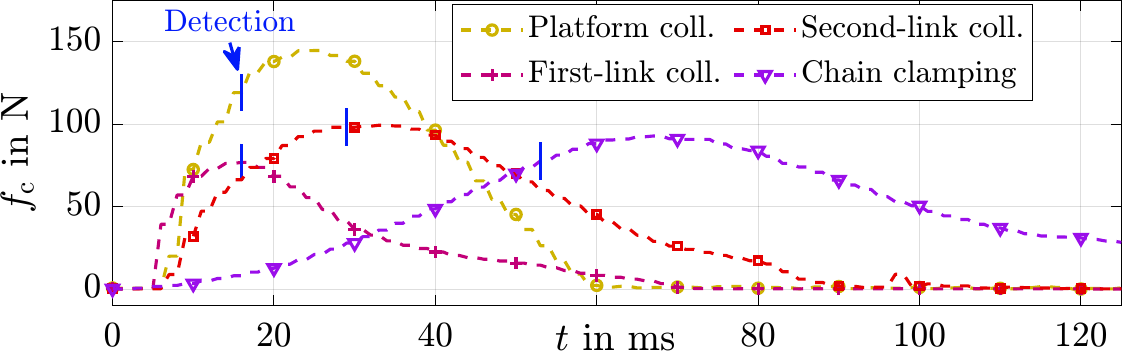}
				\caption{Measured forces of collision and clamping scenarios at the contact location. The blue marker shows the \textcolor{blue}{detection and start of the reaction}.
				Contact forces are lower than thresholds from~\cite{InternationalOrganizationforStandardization.2016}, which shows the potential of PRs for HRC.}
				\label{fig:coll_clamp_wo_redres}
				\vspace{-1.5mm}
			\end{figure}
			\begin{figure*}[h!]
			\vspace{1.5mm}			
			\centering
			\includegraphics[width=\textwidth]{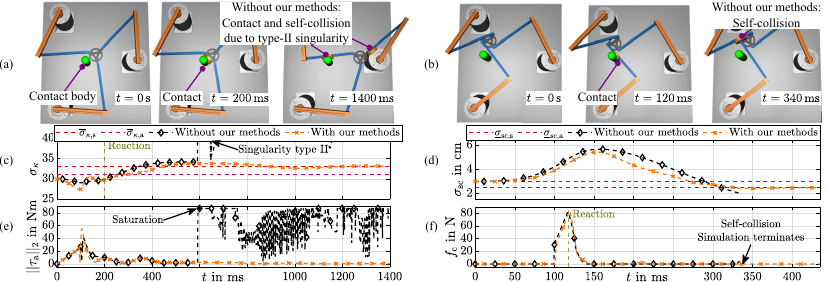}
			\caption{Results of a MuJoCo~\cite{Todorov.2012} simulation without (symbol~$\lozenge$) and with ($\mathcolor{orange}{\times}$) redundancy resolution to demonstrate a type-II singular configuration~(left) and a self-collision~(right).
				(a)--(b)~Image sections and (c)--(d)~curves with/without consideration of the IECs~$\sigma_\kappa,\sigma_\mr{sc}$, (e)--(f)~Euclidean norm~$||\bs{\tau}_\mr{a}||_2$ of the motor torques, as well as the simulated contact force~$f_\mr{c}$.
				\akr~enables safe and feasible reactions in the simulated scenarios.}
			\label{fig:100_plot_Kollision_mP_ImpReg_rrdq_SingTypII}
			\vspace{-1.5mm}
			\end{figure*}
			\begin{table}[bt!]
				\vspace{1.5mm}
				\caption{Detection and reaction results for the experiments in Fig.~\ref{fig:coll_clamp_wo_redres}}
				\label{table:coll_clamp_wo_redres}
				\centering
				\begin{tabular}{l||c|c|c|c}
				\hline
					\cellcolor{gray!0} & \cellcolor{platform!0}Platform& \cellcolor{1link!0}1st link & \cellcolor{2link!0}2nd link& \cellcolor{clamping!0}Clamping \\
					\cellcolor{gray!0} &\cellcolor{1link!0}$\mathcolor{platform}{\bs{\circ}}$& \cellcolor{1link!0}$\mathcolor{1link}{\bs{+}}$&
					\cellcolor{2link!0}$\mathcolor{2link}{\bs{\Box}}$& \cellcolor{clamping!0}$\mathcolor{clamping}{\bs{\triangledown}}$ \\\hline \hline
					\cellcolor{gray!0}$||\dot{\bs{x}}_{\mr{t}}||_2$ in m$/$s & \cellcolor{gray!0} 0.93 & \cellcolor{gray!0} 0.55 & \cellcolor{gray!0} 0.91 & \cellcolor{gray!0} 0.85\\\hline			
					\cellcolor{gray!10}$||\ddot{\bs{x}}_{\mr{t}}||_2$ in m$/$s$^2$ & \cellcolor{gray!10} 17.1 & \cellcolor{gray!10} 7.8 & \cellcolor{gray!10} 3.1 & \cellcolor{gray!10} 6.4\\\hline			
					\cellcolor{gray!0}$\Delta t_\mr{det}$ in ms & \cellcolor{gray!0} 10 & \cellcolor{gray!0} 11 & \cellcolor{gray!0} 23 & \cellcolor{gray!0} 46\\\hline
					\cellcolor{gray!10}$\Delta t_\mr{react}$ in ms & \cellcolor{gray!10} 44 & \cellcolor{gray!10} 53 & \cellcolor{gray!10} 63 & \cellcolor{gray!10} - \\\hline
					\cellcolor{gray!0}$f_\mr{c,max}$ in N & \cellcolor{gray!0} 144 & \cellcolor{gray!0} 76 & \cellcolor{gray!0} 99 & \cellcolor{gray!0} 93 \\\hline
				\end{tabular}
				\vspace{-1.5mm}
			\end{table}
			
			Figure~\ref{fig:coll_clamp_wo_redres} shows the results of the retraction motion for collisions and clamping contacts over time, while Table~\ref{table:coll_clamp_wo_redres} shows the maximum platform velocities~$||\dot{\bs{x}}_{\mr{t}}||_2$ and accelerations~$||\ddot{\bs{x}}_{\mr{t}}||_2$ during contact, the durations~$\Delta t_\mr{det},\Delta t_\mr{react}$ of the detection and reaction, as well as the maximum measured force~$f_{\mr{c,max}}$. 
			All four contacts are detected within~$10$--$\SI{46}{\milli \second}$, and the retraction movements cancel the collision contacts after~$\SI{63}{\milli \second}$ at the latest.
			The clamping force's curve in Fig.~\ref{fig:coll_clamp_wo_redres} shows that the clamping contact is not entirely canceled like the collisions.
			One possible reason is that the retraction movement with the translational platform coordinates takes place along the LoA of the estimated forces. 
			Therefore, the structure opening is only implicitly considered via the movement along the line of action, and the passive joint's angle of the clamping leg chain is not explicitly addressed in the reaction algorithm. 
			The last row in Table~\ref{table:coll_clamp_wo_redres} shows the maximum contact forces measured during the four experiments. 
			The reactions limit the contact forces to~$76$--$\SI{144}{\newton}$, so that the maximum permissible forces of transient~($280$--$\SI{320}{\newton}$) and quasi-static~($140$--$\SI{160}{\newton}$) contact with, e.g., the human arm and hands are complied with in accordance with ISO/TS 15066~\cite{InternationalOrganizationforStandardization.2016}. 
		\subsection{Simulation Results of Neglecting the Robot's Limitations}\label{ssec:Val_Simulation_wo_RR}
			The previous results assume that the reaction is neither restricted by type-II singularities nor by self-collisions, which would lead to an uncontrollable behavior with high commanded motor torques and, thus, an increased risk of injury to humans or damage to the robot. 
			The effects of the two limitations are simulated below.
			Figure~\ref{fig:100_plot_Kollision_mP_ImpReg_rrdq_SingTypII} shows the influence of a singular configuration (left) and a self-collision (right).
			
			Figure~\ref{fig:100_plot_Kollision_mP_ImpReg_rrdq_SingTypII}(c) shows the initiation of the PR's reaction to the contact at time~$\SI{200}{\milli\second}$, followed by the robot retreating.
			The condition number~$\sigma_\kappa$ increases and the PR approaches a singular configuration.
			The black curve in Fig.~\ref{fig:100_plot_Kollision_mP_ImpReg_rrdq_SingTypII}(c) corresponds to the scenario without addressing~$\sigma_\kappa$ by the redundancy resolution.
			At~$\SI{600}{\milli\second}$, the condition number increases abruptly because the PR is in a type-II singular configuration. 
			This causes a drastic increase in the motor torques in Fig.~\ref{fig:100_plot_Kollision_mP_ImpReg_rrdq_SingTypII}(e), which is limited by the software-side saturation.
			In the period~$600$--$\SI{1400}{\milli\second}$, oscillations occur until a simulation abort is triggered by a simultaneous collision of the PR with the contact body and a self-collision (see last image in Fig.~\ref{fig:100_plot_Kollision_mP_ImpReg_rrdq_SingTypII}(a)).
			
			The orange curves in Fig.~\ref{fig:100_plot_Kollision_mP_ImpReg_rrdq_SingTypII}(c),(e) now show the effect of integrating the IEC into the task-priority scheme in the algorithm.
			The IEC~$\sigma_\kappa$ is activated when the activation value~$\overline{\sigma}_{\kappa, \mr{a}}$ is exceeded, and the setpoint~$\overline{\sigma}_{\kappa, \mr{s}}$ is maintained, which prevents the PR from approaching the singular configuration.
			
			Figure~\ref{fig:100_plot_Kollision_mP_ImpReg_rrdq_SingTypII}(d),(f) depicts the PR in a configuration close to a self-collision. 
			The PR reacts to the contact with a retraction movement at the time $\SI{120}{\milli\second}$.
			As a result, a self-collision occurs at time~$\SI{340}{\milli\second}$, which is shown in the last image in Fig.~\ref{fig:100_plot_Kollision_mP_ImpReg_rrdq_SingTypII}(b) and leads to a simulation abort.
			The effect of the algorithm can also be seen here: The orange curve in Fig.~\ref{fig:100_plot_Kollision_mP_ImpReg_rrdq_SingTypII}(d) shows compliance with the IEC~$\sigma_\mr{sc}$.
			
			The simulation shows the redundancy resolution's advantage for fulfilling the PR's limits and a simultaneous reaction.
			The transfer to the real-world test bench and the evaluation during dynamic contact-free trajectories and contact experiments are carried out in the following.
		\subsection{Redundancy Resolution without Contact}\label{ssec:Val_RR_wo_Contact}
			\begin{figure}[t!]
				\vspace{1.5mm}			
				\centering
				\includegraphics[width=\columnwidth]{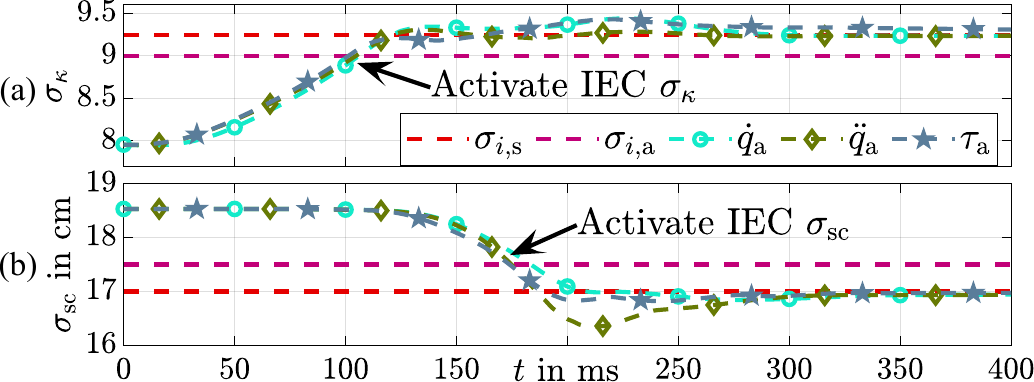}
				\caption{Tasks~(a)~$\sigma_\kappa$ and~(b)~$\sigma_\mr{sc}$ when applying the control laws~($\dot{\bs{q}}_\mr{a}$, $\ddot{\bs{q}}_\mr{a}$, $\bs{\tau}_\mr{a}$) with maximum velocities in the range~$1.25$--$\SI{1.35}{\meter / \second}$.
				All three control laws achieve the desired safety values but show overshoots\highlightred{, which are caused by high robot dynamics, mode-transition interpolation and resolving collision avoidance in the nullspace of the singularity-avoidance Jacobian.
				The safety threshold is conservatively set to compensate for this.}}
				\label{fig:110_plot_rr_wo_contact}
				\vspace{-1.5mm}
			\end{figure}
			\begin{figure}[t!]
				\vspace{1.5mm}			
				\centering
				\includegraphics[width=\columnwidth]{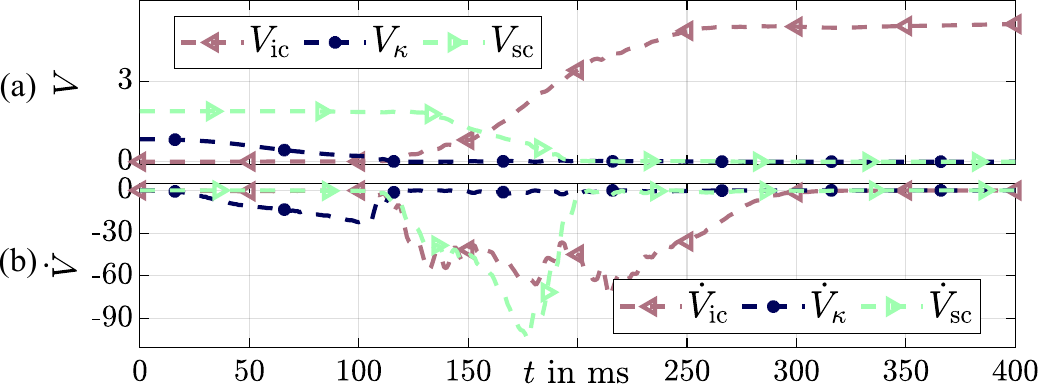}
				\caption{(a) Lyapunov function~$V_i$ and~(b) its time derivative~$\dot{V}_i$ for redundancy resolution at the torque level. Impedance control~($\mr{ic}$) is applied via the torque-based approach from Sec.~\ref{ssec:redres_dyn}.}
				\label{fig:111_plot_rr_tau_wo_contact_lyapunov}
				\vspace{-1.5mm}
			\end{figure}
			Dynamic contact-free trajectories with maximum speeds of up to~$||\dot{\bs{x}}_{\mr{t}}||_2{=}\SI{1.35}{\meter / \second}$ are chosen to evaluate the controllers' performances regarding complying with the IECs. 			
			Figure~\ref{fig:110_plot_rr_wo_contact} displays the curves of the IECs~$\sigma_\kappa$ and~$\sigma_\mr{sc}$ when applying the control laws~(\ref{eq:rr_dq}), (\ref{eq:rr_ddq}) and~(\ref{eq:rr_tau}). 
			Short-term overshoots occur, as can be seen in Fig.~\ref{fig:110_plot_rr_wo_contact}(b) for~$\sigma_\mr{sc}$ at time~$\SI{200}{\milli \second}$.
			Possible reasons are the high velocity and acceleration of the robot and the linear interpolation~(see Sec.~(\ref{ssec:IEC_Stetigkeit})) of the different modes' solution, which is calculated at the transition between two modes. 
			The acceleration-based approach shows the largest overshoot due to the empirically determined controller parameters.
			The safety threshold~$\underline{\sigma}_\mr{sc,s}$ is set to~$\SI{17}{\centi\meter}$ to consider the high dynamic movements and the thickness~($\SI{6}{\centi \meter}$) of the links since it is neglected in the minimal-distance calculation.
			However, falling below~$\underline{\sigma}_\mr{sc,s}$ only occurs for~$\SI{100}{\milli \second}$, so both IECs are then fulfilled for this trajectory. 
			
			The passivity of the torque-based control law is now investigated empirically by evaluating the Lyapunov function~(\ref{eq:lyapunov}) and its time derivative~(\ref{eq:lyapunov_derivative}) for the cases of impedance control, as well as compliance with the IECs. 
			Figure~\ref{fig:111_plot_rr_tau_wo_contact_lyapunov} depicts the results and that passivity of the single tasks shown by the positive semi-definite Lyapunov function and its negative-definite time derivative is maintained for all three cases in the example. 
			
			For the previous experiments, information on the type, location and force of the contact is not necessary. 
			However, the reactions with redundancy resolution and explicitly integrated contact coordinates require this information, \highlightred{which is why the results of contact classification, isolation and identification are presented below.
			It is obtained with the methods presented in~\cite{Mohammad.2023_IsolLoc, Mohammad.2023_Reaction}.}
			
		\highlightred{
			\subsection{Isolation and Identification}\label{ssec:Val_IsolIdenti}
				\subsubsection{Isolation of Clamping Contact}
					\begin{figure}[t!]
						\vspace{1.5mm}			
						\centering
						\includegraphics[width=1\columnwidth]{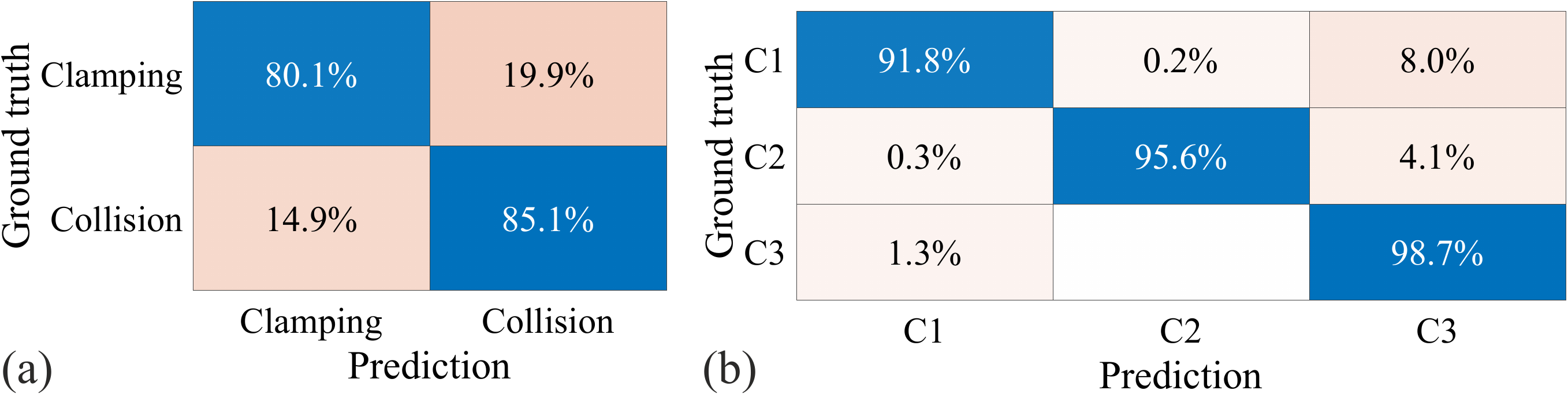}
						\caption{\highlightred{Row-normalized test results in confusion matrices (a) for the clamping/collision and (b) clamping-chain classification from~\cite{Mohammad.2023_Reaction}.
						Training and test data are obtained in different joint-angle configurations (C$i$ is the~$i$-th kinematic chain).
						Using only standard built-in sensors leads to accuracies larger than~$80\%$, which favors fast and cost-effective application of~\akr.}}
						\label{fig:140_Classification_Clamping_And_Chain_red}
						\vspace{-1.5mm}
					\end{figure}
					The feedforward neural networks (FNNs) used for clamping classification from Sec.~\ref{ssec:ContactClass} are evaluated.	
					Both FNNs are trained and evaluated using the inputs ($\hat{\bs{F}}_\mr{ext,mP}$ and $\bs{d}$) and known labeled outputs from chain clamping and collisions across the robot's structure in different joint-angle configurations.
					There are three robot configurations available in total.
					Each configuration has three clamping cases and seven collision points (six links and the platform).
					The dataset consists of 80,000 measurement samples, split into 70\% for training and 30\% for testing.
					The FNN's ability to generalize is evaluated by training and testing using data from different joint-angle configurations.
					The balanced training dataset is used to optimize hyperparameters, such as network architecture and a regularization term for the avoidance of overfitting.
					Since the classifiers are trained on a dataset that includes the sensor characteristics of current and joint-angle measurements, according sensor noise is represented in the training data.					
					The hyperparameter optimizations via a five-fold cross-validation process yielded two FNNs.
					These networks have five hidden layers, with 30 neurons per hidden layer for clamping classification and 25 neurons per layer for classifying the affected leg chain.
					Figure~\ref{fig:140_Classification_Clamping_And_Chain_red}(a) shows that 80\% of all clamping and 85\% of all collision test cases are correctly classified.\\
					If a clamping contact is classified, the second FNN predicts the affected kinematic chain.
					Figure~\ref{fig:140_Classification_Clamping_And_Chain_red}(b) shows the test results of the clamping-chain classification.
					The clamping chain is classified with an accuracy greater than 90\% due to the use of minimum distances $\bs{d}$ as a physically modeled feature in addition to estimated external forces in $\hat{\bs{F}}_\mr{ext,mP}$.
					Because the inputs to the FNNs are computed at the sampling rate, classifications are performed at the same time step as the detection.
					This enables an immediate clamping reaction.
				\subsubsection{Collision Isolation and Identification}
					\begin{figure}[t!]
						\vspace{1.5mm}			
						\centering
						\includegraphics[width=1\columnwidth]{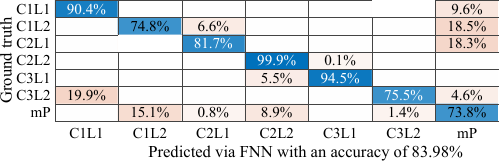}
						\caption{\highlightred{Row-normalized confusion matrices with test results of collided-body classification with the FNN from~\cite{Mohammad.2023_IsolLoc}. Training is performed in only one robot configuration while testing in different ones (C$i$L$j$ is the~$j$-th link of the~$i$-th kinematic chain and mP the mobile platform).
						The real-time capable FNN achieves an accuracy of~$84\%$, although the test data represent an unknown robot configuration.}}
						\label{fig:150_Classification_AffectedBody_Test_FFNN}
						\vspace{-1.5mm}
					\end{figure}
					\begin{figure}[t!]
						\vspace{1.5mm}			
						\centering
						\includegraphics[width=1\columnwidth]{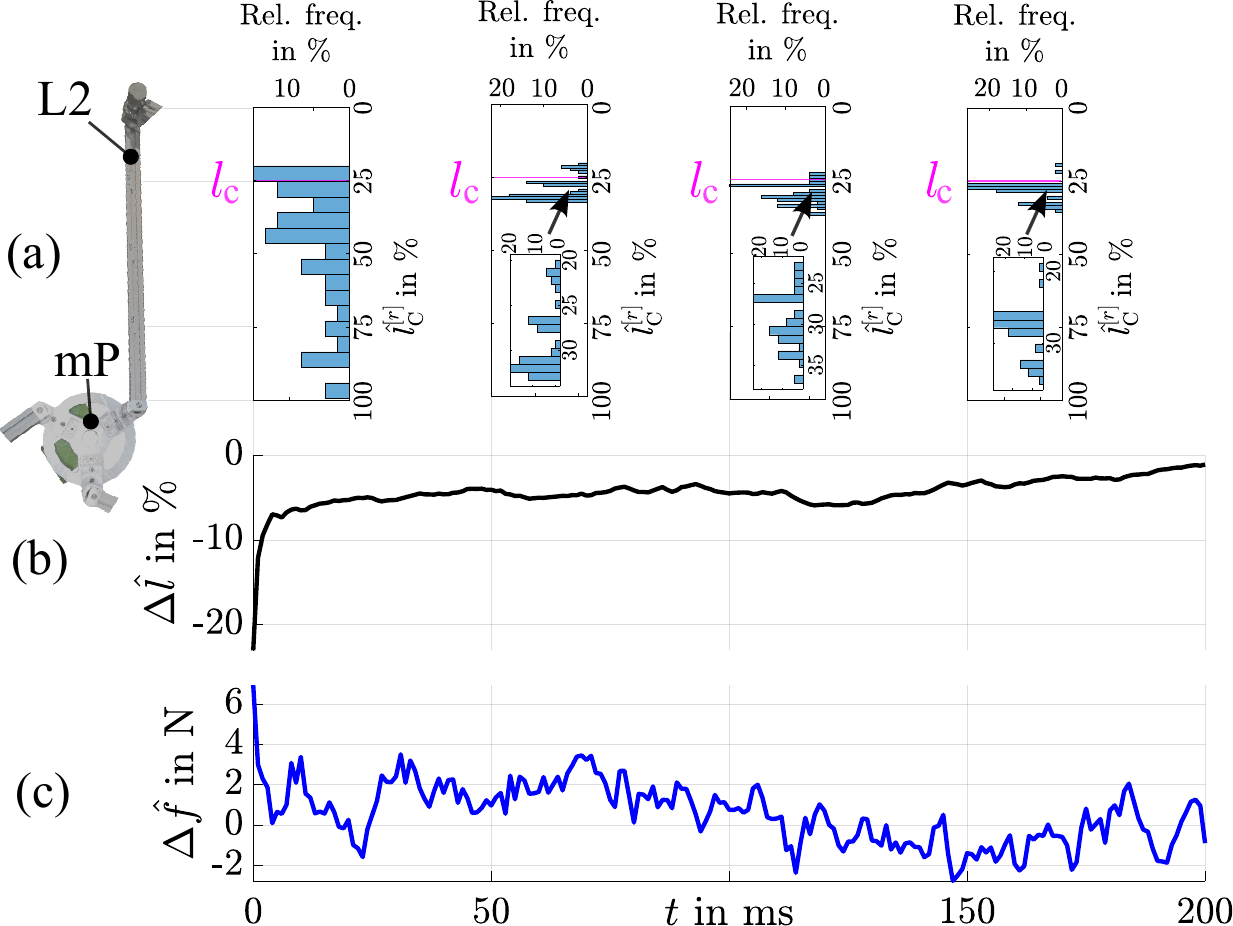}
						\caption{\highlightred{(a) Particle distribution as histograms over the second link's length at different time steps during a collision at a second link~\cite{Mohammad.2023_IsolLoc}. (b) Normalized isolation error~$\Delta\hat{l}$ and (c) identification error~$\Delta \hat{f}$.
						Low isolation and identification error enable reaction strategies for second-link collisions.}}
						\label{fig:160_pf_verteilung_querkraft_red}
						\vspace{-1.5mm}
					\end{figure}
					\begin{figure}[h!]
						\vspace{1.5mm}			
						\centering
						\includegraphics[width=1\columnwidth]{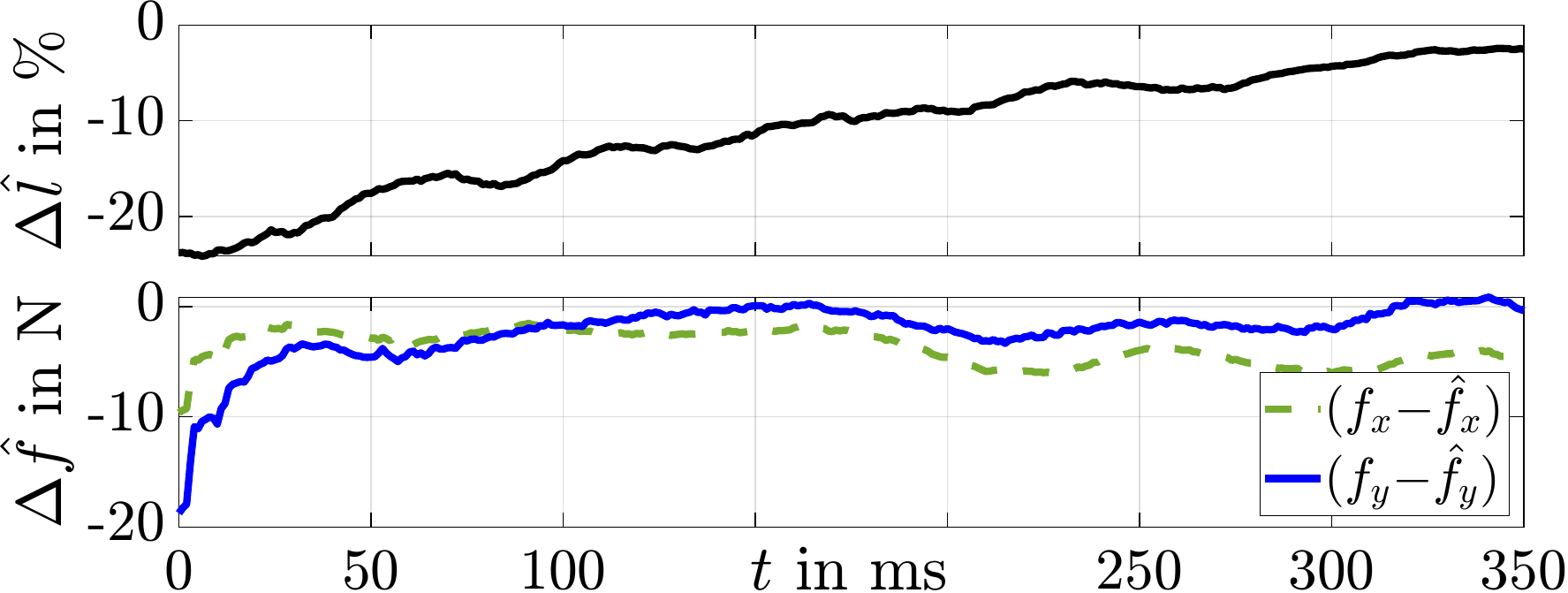}
						\caption{\highlightred{Localization- and identification-errors with the particle filter for angled collisions}}
						\label{fig:900_pf_angled}
						\vspace{-1.5mm}
					\end{figure}
					If the clamping classifier predicts a collision, a collision-body-multi-classification FNN is used to localize the contact on one of seven bodies (six links and the mobile platform).
					The training dataset contains 20 sets of measurements, comprising a total of 170,000 samples from a single configuration. This is used to determine the FNN's weights.
					Additionally, five-fold cross-validation is performed on the training dataset to determine the hyperparameters of the FNN.
					The generalization of the FNN to two different robot configurations, which together include 40 sets of measurements (310,000 samples), is evaluated using the row-normalized test results shown in Fig.~\ref{fig:150_Classification_AffectedBody_Test_FFNN}.
					The FNN achieves 84\% accuracy in prediction, despite the different joint-angle configurations.
					This can be attributed to the use of physically modeled features.
					If a second link is classified as the collided body, the particle filter~from Sec.~\ref{ssec:IsolItend_PF} is initiated with~$50$ particles.
					Figure~\ref{fig:160_pf_verteilung_querkraft_red} shows the time evolution of the particle filter's results in an experiment with a push force. 
					The trend of the estimation~$\hat{l}_\mr{c}$ toward the contact location~$l_\mr{c}$~(magenta line) is also apparent from the time history of the particle distributions in Fig.~\ref{fig:160_pf_verteilung_querkraft_red}(a).
					In Fig.~\ref{fig:160_pf_verteilung_querkraft_red}(b), the link-length normalized isolation error~$\Delta \hat{l}{=}(l_\mr{c}{-} \hat{l}_\mr{c})$ is less than~$4\%$ after~$\SI{50}{\milli \second}$, corresponding to~$\SI{24}{\milli \meter}$. 
					The identification error~$\Delta \hat{f}_y{=}(f_{y}{-}\hat{f}_{y})$ is below~$\SI{4}{\newton}$ after~$\SI{50}{\milli \second}$ in Fig.~\ref{fig:160_pf_verteilung_querkraft_red}(c).
					This is ten times shorter than the defined duration of a transient contact phase~\cite{InternationalOrganizationforStandardization.2016}.
					The particle filter is now adapted to non-orthogonal collisions via $\bs{p}_k^{[r]} {=} [l_\mr{c}^{[r]},\hat{f}_{x}^{[r]},\hat{f}_{y}^{[r]}]^\mr{T}$ with $\hat{f}_{x}^{[r]}$ and~$\hat{f}_{y}^{[r]}$ being force components for angled contact forces.
					Figure \ref{fig:900_pf_angled} shows the estimation of the extended particle filter for an angled collision at the second link of a kinematic chain.
					It demonstrates that the localization error is less than 10\% after $\SI{150}{\milli \second}$ and the identification error is less than $\SI{10}{\newton}$ after a few milliseconds.\\
					Thus, the estimation of contact location and force based on proprioceptive information is successfully applied and can be utilized in a subsequent reaction.
					Also, the estimated forces in $\hat{\bs{F}}_{\mr{ext,mP}}$ already capture different speeds and angled collisions, since the momentum observer also includes velocity-related effects.
					Since the neural network's features $\bs{d}$ and $\bs{\alpha}$ are computed based on $\hat{\bs{F}}_{\mr{ext,mP}}$, the classification based on them is also consistent.}
				
				These results demonstrate the potential of the detection, classification, isolation and identification methods, which are based on physically modeled features relying only on proprioceptive sensing and occur in the same single time step.
				Based on this, the reaction strategies~(Sec.~\ref{ssec:CollClam_Reactions}) are applied, extended by the redundancy resolution from Sec.~\ref{ssec:redres_kin} and~\ref{ssec:redres_dyn}.
				This constitutes \akr, the overall detection-reaction process for safe PRs.
				The following section shows the reaction results of \akr.
		\subsection{Contact Reaction with Redundancy Resolution}\label{ssec:Val_RR_w_Contact}
			After introducing the implementation~(\ref{ssec:Implementation}), four single collision and clamping experiments~(\ref{ssec:Single_CollCLampingExperiments}) are evaluated regarding the measured contact forces and the PR's limitation.
			A systematic analysis of 72 high-speed collision and clamping experiments (\ref{ssec:Multiple_CollCLampingExperiments}) follows.
			\subsubsection{Implementation} \label{ssec:Implementation}
				\begin{figure}[tb!]
					\removelatexerror
					\vspace{1.5mm}
					\begin{algorithm}[H]
						\caption{Detection-reaction sequence of \akr}\label{alg:Detection_Reaction_Process}
						{\small 
							\SetKwInOut{Input}{Input}
							\SetKwInOut{Output}{Output}
							\Input{$\bs{q}, \dot{\bs{q}}, \bs{x}, \dot{\bs{x}}, \hat{\bs{F}}_\mr{ext,mP}, \hat{\bs{g}}_{q_\mr{a}}, \bs{\epsilon}_\mathrm{r}, \bs{\epsilon}_\mathrm{g}$}
							\Output{Desired motor torque $\bs{\tau}_{\mr{a,d}}$}
							\uIf{$|\hat{F}_{\mr{ext,mP},i}|{\ge}\epsilon_{\mr{g},i}$\label{alg:Detection_Reaction_Process_If_zeroG}}
							{\tcp{zero-g mode as a fallback reaction}
								Set $\bs{\tau}_{\mr{a,d}}{:=}\hat{\bs{g}}_{q_\mr{a}}$\label{alg:Detection_Reaction_Process_zeroG}\;
							}
							\uElseIf{$|\hat{F}_{\mr{ext,mP},i}|{\ge}\epsilon_{\mr{r},i}$}
							{\tcp{reaction strategies from Sec.~\ref{ssec:CollClam_Reactions}}								
								$b_\mr{clamp}\gets$ Clamping classified by FNN (Sec.~\ref{ssec:ContactClass_I})\label{alg:Detection_Reaction_Process_clamp_start}\;
								\uIf{$b_\mr{clamp}$}
								{\tcp{structure opening - Eq. \ref{eq:react_Cla_Chain}}
									$j\gets$ Chain classified by FNN (Sec.~\ref{ssec:ContactClass_II})\label{alg:Detection_Reaction_Process_clamp_end}\;
									$\sigma_\mr{r1},\sigma_{\mr{r}2}\gets$ Active \& passive joints' angles $\sigma_{\mr{a},j},\sigma_{\mr{p},j}$\;
								}
								\Else
								{\tcp{retraction movement}
									$j\gets$ Collided body classified by FNN (Sec.~\ref{ssec:ClassAlg_FNN})\label{alg:Detection_Reaction_Process_collision_body_predicted}\;
									\uIf{$j{=}$mobile platform \label{alg:Detection_Reaction_Process_collision_body_platform}}
									{\tcp{mobile platform - Eq. \ref{eq:react_Coll_mobPl}}
										$\bs{\sigma}_\mr{r1}\gets$ Desired platform position $\tilde{\bs{x}}_\mr{t,d}$\;
									}
									\uElseIf{$j{\in}\{\mr{C1L1},\mr{C2L1},\mr{C3L1}\}$}
									{\tcp{first link - Eq. \ref{eq:react_Coll_1Link}}
										$\sigma_\mr{r1}\gets$ Active joint's angle $q_{\mr{a},j}$\;
									}
									\ElseIf{$j{\in}\{\mr{C1L2},\mr{C2L2},\mr{C3L2}\}$}
									{\tcp{second link - Eq. \ref{eq:react_Coll_2Link}}
										$\hat{l}_\mr{c},\hat{f}_\mr{c}\gets$ Particle filter's first estimation of location and force (Sec.~\ref{ssec:IsolItend_PF})\label{alg:Detection_Reaction_Process_collision_body_2link}\;
										$\bs{\sigma}_\mr{r1}\gets$ Contact point's velocity $\dot{\bs{x}}_{\mr{c,t}}$\;
									}
								}
								$\bs{\sigma}_\mr{r},\dot{\bs{\sigma}}_\mr{r},\ddot{\bs{\sigma}}_\mr{r}\gets$ Trajectory planning for selected tasks\label{alg:Detection_Reaction_Process_smooth_trajectory}\;
								$\bs{\tau}_{\mr{a,d}}\gets$ Desired motor torque by Alg.~\ref{alg:mode_selection}\;
							}
							\Else 
							{\tcp{no detection and reaction}
								$\bs{\tau}_{\mr{a,d}}\gets$ Preused control law of the previous task\;
							}
						}
					\end{algorithm}
					\vspace{0mm}
				\end{figure}
				\begin{figure}[t!]
					\vspace{1.5mm}			
					\centering
					\includegraphics[width=1\columnwidth]{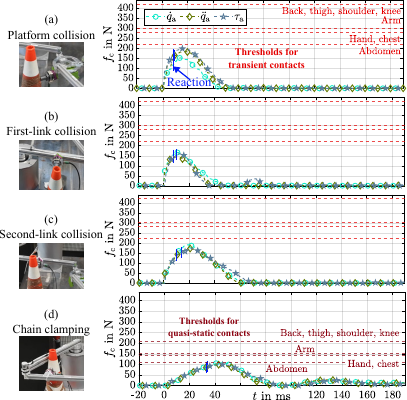}
					\caption{Measured contact forces in collision and clamping experiments with a pylon as a fixed contact body.
					\highlightred{Red dotted lines indicate permissible transient and quasi-static contact force thresholds as specified in ISO/TS 15066, which vary by human body region.} 
			 		\akr~enables safe reaction methods in high-speed scenarios with estimated contact type, location and force.
					Corresponding~$\sigma_\kappa$ and~$\sigma_\mr{sc}$ are shown in Fig.~\ref{fig:170_plot_kollision_klemmung_rr_iec_red}.}
					\label{fig:180_plot_kollision_klemmung_rr_Fc_red}
					\vspace{-1.5mm}
				\end{figure}
				\begin{figure}[t!]
					\vspace{1.5mm}			
					\centering
					\includegraphics[width=1\columnwidth]{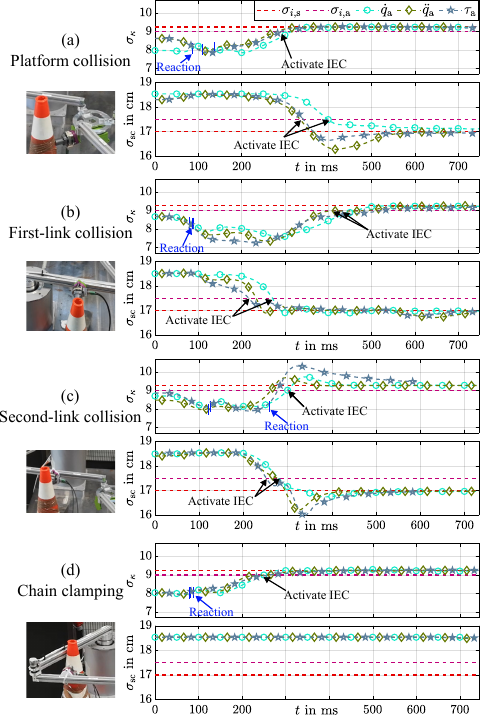}
					\caption{$\sigma_\kappa$ and~$\sigma_\mr{sc}$ before and after (a)--(c) collision and (d) clamping contacts with different redundancy resolution approaches.
					\highlightred{Overshoots, for example in (c) for $\sigma_\kappa$ at $\SI{300}{\milli\second}$, result from high robot dynamics and interpolation during mode transitions. The overshoot of $\sigma_\mathrm{sc}$ at $\SI{300}{\milli\second}$ in (c) is further amplified by resolving collision avoidance in the nullspace of the singularity-avoidance Jacobian.}
					\akr~enables feasible reactions by avoiding type-II singularities and self-collisions.
					Corresponding measured contact forces are shown in Fig.~\ref{fig:180_plot_kollision_klemmung_rr_Fc_red}.}
					\label{fig:170_plot_kollision_klemmung_rr_iec_red}
					\vspace{-1.5mm}
				\end{figure}
				\begin{figure*}[t!]
					\vspace{1.5mm}			
					\centering
					\includegraphics[width=1\textwidth]{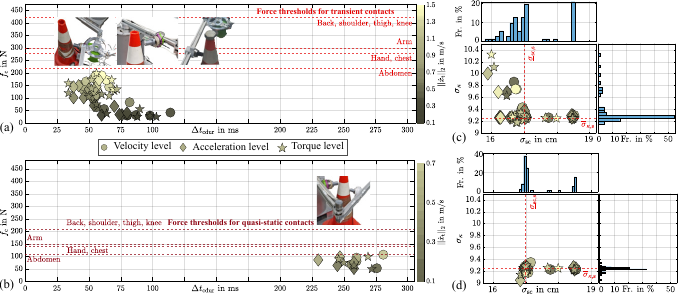}
					\caption{Maximum measured contact forces~$f_\mr{c}$ over contact duration~$\Delta t_\mr{cdur}$ with maximum platform speed in color-code in (a) collision and (b) clamping experiments. Red dotted lines show permissible (a) transient and (b) quasi-static contact force thresholds from ISO/TS 15066, depending on the human body region. 
					Corresponding (c) maximum/minimum and (d) final condition numbers and self-collision distances.
					The use of \akr~leads to feasible reactions with contact forces lower than the thresholds from~\cite{InternationalOrganizationforStandardization.2016}.}
					\label{fig:200_plot_all_collisions_clamping_red}
					\vspace{-1.5mm}
				\end{figure*}
				
				Algorithm~\ref{alg:Detection_Reaction_Process} shows the implementation of the detection-reaction sequence.
				The inputs are the current joint angles~$\bs{q}$, the platform pose~$\bs{x}$, their time derivatives~$\dot{\bs{q}}$ and~$\dot{\bs{x}}$ from~(\ref{eq:DifKin_Jac1}) and~(\ref{eq:DifKin_Jac2}), the estimated forces~$\hat{\bs{F}}_\mr{ext,mP}$ by the MO in~(\ref{eq:mo}) and the modeled gravitational terms~$\hat{\bs{g}}_{q_\mr{a}}$ from~(\ref{eq:dyn_qa}). 	
				To avoid dangerous consequences resulting from false classification regarding contact type or location, a set of thresholds~$\bs{\epsilon}_\mr{g}$ is defined. 
				As soon as~$|\hat{F}_{\mr{ext,mP},i}|{>}\epsilon_{\mr{g},i}$,~$\bs{\tau}_\mr{a,d}{=}\hat{\bs{g}}_{q_\mr{a}}$ is chosen in line~\ref{alg:Detection_Reaction_Process_zeroG}, assuming that a previously initiated reaction is dangerous~\cite{Mohammad_2023_UncQuant}.
				Thresholds~$\bs{\epsilon}_\mr{r}{<}\bs{\epsilon}_\mr{g}$ for contact reaction are defined and if only~$\epsilon_{\mr{r},i}$ is exceeded, the retraction movement or structure opening will be initiated.
				If a clamping contact at the~$j$-th kinematic chain is predicted~(lines~\ref{alg:Detection_Reaction_Process_clamp_start}--\ref{alg:Detection_Reaction_Process_clamp_end}), the corresponding active and passive joints' angles are considered for an opening of the clamping gap.
				In the case of a classified collision, the affected contact body is predicted in line~\ref{alg:Detection_Reaction_Process_collision_body_predicted}, and then one of the three reactions in lines~\ref{alg:Detection_Reaction_Process_collision_body_platform}--\ref{alg:Detection_Reaction_Process_collision_body_2link} is carried out.
				Since the IECs have the highest priorities, the results of the formulations in~(\ref{eq:react_Coll_mobPl}) and~(\ref{eq:react_Coll_mobPl_1D}) are similar, as well as the results of using~(\ref{eq:react_Coll_2Link}) and~(\ref{eq:react_Coll_2Link_1D}).
				In line~\ref{alg:Detection_Reaction_Process_collision_body_2link}, the first estimation of the particle filter is chosen since it is accurate enough for the subsequent reaction. 
				Line~\ref{alg:Detection_Reaction_Process_smooth_trajectory} forms a smooth reaction trajectory\footnote{Jerk-limited trajectory consisting of trapezoidal acceleration profiles} for the task coordinates.
				The trajectories are processed in the PR's control scheme in Alg.~\ref{alg:mode_selection}, which depends on the IECs of singularity and self-collision avoidance.
			\subsubsection{Detailed Evaluation of Collision and Clamping Experiments}\label{ssec:Single_CollCLampingExperiments}
				Figures~\ref{fig:180_plot_kollision_klemmung_rr_Fc_red} and~\ref{fig:170_plot_kollision_klemmung_rr_iec_red} show the measured forces~$f_\mr{c}$ of collision and clamping experiments, as well as the corresponding condition number~$\sigma_\kappa$ and the self-collision distance~$\sigma_\mr{sc}$ with platform speeds in the range of~$0.9$--$\SI{1.4}{\meter / \second}$ (during contact) on the PR's entire structure.
				The redundancy-resolution approach is varied by selecting one of the formulations at velocity, acceleration or torque level.
				In Fig.~\ref{fig:180_plot_kollision_klemmung_rr_Fc_red}, the blue bars represent the detection and reaction's start.
				At this time step, the complete Algorithm~\ref{alg:Detection_Reaction_Process} is executed.
				Figure~\ref{fig:180_plot_kollision_klemmung_rr_Fc_red}(a) shows that the contact occurs at the time~$\SI{0}{\milli \second}$ and is detected within approximately~$\SI{10}{\milli \second}$ in all three cases. 
				As a result of the reaction, the contact is removed after a further~$\SI{40}{\milli \second}$ \highlightred{in Fig.~\ref{fig:180_plot_kollision_klemmung_rr_Fc_red}(a)}. 
				Maximum forces of~$\SI{200}{\newton}$ are measured during the contact period.
				
				Figure~\ref{fig:170_plot_kollision_klemmung_rr_iec_red}(a) shows the previous and further development after the contact is removed.
				The retraction movement of the mobile platform leads to the IEC for singularity avoidance being activated at~$\SI{300}{\milli \second}$, which is considered the highest priority and successfully maintained.
				
				However, this leads to activation of the IEC for self-collision avoidance via~$\sigma_\mr{sc}$ in~$330$--$\SI{400}{\milli \second}$.
				Afterward, both IEC tasks are now active and taken into account as equation constraints in the redundancy resolution. 
				This causes the approach at the acceleration level to overshoot (see~$\mathcolor{ddq}{\lozenge}$,~$\sigma_\mr{sc}$ in Fig.~\ref{fig:170_plot_kollision_klemmung_rr_iec_red}(a) in the period~$400$--$\SI{500}{\milli\second}$), which results from the dynamic movement and the linear interpolation in Sec.~\ref{ssec:IEC_Stetigkeit} to maintain the continuity of the control law.
				However, both IECs are maintained afterward. 
				The previous considerations also apply to the results of the collisions at the first and second links in Figures~\ref{fig:180_plot_kollision_klemmung_rr_Fc_red}(b)--(c) and~\ref{fig:170_plot_kollision_klemmung_rr_iec_red}(b)--(c), so that contacts are detected and canceled within~$30$--$\SI{60}{\milli \second}$.
				The IECs are activated and successfully maintained.
				
				\highlightred{Figure~\ref{fig:180_plot_kollision_klemmung_rr_Fc_red}(d) shows that during the clamping experiment, the contact force temporarily is close to zero at around $\SI{100}{\milli\second}$.
				The later increase of the force indicates an incomplete release, and the full clamping duration extends longer ($\SI{300}{\milli\second}$).				
				It cannot be guaranteed that the duration of this temporary relief is sufficient for a human to withdraw the affected body part in time.}
				The results of the IECs during a clamping reaction are described, as shown in Fig.~\ref{fig:170_plot_kollision_klemmung_rr_iec_red}(d).
				Compared to the collision tests, only the IEC of singularity avoidance is activated here without the risk of self-collision.
				The reason is that the reaction with~$\bs{\tau}_{\mr{a,d}}$ to maintain~$\overline{\sigma}_{\kappa,\mr{s}}$ with simultaneous gap opening with~$\sigma_\mr{r1},\sigma_\mr{r2}$ does not influence~$\sigma_\mr{sc}$, which is recognizable from the constant time course in Fig.~\ref{fig:170_plot_kollision_klemmung_rr_iec_red}(d).
				
				\highlightred{
				A platform-collision experiment is performed, and the CPU load is measured to evaluate the real-time feasibility of the full SafePR pipeline.
				The real-time PC is equipped with an Intel Core i5-6600 CPU (4 cores, 3.3 GHz base frequency, 3.9 GHz max turbo), 8 GB DDR4 RAM, and a 64-bit Ubuntu 20.04 operating system with a preempt RT patch.
				To ensure compliance with real-time ability, a monitoring mechanism based on the external-mode patch was employed.
				This mechanism continuously monitors deadline misses by checking whether the loop starts more than $\SI{0.5}{\milli\second}$ late and whether the loop execution exceeds the sample time.
				Additionally, execution-time overruns are detected and counted if the control step exceeds the sampling interval.  
				Throughout the experiments, no deadline misses or execution overruns were observed.  
				In parallel, the CPU load was logged and evaluated based on kernel idle time and total runtime using a sampling interval of $\SI{100}{\milli \second}$.  
				The CPU load stays below 30\% during execution with~$\SI{1}{\kilo\hertz}$ the SafePR pipeline.}
				
				
				The previous results show that the explicit formulation of clamping joints and collision points in the redundancy resolution based on the obtained contact information \textit{cancels collision and clamping contacts, which confirms contribution~\ref{contribution:Detection_Reaction}}.
				The results of singularity self-collision avoidance also underline \textit{contribution~\ref{contribution:RedRes}}.
			\subsubsection{Statistical Evaluation of Collision and Clamping Experiments}\label{ssec:Multiple_CollCLampingExperiments}
				A systematic evaluation of 72 collision and clamping experiments now follows, varying the redundancy resolution's level, the contact body and the end-effector speed.
				Figure~\ref{fig:200_plot_all_collisions_clamping_red}(a)--(b) shows the results with the maximum measured contact force~$f_\mr{c}$ over the time~$\Delta t_\mr{cdur}$ between contact occurrence and removal determined via the FTS' measurements.
				In addition, maximum permissible force limits, according to~\cite{InternationalOrganizationforStandardization.2016}, for transient and quasi-stationary contacts in different regions of the human body are shown in order to assess the results of the reactions.
				
				Figure~\ref{fig:200_plot_all_collisions_clamping_red}(a) depicts that all collisions are detected and canceled within~$25$--$\SI{125}{\milli \second}$, which is a quarter of the permissible duration of~$\SI{500}{\milli\second}$ for transient contacts~\cite{InternationalOrganizationforStandardization.2016}.
				The color codes the maximum end-effector speeds during the contact duration in the range of~$0.1$--$\SI{1.5}{\meter / \second}$.
				This shows that collisions with higher speeds have a lower~$\Delta t_\mr{cdur}$ and are therefore detected and canceled faster.
				Figure~\ref{fig:200_plot_all_collisions_clamping_red}(b) presents the results of the clamping contacts.
				The permissible forces for quasi-stationary contacts are chosen since a retraction of the fixed pylon is not possible.
				Compared to the previous collision results, it is noticeable that the detection and removal of the clamping contacts require up to~$\SI{300}{\milli \second}$.
				During the collisions and clamping experiments, all the force limits shown for transient and quasi-stationary contacts at the different body regions are maintained.
				The maximum and minimum values of the tasks~$\sigma_\kappa$ and~$\sigma_\mr{sc}$ are depicted in Fig.~\ref{fig:200_plot_all_collisions_clamping_red}(c), which are caused by the high velocities and the interpolation between the two modes.
				However, these situations are not safety-critical because of the chosen values for~$\overline{\sigma}_{\kappa,\mr{s}}$ and~$\underline{\sigma}_{\mr{sc,s}}$.
				Finally, Fig.~\ref{fig:200_plot_all_collisions_clamping_red}(d) shows that the steady-state final values comply with the limits~$\overline{\sigma}_{\kappa,\mr{s}}$ and~$\underline{\sigma}_{\mr{sc},\mr{s}}$ in the majority of the experiments. 
				\begin{figure}[tb!]
					\vspace{1.5mm}
					\centering
					\includegraphics[width=\columnwidth]{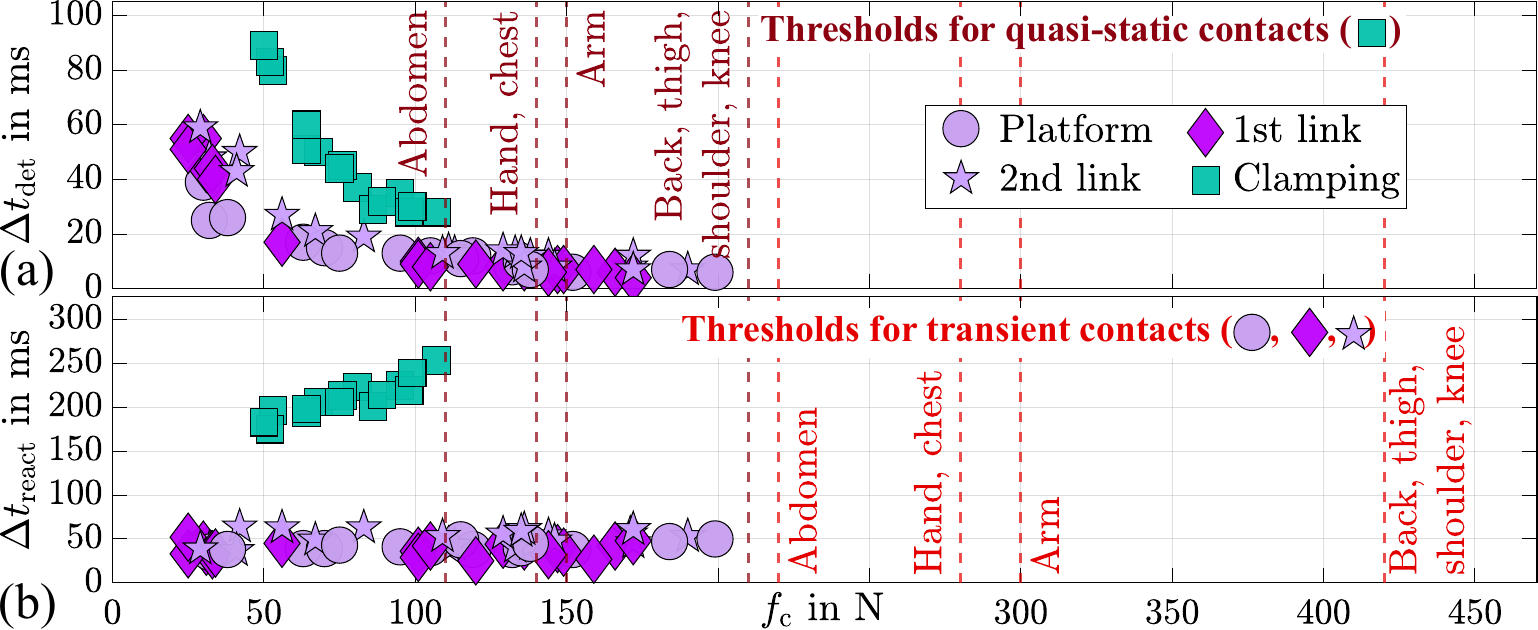}
					\caption{(a) Detection~$\Delta t_\mr{det}$ and (b) reaction duration~$\Delta t_\mr{react}$ over maximum measured contact forces $f_\mr{c}$ during the collision and clamping experiments. Vertical lines show transient and quasi-static contacts' permissible force thresholds from ISO/TS 15066, which are higher than the measured forces and highlight the potential of PRs for HRC.}
					\label{fig:220_plot_tdet_treact_fmax}
					\vspace{-1.5mm}
				\end{figure}
				
				Figure~\ref{fig:220_plot_tdet_treact_fmax} provides a more detailed insight into the results to analyze the duration~$\Delta t_\mr{det}$ between contact occurrence according to the FTS' measurements and detection performed by MO, as well as the duration~$\Delta t_\mr{react}$ between the detection and the contact removal~(FTS' measurement equals~${\approx}\SI{0}{\newton}$).
				In Fig.~\ref{fig:220_plot_tdet_treact_fmax}(a), it is noticeable that collision and clamping contacts are detected faster with increasing maximum force.
				This can be explained by the faster increase in contact force, which affects the actuators and mobile platform in a configuration-dependent way. 
				The effects on the platform are estimated by the generalized-momentum observer, ultimately leading to contact detection.
				Again, classification, localization and isolation are performed at the same time step as the detection in all~72 experiments, and only the measured joint angles and motor-current measurements are used.
				The reaction results in Fig.~\ref{fig:220_plot_tdet_treact_fmax}(b) show that collisions at the platform, at the first and second link, are canceled in~$50$--$\SI{75}{\milli \second}$ across all experiments.
				The clamping experiments in Fig.~\ref{fig:220_plot_tdet_treact_fmax}(b) show a different effect: With increasing maximum contact force, the reaction requires more time to release the clamping.
				A possible cause of this observation is the greater penetration, which increases with higher speed and must be removed by the reaction. 

				\begin{figure}[t!]
					\vspace{1.5mm}
					\centering
					\includegraphics[width=\columnwidth]{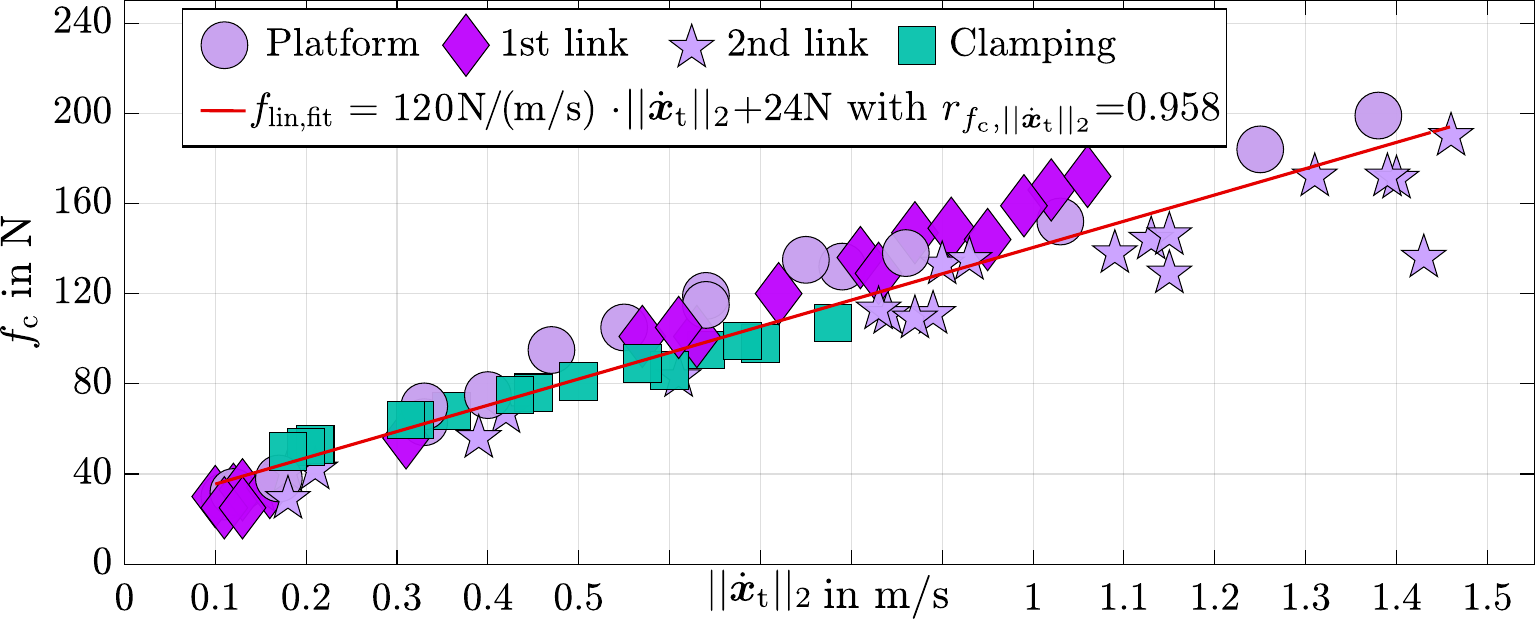}
					\caption{Maximum contact forces over platform speed during the collision and clamping phases. The red line corresponds to a linear fit and enables an estimation of maximum platform speeds.}
					\label{fig:210_plot_fmax_dxmax}
					\vspace{-1.5mm}
				\end{figure}
				\begin{table}[t!]
				\vspace{1.5mm}
				\caption{Pearson correlation coefficient and identified linear parameters for velocity-force model}
				\label{table:identified_contacts}
				\centering
				\begin{tabular}{l||c|c|c|c}
					\hline
					\cellcolor{gray!0} & \cellcolor{platform!0}Platform& \cellcolor{1link!0}1st link & \cellcolor{2link!0}2nd link& \cellcolor{clamping!0}Clamping \\
					\cellcolor{gray!0} &\cellcolor{1link!0}$\mathcolor{platform2}{\bs{\CIRCLE}}$& \cellcolor{1link!0}$\mathcolor{1link2}{\bs{\blacklozenge}}$&
					\cellcolor{2link!0}$\mathcolor{2link2}{\bs{\bigstar}}$& \cellcolor{clamping!0}$\mathcolor{clamping2}{\bs{\blacksquare}}$ \\\hline \hline
					\cellcolor{gray!0}$r_{f_\mr{c},||\dot{\bs{x}}_{\mr{t}}||_2}$ & \cellcolor{gray!0} $0.989$ & \cellcolor{gray!0} $0.997$ & \cellcolor{gray!0} $0.973$ & \cellcolor{gray!0} $0.997$\\\hline			
					\cellcolor{gray!10}$\frac{\Delta f_\mr{c}}{\Delta ||\dot{\bs{x}}_{\mr{t}}||_2}$ in $\frac{\mr{N}}{\mr{m}/\mr{s}}$& \cellcolor{gray!10} $132$ & \cellcolor{gray!10} $150$ & \cellcolor{gray!10} $108$ & \cellcolor{gray!10} $94$\\\hline			
					\cellcolor{gray!0}$f_0$ in $\mr{N}$ & \cellcolor{gray!0} $23$ & \cellcolor{gray!0} $11$ & \cellcolor{gray!0} $19$ & \cellcolor{gray!0} $34$\\\hline			
					\cellcolor{gray!10}$||\dot{\bs{x}}_{\mr{t}}||_2$ for $f_\mr{c}{=}\SI{280}{\newton}$& \cellcolor{gray!10} $1.95$ & \cellcolor{gray!10} $1.79$ & \cellcolor{gray!10} $2.42$ & \cellcolor{gray!10} $2.62$\\ \hline			
					\cellcolor{gray!0}$||\dot{\bs{x}}_{\mr{t}}||_2$ for $f_\mr{c}{=}\SI{140}{\newton}$& \cellcolor{gray!0} - & \cellcolor{gray!0} - & \cellcolor{gray!0} - & \cellcolor{gray!0} $0.97$\\ \hline			
				\end{tabular}
				\end{table}
				
				The collision and clamping results are analyzed regarding the maximum contact forces and platform speeds in Fig.~\ref{fig:210_plot_fmax_dxmax} and Table~\ref{table:identified_contacts}.
				The Pearson correlation coefficient\footnotemark\footnotetext{For a linear relationship between two variables, the coefficient is equal to one.}~$r_{f_\mr{c},||\dot{\bs{x}}_{\mr{t}}||_2}$ and parameters of linear models~$f_{\mr{c},i}(||\dot{\bs{x}}_{\mr{t}}||_2)$ are identified for the contacts in each case and in total.
				Based on the correlation coefficients of $0.958$--$0.997$, linear models are assumed with slopes in the range of~$94$--$\SI{150}{N/(m/s)}$.				
				Using the linear models, maximum platform speeds are estimated based on the force thresholds of~\cite{InternationalOrganizationforStandardization.2016 }, which are exemplary for transient contact with the human hand with $f_{\mr{c}}{=}\SI{280}{\newton}$~($\SI{140}{\newton}$ for quasi-static scenarios).
				Table~\ref{table:identified_contacts} lists the estimated platform speeds, which range from~$1$--$\SI{2.6}{\meter /\second}$.
				Considering the observations from Fig.~\ref{fig:200_plot_all_collisions_clamping_red} and~\ref{fig:220_plot_tdet_treact_fmax} that collisions with higher platform speeds are detected and canceled faster, we assume that these theoretical contacts are also eliminated within less than $\SI{100}{\milli \second}$~($\SI{300}{\milli \second}$ for clamping contacts) using the presented methods.
				However, this assumption is limited by the different material parameters of the contact partners since the pylon's effective mass and stiffness differ from the human hand.
				\highlightred{Another limitation is that the identified linear force-velocity model is based on data from orthogonal contacts, excluding frictional and angled collisions.
				Nevertheless, the linear relationship is known and relevant in the state-of-the-art, as detailed in ISO/TS 15066~\cite{InternationalOrganizationforStandardization.2016}.
				The effective stiffness and the mass of both the robot and the affected body region are relevant factors here.}
				
				Also, the selection of the presented clamping scenarios as quasi-static contacts is debatable: Although the pylon is fixed, in a realistic clamping scenario a retraction of the affected body part in the opposite direction to the clamping joint of the PR would be possible. 
				In this case, the clamping, like the collision, would have to be regarded as a transient contact with its higher force threshold so that velocities in the range of~$1.8$--$\SI{2.6}{\meter /\second}$ would be achievable.
			
	\section{\highlightred{Contributions and Limitations of SafePR}}\label{sec:Contributions_limitations}
		The presented results show that parallel robots provide the potential for HRC to achieve high platform speeds with a safety level in accordance with the force thresholds. 
		Contribution~\ref{contribution:Reactions_RedRes_Results} is proved since the considered contact scenarios are (i) detected and eliminated within $\SI{300}{\milli \second}$ for higher speeds and (ii) show the PR's potential for safe HRC according to the technical specification~\cite{InternationalOrganizationforStandardization.2016}, (iii) while the inherent limitations of parallel robotics, type-II singularities and self-collisions, are successfully avoided.
		The presented contact-detection and reaction methods are based only on built-in sensors, which represent advantages in terms of cost and robustness compared to sensors with exteroceptive information.\\
		\highlightred{When including (still) non-standard sensors like IMUs, the detection time of $6$--$\SI{20}{\milli\second}$ can be reduced by another $\SI{3}{\milli\second}$~\cite{Mohammad.2025}.
		This approach also requires mounting the IMU and incorporating sensor fusion, but it can also be integrated into \akr~to reduce detection time.
		Other parts of \akr~would remain unaffected.\\}
		\highlightred{For the application of \akr~on six-DoF parallel robots, datasets for supervised-learning algorithms must be collected for classification, which is likely more complex due to the larger dimensionality of the configuration and parameter space.
		Additionally, applying the kinematics and dynamics modeling to the six-DoF case is necessary.
		Also, if the drives are equipped with gears, the effects of backlash and gear friction can be addressed in \akr~by tuning observer thresholds and incorporating uncertainty into an interval-based contact-force estimation.
		Furthermore, the moments $\bs{m}_\mr{ext,mP}$ in the minimum-distance vector $\bs{r}_\mr{mP,LoA}$ from (\ref{eq:CalcMPLoA}) would have to be formed via the geometric Jacobian matrix.
		The reaction methods are applicable to six-DoF parallel robots due to the generally formulated nullspace projections.
		Similarly, the reactions to avoid singularities and self-collisions are based on the calculation of the Jacobian matrices $\bs{J}_\kappa$ and $\bs{J}_\mr{sc}$, for which Algorithm~\ref{alg:Jacobian} would still be applicable.}\\
		\highlightred{
			\akr~has the following limitations regarding the soft fulfillment of constraints and single-contact scenarios.\\
			(i) Strictly fulfilling the inequality constraints is not possible due to mode-transition interpolation for the control law's continuity.
			One solution is to perform model-predictive trajectory planning within an optimization scheme that enables switching modes at the time of detection and reaction initiation.
			The model predictive control could use the direct dynamics in~(\ref{eq:dyn_sigma1}) expressed in the coordinates of the inequality constraints to ensure strict compliance with the constraints.\\
			(ii) \akr~can only handle single contacts, since the planar parallel robot has only three current-measurement sensors.
			In a multi-contact scenario, the momentum observer estimates the combined effect of the contact forces on the drives according to their contact Jacobian matrices.
			Consequently, it would be possible to detect contact as long as the forces do not balance out completely.
			However, the subsequent steps required to determine the contact type and location would not be applicable, as these classifiers have only been trained using single-contact datasets, and the required information would not be available in the multi-contact case.\\
			\akr~can be extended to handle multiple simultaneous contacts if the robot would be equipped with additional force-torque sensors at the base joints.
			But since the proposed framework is designed to maximize applicability to other parallel robots, a minimized sensor setup is beneficial for reduced cost and system complexity.			
		}
				

	\section{Conclusion} \label{sec:conlusions}
		This work presents \akr~--- a detection-reaction approach for parallel robots that enables safe physical interaction while also being capable of high-speed operations due to their lower moving masses.		
		Based on a generalized-momentum observer, contacts with end-effector speeds of up to~$\SI{1.5}{\meter / \second}$ are detected within $10$--$\SI{100}{\milli \second}$.
		Interestingly, faster robot movements led to shorter detection times and, finally, to sooner safe reactions in the experiments, which underlines the parallel robot's potential for highly dynamic physical interaction with humans. 
		
		Only joint angles and motor currents are used as input into the real-time capable kinematics and dynamics modeling at~$\SI{1}{\kilo \hertz}$.
		Physically modeled features enable neural networks to classify contact type and location with accuracies in the range of~$80$--$95\%$.
		Using a fallback reaction in dependence on the estimated force prevents dangerous consequences resulting from false classification.
		These accuracies are achieved even in joint-angle configurations of the parallel robot that were unknown during training. 
		This is an essential step towards more reliable data-driven modeling in real-world applications.
		
		The previously gathered contact information is used for an immediate reaction strategy that explicitly addresses the collided or clamped robot body. 
		The retraction movements and opening of clamping chains cancel collisions within~$\SI{50}{\milli \second}$ and clamping contacts within~$\SI{250}{\milli \second}$. 
		This results in measured contact forces lower than the transient and quasi-static contacts' thresholds mentioned in ISO/TS~15066.
		Inherent limitations of parallel robots are successfully incorporated in these reaction strategies to avoid singularities and self-collisions.
		
		In 72 real-world collision and clamping experiments with one object, we demonstrate the safe and feasible detection and reaction capability with a planar parallel robot.
		Collisions and clamping contacts are removed within less than~$\SI{300}{\milli \second}$ while reaching end-effector speeds of up to~$\SI{1.5}{\meter / \second}$ in the pre-detection phase.
		As soon as contacts are detected, the type classification and localization are performed in the same time step (one millisecond) based on physically modeled features, and the reaction is initiated.
		
		These results underline this work's contribution and show that \akr~enables the usage of parallel robots in physical human-robot interaction in real-world applications due to the introduced safe and feasible reactions.
		Our methods rely only on standard built-in sensors like encoders and motor-current measurements, which benefits the long-term commercial utilization.
		Future research should focus on the transfer and demonstration of the detection-reaction sequence to spatial six-DoF parallel robots with gear friction affecting the motor-current-based force estimation and using diverse contact objects to investigate the higher velocities proposed in this work.
		\highlightred{Also, a model predictive optimization is a promising approach for future work to ensure strict compliance with the inequality constraints.}

	\addtolength{\textheight}{0cm} 
	\bibliographystyle{IEEEtran}
	\bibliography{literatur}	

@article{AbhishekAgarwal.2016,
 author = {Agarwal, Abhishek and Nasa, Chaman and Bandyopadhyay, Sandipan},
 year = {2016},
 title = {Dynamic singularity avoidance for parallel manipulators using a task-priority based control scheme},
 pages = {107--126},
 volume = {96},
 issn = {0094114X},
 journal = {Mechanism and Machine Theory},
 doi = {10.1016/j.mechmachtheory.2015.07.013}
}

@inproceedings{Albini.2017,
 author = {Albini, Alessandro and Denei, Simone and Cannata, Giorgio},
 title = {Human hand recognition from robotic skin measurements in human-robot physical interactions},
 pages = {4348--4353},
 booktitle = {2017 IEEE/RSJ International Conference on Intelligent Robots and Systems (IROS)},
 year = {2017},
 doi = {10.1109/IROS.2017.8206300}
}

@inproceedings{AlbuSchaffer.2003,
 author = {Albu-Sch{\"a}ffer, A. and Ott, C. and Frese, U. and Hirzinger, G.},
 title = {Cartesian impedance control of redundant robots: recent results with the {DLR}-light-weight-arms},
 pages = {3704--3709},
 volume = {3},
 booktitle = {ICRA},
 year = {2003},
 doi = {10.1109/ROBOT.2003.1242165}
}

@ARTICLE{AlbuSchaeffer.2023,
	author={Albu-Schäffer, Alin and Sachtler, Arne},
	journal={TRO}, 
	title={Redundancy Resolution at Position Level}, 
	year={2023},
	volume={39},
	number={6},
	pages={4240-4261},
	doi={10.1109/TRO.2023.3309097}}

@article{AlexanderDietrich.2015,
 author = {Dietrich, Alexander and Ott, Christian and Albu-Sch{\"a}ffer, Alin},
 year = {2015},
 title = {An overview of null space projections for redundant, torque-controlled robots},
 pages = {1385--1400},
 volume = {34},
 number = {11},
 issn = {0278-3649},
 journal = {IJRR},
 doi = {10.1177/0278364914566516}
}

@inproceedings{Antonelli.2009b,
 author = {Antonelli, Gianluca and Indiveri, Giovanni and Chiaverini, Stefano},
 title = {Prioritized closed-loop inverse kinematic algorithms for redundant robotic systems with velocity saturations},
 pages = {5892--5897},
 booktitle = {IROS},
 year = {2009},
 doi = {10.1109/IROS.2009.5354636}
}

@inproceedings{Arrichiello.2017,
 author = {Arrichiello, Filippo and {Di Lillo}, Paolo and {Di Vito}, Daniele and Antonelli, Gianluca and Chiaverini, Stefano},
 title = {Assistive robot operated via P300-based brain computer interface},
 pages = {6032--6037},
 booktitle = {ICRA},
 year = {2017},
 doi = {10.1109/ICRA.2017.7989714}
}

@article{CavacantiSantos.2017,
 author = {Santos, Jo{\~a}o Cavacanti and Silva, Ma{\'i}ra Da Martins},
 year = {2017},
 title = {Redundancy Resolution of Kinematically Redundant Parallel Manipulators Via Differential Dynamic Programing},
 volume = {9},
 number = {4},
 issn = {1942-4302},
 journal = {Journal of Mechanisms and Robotics},
 doi = {10.1115/1.4036739}
}

@inproceedings{Cha.2007,
 author = {Cha, Sung-Hoon and Lasky, Ty A. and Velinsky, Steven A.},
 title = {Singularity Avoidance for the 3-{RRR} Mechanism Using Kinematic Redundancy},
 pages = {1195--1200},
 booktitle = {ICRA},
 year = {2007},
 doi = {10.1109/ROBOT.2007.363147}
}

@article{Dahiya.2013,
 author = {Dahiya, Ravinder S. and Mittendorfer, Philipp and Valle, Maurizio and Cheng, Gordon and Lumelsky, Vladimir J.},
 year = {2013},
 title = {Directions Toward Effective Utilization of Tactile Skin: A Review},
 pages = {4121--4138},
 volume = {13},
 number = {11},
 issn = {1530-437X},
 journal = {IEEE Sensors Journal},
 doi = {10.1109/JSEN.2013.2279056}
}

@inproceedings{Dehio.2019,
 author = {Dehio, Niels and Steil, Jochen J.},
 title = {Dynamically-consistent Generalized Hierarchical Control},
 pages = {1141--1147},
 booktitle = {ICRA},
 year = {2019},
 doi = {10.1109/ICRA.2019.8793553}
}

@inproceedings{Dietrich.2011,
 author = {Dietrich, Alexander and Wimb{\"o}ck, Thomas and Albu-Sch{\"a}ffer, Alin},
 title = {Dynamic whole-body mobile manipulation with a torque controlled humanoid robot via impedance control laws},
 pages = {3199--3206},
 booktitle = {IROS},
 year = {2011},
 doi = {10.1109/IROS.2011.6094445}
}

@article{Dietrich.2012,
 author = {Dietrich, Alexander and Wimb{\"o}ck, Thomas and Albu-Sch{\"a}ffer, Alin and Hirzinger, Gerd},
 year = {2012},
 title = {Integration of Reactive, Torque-Based Self-Collision Avoidance Into a Task Hierarchy},
 pages = {1278--1293},
 volume = {28},
 number = {6},
 issn = {1552-3098},
 journal = {TRO},
 doi = {10.1109/TRO.2012.2208667}
}

@ARTICLE{Dietrich.2020,
	author={Dietrich, Alexander and Ott, Christian},
	journal={TRO}, 
	title={Hierarchical Impedance-Based Tracking Control of Kinematically Redundant Robots}, 
	year={2020},
	volume={36},
	number={1},
	pages={204-221},
	doi={10.1109/TRO.2019.2945876}}

@article{Dutta.2019,
 author = {Dutta, Anirvan and Salunkhe, Durgesh Haribhau and Kumar, Shivesh and Udai, Arun Dayal and Shah, Suril V.},
 year = {2019},
 title = {Sensorless full body active compliance in a 6-{DOF} parallel manipulator},
 pages = {278--290},
 volume = {59},
 issn = {0736-5845},
 journal = {Robotics and Computer-Integrated Manufacturing},
 doi = {10.1016/j.rcim.2019.04.010}
}

@article{Escarabajal.2023,
 author = {Escarabajal, Rafael J. and Pulloquinga, Jos{\'e} L. and Valera, {\'A}ngel and Mata, Vicente and Vall{\'e}s, Marina and Castillo-Garc{\'i}a, Fernando J.},
 year = {2023},
 title = {Combined Admittance Control With Type II Singularity Evasion for Parallel Robots Using Dynamic Movement Primitives},
 pages = {2224--2239},
 volume = {39},
 number = {3},
 issn = {1552-3098},
 journal = {TRO},
 doi = {10.1109/TRO.2023.3238136}
}

@article{Ficuciello.2015,
 author = {Ficuciello, Fanny and Villani, Luigi and Siciliano, Bruno},
 year = {2015},
 title = {Variable Impedance Control of Redundant Manipulators for Intuitive Human--Robot Physical Interaction},
 pages = {850--863},
 volume = {31},
 number = {4},
 issn = {1552-3098},
 journal = {TRO},
 doi = {10.1109/TRO.2015.2430053}
}

@article{Fiore.2023,
 author = {Fiore, Mario Daniele and Meli, Gaetano and Ziese, Anton and Siciliano, Bruno and Natale, Ciro},
 year = {2023},
 title = {A General Framework for Hierarchical Redundancy Resolution Under Arbitrary Constraints},
 pages = {2468--2487},
 volume = {39},
 number = {3},
 issn = {1552-3098},
 journal = {TRO},
 doi = {10.1109/TRO.2022.3232266}
}

@inproceedings{Flacco.2012,
 author = {Flacco, Fabrizio and de Luca, Alessandro and Khatib, Oussama},
 title = {Motion control of redundant robots under joint constraints: Saturation in the Null Space},
 pages = {285--292},
 booktitle = {ICRA},
 year = {2012},
 doi = {10.1109/ICRA.2012.6225376}
}

@inproceedings{Flacco.2012b,
 author = {Flacco, Fabrizio and de Luca, Alessandro and Khatib, Oussama},
 title = {Prioritized multi-task motion control of redundant robots under hard joint constraints},
 pages = {3970--3977},
 booktitle = {IROS},
 year = {2012},
 doi = {10.1109/IROS.2012.6385619}
}

@inproceedings{Golz.2015,
 author = {Golz, Saskia and Osendorfer, Christian and Haddadin, Sami},
 title = {Using tactile sensation for learning contact knowledge: Discriminate collision from physical interaction},
 pages = {3788--3794},
 isbn = {1050-4729},
 booktitle = {2015 IEEE International Conference on Robotics and Automation (ICRA)},
 year = {2015},
 doi = {10.1109/ICRA.2015.7139726}
}

@article{Gosselin.2016,
 author = {Gosselin, Cl{\'e}ment and Schreiber, Louis-Thomas},
 year = {2016},
 title = {Kinematically Redundant Spatial Parallel Mechanisms for Singularity Avoidance and Large Orientational Workspace},
 pages = {286--300},
 volume = {32},
 number = {2},
 issn = {1552-3098},
 journal = {TRO},
 doi = {10.1109/TRO.2016.2516025}
}

@article{Gosselin.2018,
	title={Redundancy in Parallel Mechanisms: A Review},
	author={Cl{\'e}ment Gosselin and Louis-Thomas Schreiber},
	journal={Applied Mechanics Reviews},
	year={2018},
	volume={70},
	pages={010802}
}

@article{Haddadin.2017,
 author = {Haddadin, Sami and de Luca, Alessandro and Albu-Sch{\"a}ffer, Alin},
 year = {2017},
 title = {Robot Collisions: A Survey on Detection, Isolation, and Identification},
 pages = {1292--1312},
 volume = {33},
 number = {6},
 issn = {1552-3098},
 journal = {IEEE Transactions on Robotics (TRO)},
 doi = {10.1109/TRO.2017.2723903}
}

@article{Hermus.2022,
 author = {Hermus, James and Lachner, Johannes and Verdi, David and Hogan, Neville},
 year = {2022},
 title = {Exploiting Redundancy to Facilitate Physical Interaction},
 pages = {599--615},
 volume = {38},
 number = {1},
 issn = {1552-3098},
 journal = {TRO},
 doi = {10.1109/TRO.2021.3086632}
}

@article{Hoang.2022,
 author = {Hoang, Xuan-Bach and Pham, Phu-Cuong and Kuo, Yong-Lin},
 year = {2022},
 title = {Collision Detection of a {H}exa Parallel Robot Based on Dynamic Model and a Multi-Dual Depth Camera System},
 volume = {22},
 number = {15},
 issn = {1424-8220},
 journal = {Sensors},
 doi = {10.3390/s22155923}
}

@misc{InternationalOrganizationforStandardization.2016,
 author = {{International Organization for Standardization}},
 title = {Robots and robotic devices --- Collaborative robots ({ISO/TS} Standard No. 15066:2016)}
}

@misc{InternationalFederationofRobotics.2024,
	author = {{International Federation of Robotics}},
	title = {Position Paper November 2024 --- Collaborative Robots --- How Robots Work Alongside Humans},
}

@article{JunNakanishi.2008,
 author = {Nakanishi, Jun and Cory, Rick and Mistryl, Michael and Peters, Jan and Schaal, Stefan},
 year = {2008},
 title = {Operational Space Control: A Theoretical and Empirical Comparison},
 pages = {737--757},
 volume = {27},
 number = {6},
 issn = {0278-3649},
 journal = {IJRR},
 doi = {10.1177/0278364908091463}
}

@article{Khatib.1987,
 author = {Khatib, O.},
 year = {1987},
 title = {A unified approach for motion and force control of robot manipulators: The operational space formulation},
 pages = {43--53},
 volume = {3},
 number = {1},
 issn = {2374-8710},
 journal = {IEEE Journal on Robotics and Automation},
 doi = {10.1109/JRA.1987.1087068}
}

@article{Khurana.2024,
 author = {Khurana, Harshit and Billard, Aude},
 year = {2024},
 title = {Motion Planning and Inertia-Based Control for Impact Aware Manipulation},
 pages = {2201--2216},
 volume = {40},
 issn = {1552-3098},
 journal = {TRO},
 doi = {10.1109/TRO.2023.3319895}
}

@inproceedings{Kotlarski.2010,
 author = {Kotlarski, Jens and {Do Thanh}, Trung and Heimann, Bodo and Ortmaier, Tobias},
 title = {Optimization strategies for additional actuators of kinematically redundant parallel kinematic machines},
 pages = {656--661},
 booktitle = {ICRA},
 year = {2010},
 doi = {10.1109/ROBOT.2010.5509982}
}

@article{Lachner.2021,
 author = {Lachner, Johannes and Allmendinger, Felix and Hobert, Eddo and Hogan, Neville and Stramigioli, Stefano},
 year = {2021},
 title = {Energy budgets for coordinate invariant robot control in physical human--robot interaction},
 pages = {968--985},
 volume = {40},
 number = {8-9},
 issn = {0278-3649},
 journal = {The International Journal of Robotics Research (IJRR)},
 doi = {10.1177/02783649211011639}
}

@inproceedings{Lippi.2021,
 author = {Lippi, Martina and Gillini, Giuseppe and Marino, Alessandro and Arrichiello, Filippo},
 title = {A Data-Driven Approach for Contact Detection, Classification and Reaction in Physical Human-Robot Collaboration},
 pages = {3597--3603},
 booktitle = {ICRA},
 year = {2021},
 doi = {10.1109/ICRA48506.2021.9561827}
}

@article{Liu.2016,
 author = {Liu, Mingxing and Tan, Yang and Padois, Vincent},
 year = {2016},
 title = {Generalized hierarchical control},
 pages = {17--31},
 volume = {40},
 number = {1},
 issn = {1573-7527},
 journal = {Autonomous Robots},
 doi = {10.1007/s10514-015-9436-1}
}

@inproceedings{Luca.2006,
 author = {de Luca, Alessandro and Albu-Sch{\"a}ffer, Alin and Haddadin, Sami and Hirzinger, Gerd},
 title = {Collision Detection and Safe Reaction with the {DLR}-{III} Lightweight Manipulator Arm},
 pages = {1623--1630},
 booktitle = {IROS},
 year = {2006},
 doi = {10.1109/IROS.2006.282053}
}

@inproceedings{Luca.2008,
 author = {de Luca, Alessandro and Ferrajoli, Lorenzo},
 title = {Exploiting Robot Redundancy in Collision Detection and Reaction},
 pages = {3299--3305},
 isbn = {978-1-4244-2057-5},
 booktitle = {IROS},
 year = {2008},
 doi = {10.1109/IROS.2008.4651204}
}

@inproceedings{Magrini.2014,
 author = {Magrini, Emanuele and Flacco, Fabrizio and de Luca, Alessandro},
 title = {Estimation of contact forces using a virtual force sensor},
 pages = {2126--2133},
 booktitle = {IROS},
 year = {2014},
 doi = {10.1109/IROS.2014.6942848}
}

@inproceedings{Magrini.2015,
 author = {Magrini, Emanuele and Flacco, Fabrizio and de Luca, Alessandro},
 title = {Control of generalized contact motion and force in physical human-robot interaction},
 pages = {2298--2304},
 isbn = {978-1-4799-6923-4},
 booktitle = {ICRA},
 year = {2015},
 doi = {10.1109/ICRA.2015.7139504}
}

@inproceedings{Mansfeld.2017,
 author = {Mansfeld, Nico and Djellab, Badis and Veuthey, Jaime Raldua and Beck, Fabian and Ott, Christian and Haddadin, Sami},
 title = {Improving the performance of biomechanically safe velocity control for redundant robots through reflected mass minimization},
 pages = {5390--5397},
 booktitle = {IROS},
 year = {2017},
 doi = {10.1109/IROS.2017.8206435}
}

@inproceedings{Meguenani.2015,
 author = {Meguenani, Anis and Padois, Vincent and Bidaud, Philippe},
 title = {Control of robots sharing their workspace with humans: An energetic approach to safety},
 pages = {4678--4684},
 booktitle = {IROS},
 year = {2015},
 doi = {10.1109/IROS.2015.7354043}
}

@book{Merlet.2006,
 author = {Merlet, J-P},
 year = {2006},
 title = {Parallel robots},
 edition = {2nd ed.},
 volume = {74},
 publisher = {Springer},
 isbn = {1402041322},
 series = {Solid mechanics and its applications}
}

@article{Merlet.2006_JMD,
	author = {Merlet, J. P.},
	title = {Jacobian, Manipulability, Condition Number, and Accuracy of Parallel Robots},
	journal = {Journal of Mechanical Design},
	volume = {128},
	number = {1},
	pages = {199-206},
	year = {2005},
	issn = {1050-0472},
	doi = {10.1115/1.2121740},
}

@article{Moe.2016,
 author = {Moe, Signe and Antonelli, Gianluca and Teel, Andrew R. and Pettersen, Kristin Y. and Schrimpf, Johannes},
 year = {2016},
 title = {Set-Based Tasks within the Singularity-Robust Multiple Task-Priority Inverse Kinematics Framework: General Formulation, Stability Analysis, and Experimental Results},
 volume = {3},
 issn = {2296-9144},
 journal = {Frontiers in Robotics and AI},
 doi = {10.3389/frobt.2016.00016}
}

@inproceedings{Mohammad.2023,
 author = {Mohammad, Aran and Schappler, Moritz and Ortmaier, Tobias},
 title = {Towards Human-Robot Collaboration with Parallel Robots by Kinetostatic Analysis, Impedance Control and Contact Detection},
 pages = {12092--12098},
 booktitle = {ICRA},
 year = {2023},
 doi = {10.1109/ICRA48891.2023.10161217}
}

@inproceedings{Mohammad.2023_IsolLoc,
 author = {Mohammad, Aran and Schappler, Moritz and Ortmaier, Tobias},
 title = {Collision Isolation and Identification Using Proprioceptive Sensing for Parallel Robots to Enable Human-Robot Collaboration},
 pages = {5910--5917},
 booktitle = {IROS},
 year = {2023},
 doi = {10.1109/IROS55552.2023.10342345}
}

@inproceedings{Mohammad.2023_Reaction,
 author = {Mohammad, Aran and Schappler, Moritz and Habich, Tim-Lukas and Ortmaier, Tobias},
 title = {Safe Collision and Clamping Reaction for Parallel Robots During Human-Robot Collaboration},
 pages = {5966--5973},
 booktitle = {IROS},
 year = {2023},
 doi = {10.1109/IROS55552.2023.10341581}
}

@inproceedings{Mohammad_2023_UncQuant,
 author = {Mohammad, Aran and Muscheid, Hendrik and Schappler, Moritz and Seel, Thomas},
 title = {Quantifying Uncertainties of~Contact Classifications in~a~Human-Robot Collaboration with~Parallel Robots},
 pages = {137--150},
 publisher = {{Springer Nature Switzerland}},
 isbn = {978-3-031-55000-3},
 booktitle = {Human-Friendly Robotics 2023},
 year = {2024},
 address = {Cham}
}

@ARTICLE{Mohammad.2025,
	author={Mohammad, Aran and Piosik, Jan and Lehmann, Dustin and Seel, Thomas and Schappler, Moritz},
	journal={IEEE Robotics and Automation Letters}, 
	title={Fast Contact Detection Via Fusion of Joint and Inertial Sensors for Parallel Robots in Human-Robot Collaboration}, 
	year={2025},
	volume={10},
	number={7},
	pages={7547-7554},
	doi={10.1109/LRA.2025.3575326}}

@inproceedings{MunozOsorio.2019,
 author = {Osorio, Juan D. Mu{\~n}oz and Allmendinger, Felix and Fiore, Mario and Zimmermann, Uwe and Ortmaier, Tobias},
 title = {Physical Human-Robot Interaction under Joint and {C}artesian Constraints},
 pages = {185--191},
 year = {2019},
 booktitle = {19th International Conference on Advanced Robotics},
 doi = {10.1109/ICAR46387.2019.8981579}
}

@inproceedings{Osorio.2020,
 author = {Osorio, Juan D. Mu{\~n}oz and Abdelazim, Abdelrahman and Allmendinger, Felix and Zimmermann, Uwe E.},
 title = {Unilateral Constraints for Torque-based Whole-Body Control},
 pages = {7623--7628},
 booktitle = {IROS},
 year = {2020},
 doi = {10.1109/IROS45743.2020.9341241}
}

@book{Ott.2008,
 author = {Ott, Christian},
 year = {2008},
 title = {Cartesian Impedance Control of Redundant and Flexible-Joint Robots},
 price = {Gb. : EUR 85.55 (freier Pr.), sfr 133.00 (freier Pr.)},
 volume = {49},
 isbn = {978-3-540-69253-9},
 series = {Springer tracts in advanced robotics},
 doi = {10.1007/978-3-540-69255-3}
}

@article{OussamaKhatib.1993,
 author = {Khatib, Oussama},
 year = {1993},
 title = {The Operational Space Framework},
 pages = {277--287},
 volume = {36},
 number = {3},
 journal = {JSME international journal. Ser. C, Dynamics, control, robotics, design and manufacturing},
 doi = {10.1299/jsmec1993.36.277}
}

@InProceedings{Schappler.2021,
	author    = {Schappler, Moritz and Ortmaier, Tobias},
	booktitle = {Proceedings of the 18th International Conference on Informatics in Control, Automation and Robotics},
	title     = {Singularity Avoidance of Task-Redundant Robots in Pointing Tasks: On Nullspace Projection and {C}ardan Angles as Orientation Coordinates},
	year      = {2021},
	doi       = {10.5220/0010621103380349},
}

@inproceedings{Schappler.2023,
 author = {Schappler, Moritz},
 title = {Pose Optimization of~Task-Redundant Robots in~Second-Order Rest-to-Rest Motion with~Cascaded Dynamic Programming and~Nullspace Projection},
 pages = {106--131},
 publisher = {{Springer International Publishing}},
 isbn = {978-3-031-26474-0},
 booktitle = {Informatics in Control, Automation and Robotics},
 year = {2023},
 address = {Cham}
}

@article{Sun.2024,
 author = {Sun, Tao and Sun, Jiarui and Lian, Binbin and Li, Qi},
 year = {2024},
 title = {Sensorless admittance control of 6-{DoF} parallel robot in human-robot collaborative assembly},
 pages = {102742},
 volume = {88},
 issn = {0736-5845},
 journal = {Robotics and Computer-Integrated Manufacturing},
 doi = {10.1016/j.rcim.2024.102742}
}

@inproceedings{Sutjipto.2021,
 author = {Sutjipto, Sheila and Woolfrey, Jon and Carmichael, Marc G. and Paul, Gavin},
 title = {Cartesian Inertia Optimization via Redundancy Resolution for Physical Human-Robot Interaction},
 pages = {570--575},
 booktitle = {IEEE 17th International Conference on Automation Science and Engineering (CASE)},
 year = {2021},
 doi = {10.1109/CASE49439.2021.9551663}
}

@book{Taghirad.2013,
 author = {Taghirad, Hamid D.},
 year = {2013},
 title = {Parallel Robots: Mechanics and control},
 address = {Boca Raton, FL},
 edition = {1st edition},
 publisher = {{CRC Press}},
 isbn = {9780429097454},
 doi = {10.1201/b16096}
}

@inproceedings{Thanh.2009,
 author = {Thanh, Trung Do and Kotlarski, Jens and Heimann, Bodo and Ortmaier, Tobias},
 title = {On the inverse dynamics problem of general parallel robots},
 pages = {1--6},
 booktitle = {IEEE International Conference on Mechatronics},
 year = {2009},
 doi = {10.1109/ICMECH.2009.4957202}
}

@article{Thanh.2012,
 author = {Thanh, Trung Do and Kotlarski, Jens and Heimann, Bodo and Ortmaier, Tobias},
 year = {2012},
 title = {Dynamics identification of kinematically redundant parallel robots using the direct search method},
 pages = {277--295},
 volume = {52},
 issn = {0094114X},
 journal = {Mechanism and Machine Theory},
 doi = {10.1016/j.mechmachtheory.2012.02.002}
}

@inproceedings{Todorov.2012,
 author = {Todorov, Emanuel and Erez, Tom and Tassa, Yuval},
 title = {{MuJoCo}: A physics engine for model-based control},
 pages = {5026--5033},
 isbn = {2153-0866},
 booktitle = {IROS},
 year = {2012},
 doi = {10.1109/IROS.2012.6386109}
}

@ARTICLE{Vorndamme.2024,
	author={Vorndamme, Jonathan and Melone, Alessandro and Kirschner, Robin and Figueredo, Luis and Haddadin, Sami},
	journal={TRO}, 
	title={Safe Robot Reflexes: A Taxonomy-based Decision and Modulation Framework}, 
	year={2024},
	volume={},
	number={},
	pages={1-20},
	doi={10.1109/TRO.2024.3519421}}

@article{Walker.1994,
 author = {Walker, I. D.},
 year = {1994},
 title = {Impact configurations and measures for kinematically redundant and multiple armed robot systems},
 pages = {670--683},
 volume = {10},
 number = {5},
 journal = {IEEE Transactions on Robotics and Automation},
 doi = {10.1109/70.326571}
}

@article{Wen.2021,
 author = {Wen, Kefei and Nguyen, Tan Sy and Harton, David and Lalibert{\'e}, Thierry and Gosselin, Cl{\'e}ment},
 year = {2021},
 title = {A Backdrivable Kinematically Redundant (6+3)-Degree-of-Freedom Hybrid Parallel Robot for Intuitive Sensorless Physical Human--Robot Interaction},
 pages = {1222--1238},
 volume = {37},
 number = {4},
 issn = {1552-3098},
 journal = {TRO},
 doi = {10.1109/TRO.2020.3043723}
}

@inproceedings{Yigit.2023_IROS,
 author = {Yi{\u{g}}it, Arda and Nguyen, Tan-Sy and Gosselin, Cl{\'e}ment},
 title = {Exploiting the Kinematic Redundancy of a Backdrivable Parallel Manipulator for Sensing During Physical Human-Robot Interaction},
 pages = {9788--9793},
 booktitle = {IROS},
 year = {2023},
 doi = {10.1109/IROS55552.2023.10341495}
}

@article{Zhang.2021,
 author = {Zhang, Zengjie and Qian, Kun and Schuller, Bj{\"o}rn W. and Wollherr, Dirk},
 year = {2021},
 title = {An Online Robot Collision Detection and Identification Scheme by Supervised Learning and {B}ayesian Decision Theory},
 pages = {1144--1156},
 volume = {18},
 number = {3},
 issn = {1558-3783},
 journal = {IEEE Transactions on Automation Science and Engineering},
 doi = {10.1109/TASE.2020.2997094}
}
\end{document}